\documentclass[10pt,journal, compsoc]{IEEEtran}
\usepackage{amsmath,amsfonts}
\usepackage{algpseudocode}
\usepackage{algorithm}
\usepackage{array}
\usepackage[caption=false,font=footnotesize,labelfont=sf,textfont=sf]{subfig}
\usepackage{enumitem}
\usepackage{textcomp}
\usepackage{booktabs}
\usepackage{multirow}
\usepackage{stfloats}
\usepackage{url}
\usepackage{verbatim}
\usepackage{graphicx}
\usepackage{tabularx}
\usepackage{cite}
\usepackage{comment}
\usepackage{balance}
\usepackage[framemethod=default]{mdframed}
\newmdenv[backgroundcolor=orange, linecolor=orange]{highlightedbox}
\usepackage[dvipsnames]{xcolor}
\usepackage{soul}
\usepackage{colortbl}
\usepackage{multicol}
\usepackage{hyperref}
\usepackage[normalem]{ulem}
\hypersetup{colorlinks=false, urlcolor=false}

\definecolor{ModalitySelection}{HTML}{E1EEC4}
\definecolor{ClientSelection}{HTML}{F9D1D4}

\colorlet{Up}{blue!10}
\colorlet{StrongUp}{blue!20}
\colorlet{Down}{red!10}
\colorlet{StrongDown}{red!20}

\def\BibTeX{{\rm B\kern-.05em{\sc i\kern-.025em b}\kern-.08em
    T\kern-.1667em\lower.7ex\hbox{E}\kern-.125emX}}
\begin{document}

\title{
Communication-Efficient Multimodal Federated Learning: Joint Modality and Client Selection
}
    
\author{Liangqi Yuan,~\IEEEmembership{Graduate Student Member,~IEEE,}
        Dong-Jun Han,~\IEEEmembership{Member,~IEEE,}
        Su Wang,
        Devesh Upadhyay,~~\IEEEmembership{Senior Member,~IEEE}
        and Christopher G. Brinton,~\IEEEmembership{Senior Member,~IEEE}
\IEEEcompsocitemizethanks{\IEEEcompsocthanksitem This work was supported by the Office of Naval Research (ONR) under Grant N00014-23-C-1016, the National Science Foundation (NSF) under Grant CPS-2313109, and the Air Force Office of Scientific Research (AFOSR) under Grant FA9550-24-1-0083.
\IEEEcompsocthanksitem An abridged version of this paper was presented at the 2024 IEEE International Conference on Communications (ICC)~\cite{yuan2024fedmfs}.
\IEEEcompsocthanksitem L. Yuan and C. G. Brinton are with the School of Electrical and Computer Engineering, Purdue University, West Lafayette, 47907, USA. E-mail: \{liangqiy, cgb\}@purdue.edu.
\IEEEcompsocthanksitem D.-J. Han is with the Department of Computer Science and Engineering, Yonsei University, Seoul, South Korea. E-mail: djh@yonsei.ac.kr.
\IEEEcompsocthanksitem S. Wang is with the School of Electrical and Computer Engineering, Princeton University, NJ, 08540, USA. E-mail: hw5731@princeton.edu.
\IEEEcompsocthanksitem D. Upadhyay is with the Saab Inc., East Syracuse, 13057, USA. E-mail: deveshu@gmail.com.}}

\IEEEtitleabstractindextext{%
\begin{abstract}
Multimodal federated learning (MFL) aims to enrich model training in FL settings where clients are collecting measurements across multiple modalities. However, key challenges to MFL remain unaddressed, particularly in heterogeneous network settings where: (i) the set of modalities collected by each client is diverse, and (ii) communication limitations prevent clients from uploading all their locally trained modality encoders to the server. In this paper, we propose Multimodal Federated learning with joint Modality and Client selection (MFedMC), a communication-efficient MFL framework that tackles these challenges through a decoupled architecture and selective uploading. Unlike traditional holistic fusion approaches, MFedMC separates modality encoders and fusion modules: modality encoders are aggregated at the server for generalization across diverse client distributions, while fusion modules remain local to each client for personalized adaptation to individual modality configurations and data characteristics. Building on this decoupled design, our joint selection algorithm incorporates two main components: (a) A modality selection methodology for each client, which weighs (i) the impact of the modality, gauged by Shapley value analysis, (ii) the modality encoder size as a gauge of communication overhead, and (iii) the frequency of modality encoder updates, denoted recency, to enhance generalizability. (b) A client selection strategy for the server based on the local loss of modality encoders at each client. Experiments on five real-world datasets demonstrate that MFedMC achieves comparable accuracy to several baselines while reducing communication overhead by over 20$\times$. A demo video and our code are available at \href{https://liangqiy.com/mfedmc/}{https://liangqiy.com/mfedmc/}.
\end{abstract}

}

\maketitle

\IEEEdisplaynontitleabstractindextext

\IEEEpeerreviewmaketitle

\IEEEraisesectionheading{
\section{Introduction}
\label{Sec. Introduction}
}

Federated learning (FL) is a distributed machine learning (ML) approach in which users collaboratively train ML models through sharing model parameters rather than raw measurements~\cite{mcmahan2017communication,yuan2024decentralized,yuan2024digital,fang2025federated}. The FL approach establishes a federation of learners, each updating their model based on local data. Subsequently, these locally learned parameters are uploaded to a central server or shared with other clients. The local model within each client is then updated by integrating an aggregation of these parameters, ensuring that each model benefits from the collective learning experience. As Internet of Things (IoT) devices, such as smartphones, robots, and unmanned aerial vehicles (UAVs), are increasingly equipped with multimodal sensors, there has been growing interest in multimodal federated learning (MFL) frameworks~\cite{yu2021fedhar,wu2020fedhome,baghersalimi2023decentralized,lin2023federated,chen2024fedmbridge}. For example, in federated settings with connected and automated vehicles (CAVs), diverse sensors such as cameras, LiDAR, and radar provide complementary multimodal measurements that enable robust control decisions across varying weather conditions and fields of view~\cite{yuan2023peer,chellapandi2023survey,chellapandi2023federated}.

\begin{figure}[t]
\centering
\includegraphics[width=\linewidth]{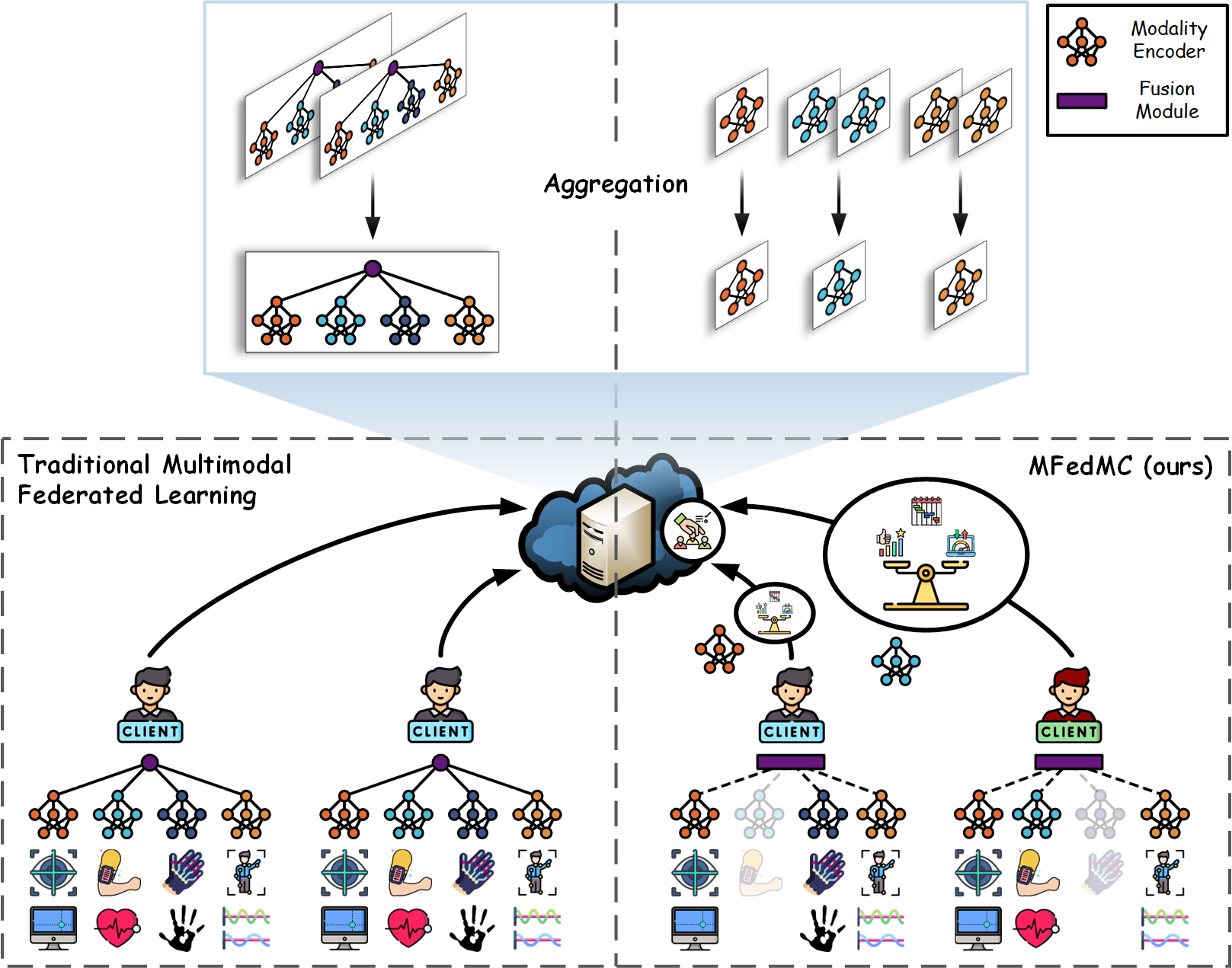}
\caption{Comparison between traditional multimodal federated learning vs. the proposed MFedMC.}
\label{Fig. MFedMC}
\end{figure}

Multimodal fusion combines information from two or more modalities to enhance ML model performance and robustness through cross-modality interactions~\cite{lee2019adaptive, li2022adaptive, zhang2023multimodal}. Fusion methods are generally categorized as data, feature, and decision, depending on the network layer where the modalities are merged. Data-level fusion concatenates at the input level (e.g., combining RGB and depth images into RGB-D). Feature fusion merges features from intermediate network layers, enabling nuanced interactions but increasing architectural complexity. Decision fusion aggregates outputs from separate models, offering flexibility but potentially losing fine-grained information during integration. This trade-off between information retention and fusion sophistication is a key consideration in the design of multimodal learning systems.

\subsection{Motivation}

In the context of the IoT, where devices measure a diverse set of modalities, most existing MFL frameworks call for massive parallel processing of extensive sensor data streams~\cite{kairouz2021advances,zhao2022multimodal}. They utilize multimodal fusion to boost model performance, especially in scenarios where heterogeneous clients lack one or more modalities. However, IoT devices often possess communication constraints due to bandwidth limitations, location, and varying capabilities. This has also been a key bottleneck in conventional (single modality) FL~\cite{sattler2019robust,lim2020federated,imteaj2021survey,fang2024submodel}. Thus, there is still a need to devise strategies to reduce communication overhead and improve learning efficiency within the MFL paradigm. We summarize the existing gaps and challenges in the current MFL frameworks as follows.
\begin{enumerate}[label=(\roman*)]
    \item \textbf{Client and Modality Heterogeneity} extends beyond statistical heterogeneity (non-IID data distributions) to encompass individual, group, and system differences~\cite{zhang2022fedada,chu2022mitigating, abdelmoniem2022empirical}. Individual heterogeneity manifests in feature variations (e.g., walking frequencies), group heterogeneity arises from physiological differences (e.g., left-hander vs. right-hander), and system heterogeneity stems from external factors (e.g., device age). Critically, in MFL, modality heterogeneity is prevalent where some clients may lack certain modalities entirely (e.g., some CAVs are equipped with LiDAR while others rely solely on vision-based systems), posing unique challenges in MFL scenarios.
    \item \textbf{Communication Efficiency} is influenced not only by the balance between performance and communication overhead, but also by the varying data sizes across different modalities and data types, leading to differences in modality encoder sizes. Furthermore, the complexity of information patterns inherent in different data modalities affects the ease of learning. For instance, compared to high-resolution image data, simpler data modalities like time-series radar data, despite potentially lower accuracy, may require only simple ML models for effective training and recognition due to their smaller data size and straightforward pattern recognition.
    \item \textbf{Impact of Modality and Client} in MFL are critical yet under explored areas, particularly in the context of communication costs and client participation. It remains an open question as to which modality or combination of modalities should be prioritized in the predictions, considering their information richness. Moreover, the extent of client participation, influenced by factors such as data availability and quality, also plays a vital role in determining the effectiveness of different modalities within the MFL framework. 
\end{enumerate}
Motivated by these challenges, we aim to answer the following key questions: 

\vspace{5pt}
\begin{quote}
\textit{\textbf{In resource-constrained and heterogeneous MFL settings, (i) how should we design the learning architecture to handle clients with heterogeneous modality sets while enabling both generalization and personalization, (ii) how should each client evaluate and select the most valuable modalities for uploading to balance performance and communication overhead, and (iii) how should the server determine which clients, each with different modality encoder combinations, to aggregate from to further optimize this trade-off?}}
\end{quote}

\subsection{Overview and Contribution}

We propose Multiodal Federated learning with joint Modality and Client selection (MFedMC), an MFL methodology tailored for clients with heterogeneously absent modalities. An overview of our proposed method and comparison with traditional MFL are given in Fig. \ref{Fig. MFedMC}. We propose to decouple modality encoders and fusion modules, where \textit{global modality encoders} generate predictions that serve as input to the individual \textit{local fusion module} in each client. This modular design renders our framework adaptable and beneficial across various applications that involve multiple modalities, thus facilitating the collaboration between clients. For example, CAVs utilize a combination of cameras, LiDAR, and radar, each providing complementary information essential for autonomous driving. Our framework supports modular modality encoders on vehicles equipped with these different sensors, enabling effective MFL across different vehicles. While modality encoders are uploaded to the server for aggregation to enhance generalization, each client retains the fusion module locally to improve personalization and confidence. Retaining the fusion module local to the client minimizes the risk of leakage of client sensitive information to the server either directly or indirectly via server side inference.

Building on this foundation, we introduce the concept of \textit{joint modality and client selection} to reduce communication overhead. First, \textit{selective modality communication} acknowledges that clients may not always have the capacity or necessity to upload all modality encoders (e.g., resource-constrained IoT applications). This approach is based on factors such as modality performance or its impact on the final decision, communication overhead, and recency. To evaluate the impact of each modality, we propose employing Shapley values~\cite{shapley1953value, young1985monotonic, sundararajan2020many, lundberg2017unified, lundberg2020local2global}, measured on the fusion module, to quantify their respective performances. Communication overheads are determined by quantifying the size of modality encoders, while recency is gauged by tracking the upload history of modalities. Complementing this, \textit{selective client uploading} is proposed to further reduce communication overheads and address the effects of client heterogeneity on the global model. This involves ranking each client based on the local loss of modality encoder, thereby optimizing both the efficiency and effectiveness of our FL framework.

In this paper, we present the following contributions:
\begin{itemize} 
    \item \textbf{Multimodal Federated Learning with Decoupled Encoders and Fusion (Sec.~\ref{Sec. Proposed MFedMC Framework}).} We propose a framework that decouples traditional holistic fusion approaches into separate modality encoders and fusion modules. Modality encoders are uploaded to a server to enhance generalizability, while individual fusion modules are retained locally to promote personalization and strengthen confidence. This decoupled approach enables modular functionality of modalities and naturally accommodates client heterogeneity and missing modality scenarios.
    \item \textbf{Communication-Efficient Joint Modality and Client Selection (Sec.~\ref{Sec. Modality Selection}, Sec.~\ref{Sec. Client Selection}).} We propose employing a Shapley component to quantify the impact of modality encoders, along with assessing their sizes to estimate communication overhead. Additionally, a recency term is introduced to track the frequency of uploads, facilitating selective modality communication. Furthermore, we implement selective client uploading, which leverages the local loss of modality encoder to quantify client heterogeneity. This joint modality and client selection strategy significantly reduces communication costs while maintaining model performance, thereby enhancing learning efficiency.
    \item \textbf{Five Real-World Experiments (Sec.~\ref{Sec. Results and Overall Comparison}).} Our proposed MFedMC is evaluated for performance and communication overhead against four baseline methods across five heterogeneous multimodal datasets, including wearable sensors, healthcare, language, and satellite datasets. Experiment results highlight the superior performance of MFedMC, achieving comparable accuracy while incurring less than 25\% of the communication overhead compared to baseline methods.
    \item \textbf{Analytics on Modality Impact (Sec.~\ref{Sec. Ablation Study: Modality Selection}).} We offer analyses on modality impact within the FL process. Utilizing Shapley values, we illustrate the interplay among modalities, revealing the dynamic impact of each modality on the final decision-making process within scenarios that take into account communication overhead and recency.
\end{itemize}

This paper is an extension of our previous paper~\cite{yuan2024fedmfs}. Building upon the foundation laid by~\cite{yuan2024fedmfs}, this paper introduces the following key contributions: 
(i) For modality selection, we incorporate a new metric, recency, to prevent overemphasis on certain modalities and maintain generalization. 
(ii) We introduce a client selection strategy tailored to MFL that synergistically optimizes communication overhead in conjunction with modality selection. 
(iii) Beyond the previously experimented ActionSense dataset, we have added four new datasets, covering domains such as wearable sensors, healthcare, language, and satellite imagery, providing a holistic demonstration of the framework's applicability. 
(iv) We extend our experimental suite with additional baselines and comprehensive ablation studies across diverse federated settings, including class heterogeneity, modality heterogeneity, long-tailed distributions, and client dynamics.
 
\subsection{Organization}
The rest of this paper is organized as follows. Section~\ref{Sec. Related Works} reviews related works, and Section~\ref{Sec. Methodology} details our MFedMC algorithm. We present our experiments and their corresponding results in Section~\ref{Sec. Experiment and Results}. Finally, we summarize our conclusions and future work in Section~\ref{Sec. Conclusion}.

\section{Related Works}
\label{Sec. Related Works}

\subsection{Multimodal Federated Learning}
MFL has gained increasing attention due to the inherently multimodal nature of real-world data and the complementary benefits offered by leveraging multiple modalities. Recently, a variety of algorithms have been proposed to improve the performance of MFL frameworks based on different fusion techniques, including data-level \cite{qi2023fl}, feature-level \cite{xiong2022unified}, and decision-level \cite{chen2022towards}. However, due to client heterogeneity, different clients may possess various modalities, which is also known as modality heterogeneity challenge \cite{zhao2022multimodal, chen2022fedmsplit, bao2023multimodal, feng2023fedmultimodal}. Most current MFL implementations are confined to end-to-end model architectures, resulting in either (i) inability to aggregate on the server because of different model architectures caused by varying modalities or (ii) significant performance degradation when using zero padding or random padding to compensate for missing modality data \cite{chen2022towards, le2024cross}. Several related works aim to address these challenges, such as utilizing pre-trained autoencoders to generate missing modalities \cite{zheng2023autofed}, relying on a complete public dataset at the server \cite{yu2023multimodal}, cross-modality reconstruction to recover missing modalities \cite{xiong2023client}, and others.

However, our approach circumvents the performance degradation associated with zero padding or random padding techniques by decoupling the learning process to focus exclusively on available modalities at each client. At the server side, we aggregate modality-specific encoders rather than complete models. The fusion modules remain localized due to structural variations across clients and serve as personalization mechanisms, as global encoders may not align with local data distributions. For example, a left-handed user's client would likely experience suboptimal performance with a global model predominantly trained on right-handed users' data, thus the fusion module functions to appropriately attenuate the weights of potentially misaligned modality encoders. Furthermore, our modality selection strategy leverages the decoupled fusion module to quantify the impact contribution of each modality at the client level.

\subsection{Selection Strategies in Federated Learning}

Multi-objective selection in network scenarios is not uncommon, with numerous related works selecting devices for resource optimization based on different trade-offs \cite{yuan2025local}. In FL, due to the extensive heterogeneity among clients, including information richness, information asymmetry, and security and privacy concerns, related works have proposed static or dynamic client selection functions for uploading and aggregation \cite{xu2023federated, luo2022tackling, yuan2023federated, fu2023client}, such as network heterogeneity \cite{wang2023device}, client dataset similarity \cite{pan2023contextual}, client training loss \cite{cho2020client}, and other schemes. In the MFL scenario, beyond the aforementioned client heterogeneity, a new heterogeneity challenge emerges: modality heterogeneity among clients. To date, within the MFL context, a preliminary test in FLASH \cite{salehi2022flash} involves clients randomly selecting and uploading one of three modality encoders or a fusion module for aggregation.

However, no research has yet explored modality selection in the MFL context, primarily because modality encoders differ in training difficulty and impact during fusion. Moreover, modality encoders evidently possess different parameter quantities, further resulting in varying communication overheads. Our approach addresses the multi-objective optimization problem of performance-communication trade-offs, dynamically selecting modalities during the MFL process. We also highlight the dynamic selection process in MFL, where selected modalities should differ in each communication round to maximize modality complementarity. Our method adaptively selects different modalities at various stages of the dynamic MFL process, choosing more easily trainable modalities in early stages and more difficult but information-rich modalities in later stages. This not only ensures the convergence speed of the MFedMC framework but also guarantees its overall superior accuracy.

\section{Formulation and Methodology}
\label{Sec. Methodology}

\subsection{Proposed MFedMC Framework}
\label{Sec. Proposed MFedMC Framework}

\textbf{System Model.}
Consider an MFL system with $K$ clients. Each client $k$ possesses a local dataset $\mathbb{D}^k$ with corresponding labels $Y^k$, comprising multimodal data represented as
\begin{equation}
    \mathbb{D}^k = \{\mathcal{D}^k_1, \mathcal{D}^k_2, \dots, \mathcal{D}^k_{M_k}\},
\end{equation}
where $\mathcal{D}^k_m$ denotes the dataset for modality $m$ (e.g., text, images, LiDAR), and $M_k$ represents the number of available modalities at client $k$. Note that $M_k$ may vary across clients due to modality heterogeneity, where some clients may lack certain sensing capabilities.

\noindent\textbf{Decoupled Architecture.}
To accommodate class and modality heterogeneities, we decouple the multimodal learning process into two components: global modality encoders and local fusion modules. Each client $k$ maintains a set of modality encoders
\begin{equation}
    \Theta^k = \{\theta^k_1, \theta^k_2, \ldots, \theta^k_{M_k}\},
\end{equation}
where each encoder $\theta^k_m$ extracts discriminative features from its corresponding modality dataset $\mathcal{D}^k_m$ and produces predictions. Specifically, for each modality $m$, the encoder generates a predicted label
\begin{align}
    \hat{y}^k_m &= \theta^k_m(\mathcal{D}^k_m), \\
    \widehat{\mathbb{Y}}^k &= \{\hat{y}^k_1, \hat{y}^k_2, \dots, \hat{y}^k_{M_k}\},
\label{Eq. Modality Encoder Predicted Label}
\end{align}
where $\widehat{\mathbb{Y}}^k$ denotes the collection of predictions from all modality encoders at client $k$. A local fusion module $\omega^k$ then aggregates these predictions to produce the final output:
\begin{align}
    \widehat{Y}^k &= \omega^k \left(\theta^k_1(\mathcal{D}^k_1), \theta^k_2(\mathcal{D}^k_2), \ldots, \theta^k_{M_k}(\mathcal{D}^k_{M_k})\right) \nonumber \\
                  &= \omega^k \left(\hat{y}^k_1, \hat{y}^k_2, \ldots, \hat{y}^k_{M_k}\right) = \omega^k \left(\widehat{\mathbb{Y}}^k\right).
\end{align}
where $\widehat{Y}^k$ represents the final prediction for client $k$. This decoupled design offers several advantages: (i) modality encoders $\theta_m$ can be collaboratively trained across clients and shared via the server to enhance generalizability, (ii) fusion modules $\omega^k$ remain local to each client, enabling personalized adaptation to individual heterogeneities such as modality availability, user characteristics, and noise levels, and (iii) the modular structure naturally accommodates missing modality scenarios without requiring architectural modifications.

\noindent\textbf{Learning Objectives.}
At each client $k$, all modality encoders are trained in parallel, with each encoder independently minimizing the discrepancy between its predicted labels and the true labels using a loss function $f$ :
\begin{equation}
    \min_{\theta^k_m} f(Y^k, \hat{y}^k_m), \quad \forall m \in \{1, 2, \ldots, M_k\}.   
\end{equation}
Subsequently, the fusion module aims to minimize the discrepancy between the true labels and the fused predictions across all modalities:
\begin{equation}
\min_{\omega^k} f(Y^k, \widehat{Y}^k).
\end{equation}

\begin{figure*}[t]
\centering
\includegraphics[width=0.8\linewidth]{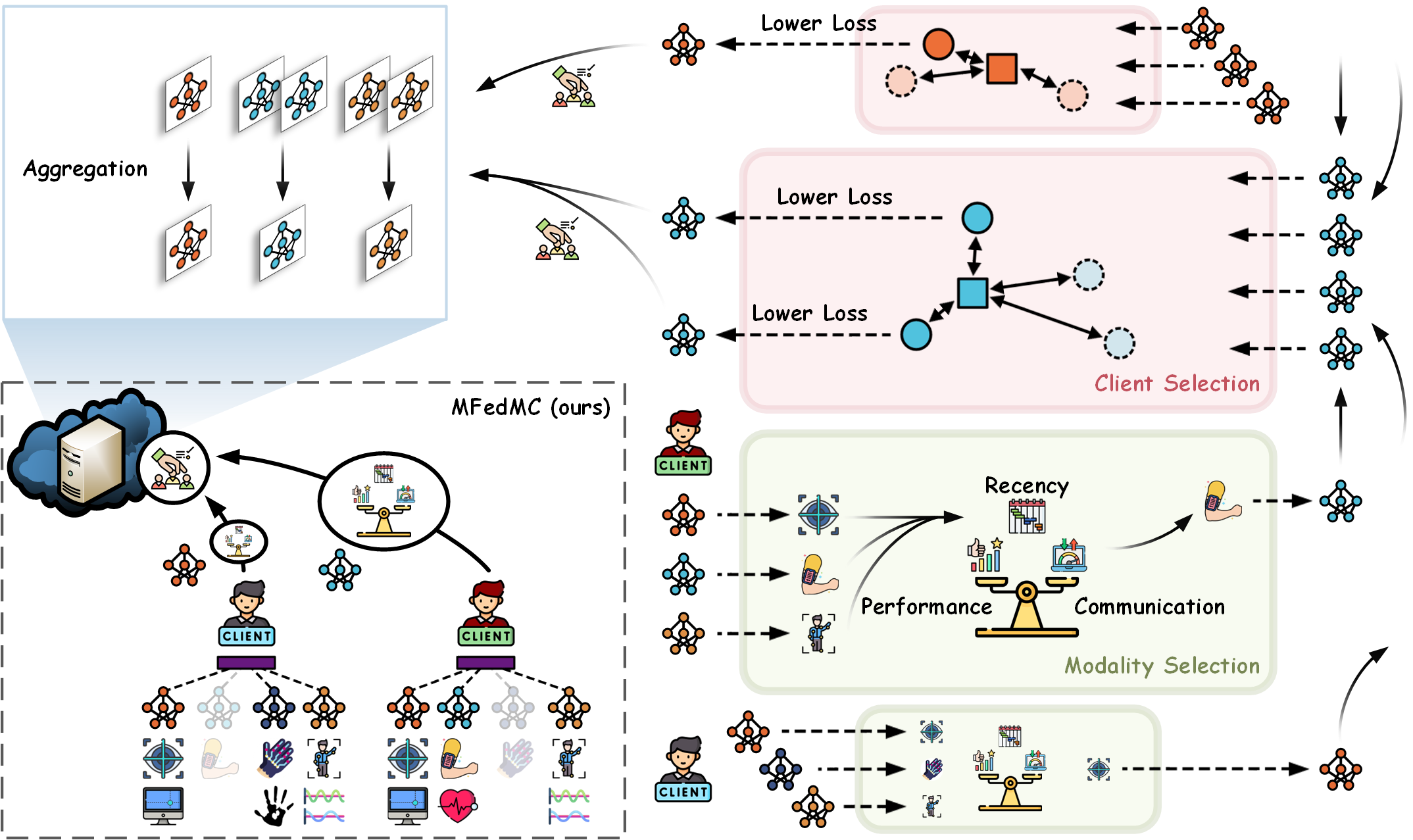}
\caption{System diagram of the proposed MFedMC illustrating the process of \textbf{Modality Selection} and \textbf{Client Selection} as detailed in Algorithm~\ref{Alg. MFedMC}.}
\label{Fig. JMCS}
\end{figure*}

\noindent\textbf{Joint Selection Overview.}
The MFedMC algorithm, illustrated in Fig.~\ref{Fig. JMCS} and detailed in Algorithm~\ref{Alg. MFedMC}, achieves communication-efficient collaborative learning through joint client and modality encoder selection. During each communication round, rather than requiring all clients to upload all modality encoder parameters, MFedMC selects a subset of clients and determines which modality encoders each selected client should upload. This selective aggregation substantially reduces communication overhead while maintaining model performance. The server aggregates the received encoder parameters and broadcasts the updated global encoders back to clients, which then perform local training on both encoders and fusion modules. Within a fixed communication budget, MFedMC maximizes model accuracy by intelligently determining which clients and modality encoders contribute most effectively to the global model in each round. In the following sections, we provide detailed descriptions of the joint selection mechanism and training procedure.

\renewcommand{\Procedure}[2]{\item[\hspace{0em}\textbf{#1}] #2}
\renewcommand{\EndProcedure}{}
\makeatletter
\newcommand{\setalglineno}[1]{%
  \setcounter{ALG@line}{\numexpr#1-1}}
\makeatother

\begin{algorithm*}[h]
\small
\caption{MFedMC: Multimodal Federated Learning with Joint \colorbox{ModalitySelection}{Modality Selection} and \colorbox{ClientSelection}{\strut Client Selection}}
\label{Alg. MFedMC}
\textbf{Input:} Communication rounds ($T$), client datasets ($\{\mathbb{D}^k\}_{k=1}^K$), client label sets ($\{Y^k\}_{k=1}^K$), local training epoch ($E$), initial models ($\theta^k_{m,0}$ and $\omega^k_0$), loss function ($f$), learning rate ($\eta$), modality encoder upload count ($\gamma$), modality selection weights ($\alpha_s$, $\alpha_c$, and $\alpha_r$), client selection ratio ($\delta$)

\textbf{Output:} Generalized global modality encoders ($\theta_m$) and personalized local fusion modules for each client ($\omega^k$)

\begin{algorithmic}[1]
\Procedure{\# Global Iteration}{}
\For{$t=0$ {\bfseries to} $T-1$}
    \begin{algorithmic}[1]
    \Procedure{}{}
    
    \begin{algorithmic}[1]
    \Procedure{\# Local Learning}
    \For{each client $k$ {\bfseries in parallel}}
        \For{each data modality $m \in \mathcal{M}^k$ {\bfseries in parallel}}
            \State Train modality encoder: $\theta^{k,t+1}_m \gets \arg\min_{\theta^{k,t}_m} f(Y^{k}, \hat{y}^{k,t}_m)$ for $E$ epochs
        \EndFor
        \State \textbf{(Stage \#1)} Freeze encoders $\{\theta^{k,t+1}_m\}_{m \in \mathcal{M}^k}$ and train fusion module $\omega^{k,t+1}$ for $E$ epochs
    \EndFor
    \EndProcedure
    \end{algorithmic}

    \begin{algorithmic}[1]
    \Procedure{\# Modality Selection}{}
    \State Compute the Shapley values for each modality $\Phi^{k,t}$ \Comment{(\ref{Eq. Shapley}), (\ref{Eq. Shapley Value Set})}
    \State Compute the modality encoder size $\Bar{\Theta}^k$ \Comment{(\ref{Eq. Modality Encoder Size})}
    \State Compute the recency $\mathcal{T}^k_m$ \Comment{(\ref{Eq. Recency})}
    \State \colorbox{ModalitySelection}{Clients select the modality encoders with top-$\gamma$ priority for uploading and update recency} \Comment{(\ref{Eq. Top Priority}), (\ref{Eq. Selected Modality}), (\ref{Eq. Selected Modality Encoder})}
    \EndProcedure
    \end{algorithmic}	

    \begin{algorithmic}[1]
    \Procedure{\# Client Selection \& Server Aggregation}{}
    \State \colorbox{ClientSelection}{Server selects top-$\delta$ clients with lowest modality encoder losses} \Comment{(\ref{Eq. Top Loss}), (\ref{Eq. Client Selection}), (\ref{Eq. Client Selection Model})}
    \For{each data modality $m \in \mathcal{M}$}
        \State Aggregate uploaded modality encoder $\theta^{t+1}_m$ \Comment{(\ref{Eq. Aggregation})} 
    \EndFor
    \EndProcedure
    \end{algorithmic}
    
    \begin{algorithmic}[1]
    \Procedure{\# Local Deploying}{}
    \For{each client $k$ {\bfseries in parallel}}
        \For{each data modality $m \in \mathcal{M}^k$ {\bfseries in parallel}}
            \State Download and deploy global encoder $\theta^{t+1}_m$
        \EndFor
        \State \textbf{(Stage \#2)} Freeze encoders $\{\theta^{t+1}_m\}_{m \in \mathcal{M}^k}$ and fine-tune fusion module $\omega^{k,t+1}$ for $E$ epochs
    \EndFor
    \EndProcedure
    \end{algorithmic}
    
    \EndProcedure
    \end{algorithmic}
    
\setalglineno{2}
\State \hspace{-2em} \textbf{end for}
\State \hspace{-2em} \textbf{return} Global encoders $\{\theta^T_m\}_{m \in \mathcal{M}}$ and local fusion modules $\{\omega^{k,T}\}_{k=1}^K$

\end{algorithmic}	
\end{algorithm*}

\subsection{Modality Selection} 
\label{Sec. Modality Selection} 

Due to the resource constraints on edge devices serving as clients, they may not possess adequate storage capacity to house extensive multimodal data, computational capability to learn on multimodal datasets, or communication bandwidth to upload several models to the server. Thus, we introduce three metrics to assist clients to determine whether to upload their models to the server:
\begin{itemize}
    \item \textbf{Shapley value} ($\varphi$) represents the impact of a modality encoder on the final prediction, where a higher value of $\varphi$ corresponds to a higher priority $\uparrow$.
    \item \textbf{Modality encoder size} ($|\theta|$) pertains to the communication overhead, where a lower value of $|\theta|$ corresponds to a higher priority $\downarrow$.
    \item \textbf{Recency} ($\mathcal{T}$) denotes the freshness or recentness of a client's modality encoder upload, where a higher value indicates that the modality has not been updated recently, thus it is given higher priority $\uparrow$.
\end{itemize}

\noindent \textbf{Shapley Value (Impact of Modality).} During each communication round, clients evaluate the impact of the modality encoders $\Theta^k$ on the outcomes utilizing interpretability techniques and choose to upload only a selected subset of encoders. We consider using the Shapley value as an assessment to evaluate the relationship between input $\widehat{\mathbb{Y}}^k$ and output $\widehat{Y}^k$ of the fusion module $\omega^k$:
\begin{equation}
    \begin{small}
    \begin{aligned}
    \varphi^k_m = \sum_{\mathcal{Y} \subseteq \widehat{\mathbb{Y}}^k \setminus \{m\}} \frac{|\mathcal{Y}|!(|\widehat{\mathbb{Y}}^k|-|\mathcal{Y}|-1)!}{|\widehat{\mathbb{Y}}^k|!} \left( \omega^k(\mathcal{Y} \cup \{m\}) - \omega^k(\mathcal{Y}) \right),
    \end{aligned}
    \end{small}
\label{Eq. Shapley}
\end{equation}
where $\varphi^k_m$ is the Shapley value of input modality $m$, $\mathcal{Y}$ is a subset of $\widehat{\mathbb{Y}}^k$ excluding modality $m$, and $\omega^k(\mathcal{Y})$ is the predicted value using only modalities in set $\mathcal{Y}$. For all modalities, we assess the magnitude of each Shapley value by taking its absolute value and construct the following set:
\begin{equation}
    \Phi^k = \left\{|\varphi^k_1|, |\varphi^k_2|, \dots, |\varphi^k_{M_k}|\right\}.
\label{Eq. Shapley Value Set}
\end{equation}

\noindent \textbf{Modality Encoder Size (Communication Overhead).} Given modality encoders with parameters $\Theta^k = \{\theta^k_1, \theta^k_2, \ldots, \theta^k_{M_k}\}$, the communication overhead for each modality encoder is directly proportional to the model size given by
\begin{equation}
    \Bar{\Theta}^k = \left\{|\theta^k_1|, |\theta^k_2|, \ldots, |\theta^k_{M_k}|\right\}. 
\label{Eq. Modality Encoder Size}
\end{equation}

\noindent \textbf{Recency.} Our primary aim with the recency metric is to encourage clients to update a specific modality encoder to the server. This is to ensure that the MFedMC framework doesn't overly prioritize data modalities that are more easily obtainable, possess simpler data structures, or have conspicuous features, thus sidelining the diversity of other data modalities. To encapsulate the concept of recency in our model, we denote $\mathcal{T}^k_m$ as:
\begin{equation}
    \mathcal{T}^k_m = t - t^k_m - 1,
\label{Eq. Recency}
\end{equation}
where $t = 1, 2, \dots, T$ represents the current communication round, and $t^k_m$ indicates the communication round during which the modality encoder from modality $m$ of client $k$ was last uploaded.

\noindent \textbf{Priority (Composite Score).} Considering the impact of the modality encoder, as quantified by the Shapley value, the communication overhead as characterized by the modality encoder size, and the timeliness of model updates captured by the recency, we propose priority $P$ as a composite score. To derive the priority, we proceed with individual normalization for each criterion:
\begin{equation}
    \begin{small}
    \left\{\!
    \begin{aligned}
    \tilde{\varphi}^k_m &= \frac{\varphi^k_m - \min(\Phi^k)}{\max(\Phi^k) - \min(\Phi^k)}, \\
    |\tilde{\theta}^k_m| &= \frac{\theta^k_m - \min(\Bar{\Theta}^k)}{\max(\Bar{\Theta}^k) - \min(\Bar{\Theta}^k)}, \\
    \tilde{\mathcal{T}}^k_m &= \frac{\mathcal{T}^k_m}{t},
    \end{aligned}
    \right. \text{for } m = 1, 2, \dots, M_k,
    \end{small}
\label{Eq. Normalization}
\end{equation}
where $\tilde{\varphi}^k_m$ represents the normalized Shapley Value, $|\tilde{\theta}^k_m|$ denotes the normalized communication overhead, and $\tilde{\mathcal{T}}^k_m$ indicates the normalized recency. With a focus on identifying modality encoders for server communication, we devise the priority $P^k_m$ for each modality and the corresponding set $\mathcal{P}^k$ to determine whether modality encoders should be sent to the server. 
They are formulated as:
\begin{equation}
\begin{aligned}
    P^k_m &= \alpha_s \cdot \tilde{\varphi}^k_m + \alpha_c \cdot (1 - |\tilde{\theta}^k_m|) + \alpha_r \cdot \tilde{\mathcal{T}}^k_m, \\
    \mathcal{P}^k &= \left\{P^k_1, P^k_2, \dots, P^k_{M_k}\right\},
\end{aligned}
\label{Eq. Priority}
\end{equation}
where $\alpha_s$, $\alpha_c$, and $\alpha_r$ are the predetermined metric weights, satisfying $\alpha_s + \alpha_c + \alpha_r = 1$. Naturally, a modality with the maximal priority is considered optimal.

\noindent \textbf{Modality Selection.} To streamline our decision-making, we focus on modalities with the highest $\gamma$ priorities:
\begin{equation}
    \mathcal{P}^k_\gamma = \mathrm{top}_\gamma\max (\mathcal{P}^k)
\label{Eq. Top Priority}
\end{equation}
Hence, the selected modality set of client $k$ becomes:
\begin{equation}
    \mathcal{M}^k_\gamma = \left\{ m \,:\, P^k_m \in \mathcal{P}^k_\gamma, \text{ for } m = 1, 2, \dots, M_k \right\}.
\label{Eq. Selected Modality}
\end{equation}
With $\mathcal{M}^k_\gamma$ determined, the set of modality encoders ready for communication to the server by the client $k$ can be defined as:
\begin{equation}
    \Theta^k_\gamma = \left\{\theta^k_m : \theta^k_m \in \Theta^k \, \text{and} \, m \in \mathcal{M}^k_\gamma\right\},
\label{Eq. Selected Modality Encoder}
\end{equation}
where the set $\Theta^k_\gamma$ represents the modality encoders corresponding to the top-$\gamma$ priority from $\mathbb{S}^k$. Each client will upload a data packet with various details to the server for aggregation, including model parameters $\theta^k$, modality information $m$, the number of samples $|\mathcal{D}^k_m|$, among others. Likewise, upon downloading from the server, this information will also be retrieved. Note that only model $\theta^k$ will be uploaded/downloaded to/from the server. The fusion module $\omega^k$ varies across clients, determined by the unique deployment scenarios of each client, such as geographical location, operational duration, external interference, etc.

\subsection{Client Selection and Joint Selection}
\label{Sec. Client Selection} 

\noindent \textbf{Client Selection.} Considering the heterogeneity of clients in practice, the proposed MFedMC is designed to further optimize communication overhead and enhance learning efficiency through a client selection strategy. During the training process of the client models, we consider loss $\ell$ as a selection criterion, focusing on the modalities chosen to communicate to the server. For each modality $m$, the set $\mathcal{L}_{\gamma}$ includes the loss values $\ell^k_{m}$ from each client $k$ that has modality $m$ in their selected set $\mathcal{M}^k_\gamma$:
\begin{equation}
\mathcal{L}_{\gamma} = \left\{ \ell^k_{m} \mid m \in \mathcal{M}^k_\gamma \text{ and } k = 1, 2, \dots, K \right\}.
\end{equation}
This collection $\mathcal{L}_{\gamma}$ allows the server to make informed decisions about which clients to select to participate in the learning process, based on the reported loss values across the different modalities. The set of the lowest $\lceil\delta K\rfloor$ loss values can be expressed as:
\begin{equation}
\mathcal{L}^\delta_{\gamma} = \mathrm{top}_{\lceil\delta K\rfloor}\min (\mathcal{L}_{\gamma})
\label{Eq. Top Loss}
\end{equation}
where the ratio parameter $\delta$ establishes the proportion of the lower loss values to be considered. This strategy utilizes the top-($\lceil\delta K\rfloor$) criterion to pick a subset of clients, concentrating on those with the lower loss values, which are indicative of their potential effect on the learning efficiency. $\lceil \cdot \rfloor$ denotes the rounding operation. Hence, the set of selected clients can be expressed as:
\begin{equation}
    \mathcal{K}^\delta_{\gamma} = \left\{ k \,:\, \ell^k \in \mathcal{L}^\delta_{\gamma}, \text{ for } k = 1, 2, \dots, K \right\}.
\label{Eq. Client Selection}
\end{equation}

\noindent \textbf{Joint Selection.} In summary, the joint selection process is completed by first identifying the modality set $\mathcal{M}^k_\gamma$ for each client $k$, followed by selecting the client set $\mathcal{K}^\delta_{\gamma}$ based on the loss values. This initial selection prioritizes modalities that are most significant for each client. Subsequently, client selection identifies clients based on the loss. This sequential selection process ensures that we first optimize the selection of modalities before determining which clients will contribute to the learning process. The identification of the client set $\mathcal{K}^\delta_{\gamma}$ and the final model set to communicate to the server is then achieved as:
\begin{equation}
    \Theta^\delta_\gamma = \left\{\theta^k_m : \theta^k_m \in \Theta^k_\gamma \, \text{and} \, k \in \mathcal{K}^\delta_{\gamma}\right\},
\label{Eq. Client Selection Model}
\end{equation}
where the set $\Theta^k_\gamma$ represents the modality encoders corresponding to the top-$\gamma$ priority defined in (\ref{Eq. Selected Modality Encoder}). Through joint modality and client selection, the expected reduction in communication overhead is quantified by the ratio $\gamma/\bar{M} \times \delta$, where $\bar{M} \geq 2$ denotes the average number of modalities per client. The factor $0 < \gamma/\bar{M} < 1$ represents the communication overhead reduction achieved through modality selection, while $0 < \delta < 1$ represents the communication overhead reduction achieved through client selection. To illustrate, consider an MFL environment comprising 100 clients, each utilizing an average of 3 modalities. In a conventional MFL paradigm, each training round necessitates 300 modality encoder uploads (100 clients $\times$ 3 modalities). Employing our joint selection strategy with parameters $\gamma = 1$ and $\delta = 0.2$, the system requires merely 20 modality encoder uploads each round, constituting a 93.3\% reduction in communication overhead. With this strategic joint selection approach, the observed performance degradation is significantly smaller than the corresponding reduction in communication overhead, thus effectively achieving our objective.

\subsection{Server Aggregation and Local Retention}
\label{Sec. Server Aggregation and Local Retention}

\noindent\textbf{Server Aggregation: Learning Generalizable Modality Encoders.}
To achieve robust cross-client generalization, modality encoders are aggregated at the server. Upon receiving the selected encoders from participating clients, the server performs weighted aggregation for each modality $m$ as
\begin{equation}
    \theta_m \leftarrow \sum_{\theta^k_m \in \Theta^\delta_{\gamma}} \frac{|\mathcal{D}^k_m|}{\sum_{k=1}^{K_m}|\mathcal{D}^k_m|} \theta^k_m,
\label{Eq. Aggregation}
\end{equation}
where $K_m$ denotes the number of clients that uploaded the encoder $\theta_m$ in the current round, and $|\mathcal{D}^k_m|$ represents the sample count of modality $m$ in client $k$.

\noindent\textbf{Local Retention: Enabling Personalized Fusion Modules.}
In stark contrast, fusion modules $\omega^k$ remain strictly local and are never transmitted to the server. This deliberate design choice enables personalization: each fusion module specializes in integrating modality predictions according to its unique local context, including modality heterogeneity, user-specific patterns, device characteristics, and individual data distributions. The fusion module undergoes a two-stage training protocol:
\begin{itemize}
    \item \textbf{Stage \#1:} After local encoder training, the fusion module is trained with frozen encoders to calculate Shapley values for modality selection.
    \item \textbf{Stage \#2:} Upon receiving updated global encoders, each client fine-tunes its fusion module to learn optimal integration strategies tailored to local conditions.
\end{itemize}

\noindent\textbf{Modality Scalability of Our Decoupled Architecture.}
Our decoupled architecture naturally accommodates dynamic modality configurations. When a client's modality set changes (e.g., adding or removing sensors), it simply downloads the corresponding global encoders from the server, adjusts the fusion module architecture, and retrains only the lightweight fusion module with frozen encoders, without retraining the entire model. In terms of computational cost, training modality encoders incurs the same overhead as classical multimodal learning, and since encoders are trained in parallel (Algorithm~\ref{Alg. MFedMC}), the wall-clock training time is comparable to classical approaches. The additional overhead stems from Shapley value calculation, which grows exponentially with modalities ($\mathcal{O}(M_k 2^{M_k})$). To address this, we employ tree-based fusion modules and interventional feature perturbation~\cite{lundberg2020local2global} to reduce complexity from $\mathcal{O}(N_{\omega^k} L_{\omega^k} M_k 2^{M_k})$ to $\mathcal{O}(N_{\omega^k} L_{\omega^k} H_{\omega^k} |D_k|)$, where $N_{\omega^k}$, $L_{\omega^k}$, $H_{\omega^k}$, and $|D_k|$ denote the number of trees, leaves, maximum depth, and background dataset size. We further subsample background data ($|D'_k| \ll |D_k|$) for efficient estimation, ensuring MFedMC scales effectively as clients incorporate increasing numbers of modalities.

\begin{table*}[t] 
\caption{Description of Datasets}
\label{Table Dataset}
\centering
\vspace{-2mm}
\begin{tabular}{@{}lcccc@{}}
\toprule
\textbf{Dataset} & \textbf{Client} & \textbf{Task} & \textbf{Modality} & \textbf{Feature} \\
\midrule
\multirow{4}{*}{ActionSense \textsuperscript{1}} & \multirow{4}{*}{9 Subjects} & \multirow{4}{*}{20 Kitchen Activities} & Eye Tracking & $2$ \\
& & & EMG $(\times2)$ (Left and Right Arm) & $8 (\times2)$ \\
& & & Tactile $(\times2)$ (Left and Right Hand) & $32\times32 (\times2)$ \\
& & & Body Tracking & $22\times3$ \\
\midrule
\multirow{2}{*}{UCI-HAR} & \multirow{2}{*}{30 Subjects} & \multirow{2}{*}{6 Daily Activities} & Accelerometer & $128\times3$ \\
& & & Gyroscope & $128\times3$ \\
\midrule
\multirow{2}{*}{PTB-XL} & \multirow{2}{*}{39 Hospitals} & \multirow{2}{*}{5 Diagnosis} & Limb Lead ECG \textsuperscript{2} & $1000\times6$ \\
& & & Precordial Lead ECG \textsuperscript{3} & $1000\times6$ \\
\midrule
\multirow{2}{*}{MELD} & \multirow{2}{*}{42 Speakers} & \multirow{2}{*}{4 Emotions} & Audio & Client max length$\times80$ \textsuperscript{4} \\
& & & Text & $100$ \\
\midrule
\multirow{2}{*}{DFC2023} & 10 Cities of GF2 Satellite & \multirow{2}{*}{12 Roof Types} & SAR & $32\times32\times1$ \\
& $+$ 17 Cities of SV Satellite & & Optical & $32\times32\times3$ \\
\bottomrule
\end{tabular}

\vspace{2pt}
\raggedright
\noindent{\textsuperscript{1} Subjects 06 through 09 miss both left and right tactile data. \hfill \textsuperscript{3} Precordial Lead ECG refers to V1–V6 leads. 

\textsuperscript{2} Limb Lead ECG refers to I, II, III, aVL, aVR, aVF leads. \hfill \textsuperscript{4} Client max length refers to the max length of audio utterances for each client.

}
\end{table*}

\section{Experiment and Results}
\label{Sec. Experiment and Results}

\subsection{Dataset}
\begin{figure}[t]
    \centering
    \includegraphics[width=1\linewidth]{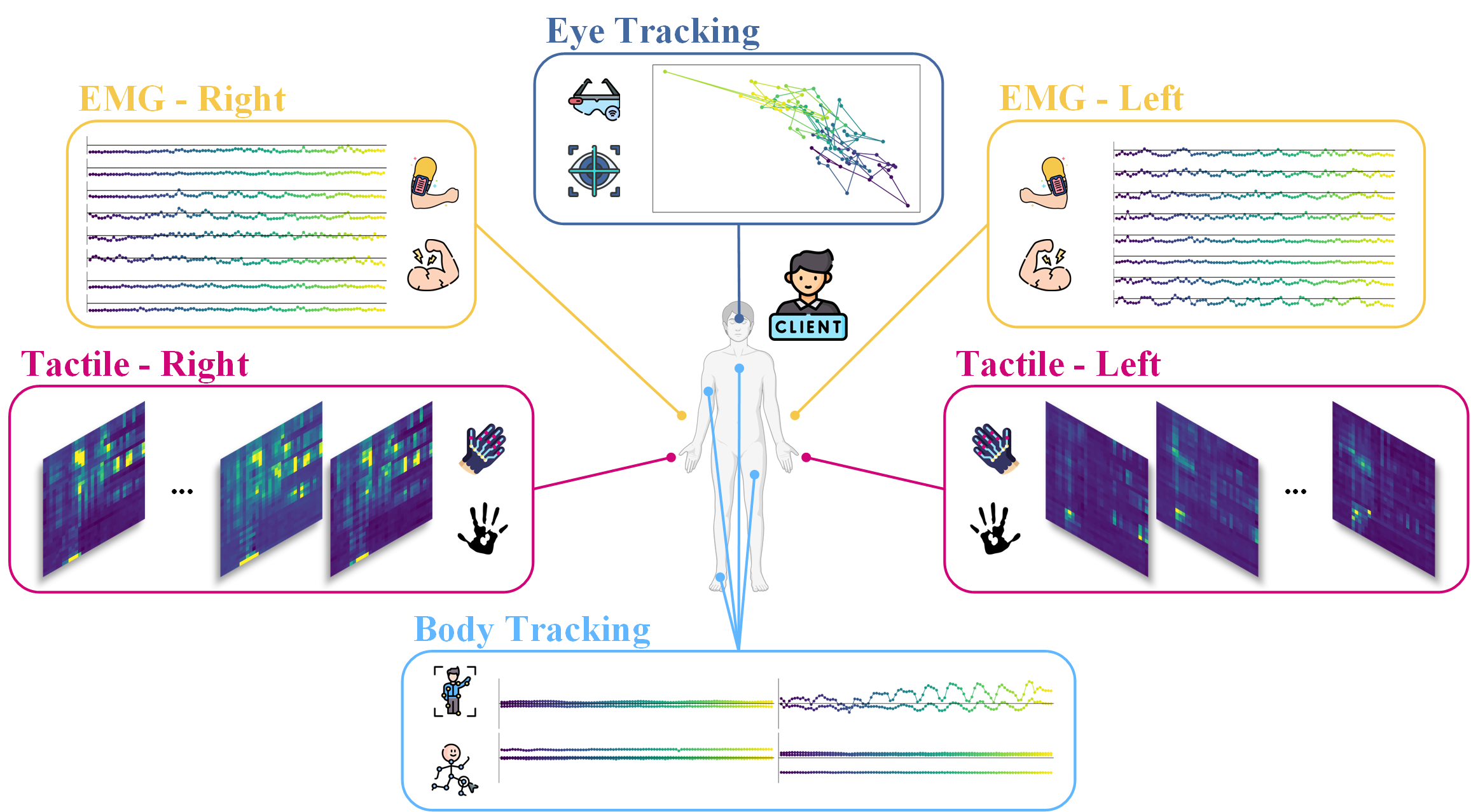}
    \caption{Data visualization of ActionSense dataset for subject 00 engaged in peeling a potato.}
    \label{Fig. ActionSense Dataset}
\end{figure}

Our experiments are conducted across five diverse multimodal real-world datasets, each incorporating varying numbers of clients, modalities, application contexts, and data types and structures, as depicted in Table~\ref{Table Dataset}. Specifically, we consider:
\begin{enumerate}[label=(\roman*)]
    \item ActionSense~\cite{delpreto2022actionsense} is an extensive multimodal dataset that captures human daily activities. It integrates data from six different types of wearable sensors, documenting human interactions with objects and the environment in a kitchen context. Activities include tasks such as peeling a cucumber, slicing a potato, cleaning a plate with a sponge, and so forth. Fig. \ref{Fig. ActionSense Dataset} illustrates the six modality data from the ActionSense dataset for Subject 00 while peeling a potato.

    \item UCI-HAR~\cite{anguita2013public}, similar to ActionSense, employs wearable sensors to monitor people during routine activities such as Walking, Walk Downstairs, Sitting, etc. Compared to ActionSense, UCI-HAR features a larger number of subjects but fewer modalities.
    \item PTB-XL~\cite{wagner2020ptb}, a substantial electrocardiography (ECG) dataset, comprises diverse clinical patient data, featuring 12-lead ECGs from various hospitals. Following established protocols in \cite{strodthoff2020deep}, we treat the Limb Lead ECG and the Precordial Lead ECG as separate modalities. Each data entry includes five labels: Normal, Myocardial Infarction, ST/T Change, Conduction Disturbance, and Hypertrophy.
    \item MELD~\cite{poria-etal-2019-meld} is a natural language processing (NLP) dataset derived from dialogues in the TV-series Friends. In each dialogue scene, each speaker is represented by audio and text modalities, reflecting different emotions such as neutral, sadness, joy, and anger.
    \item The 2023 IEEE GRSS Data Fusion Contest (DFC23)~\cite{mrnt-8w27-22} is a large-scale dataset featuring rooftop satellite imagery of buildings. It contains images from the Gaofen-2 (GF2) and SuperView-1 (SV) satellites, covering rooftops in seventeen cities across six continents. The dataset includes synthetic aperture radar (SAR) and optical images, categorizing twelve types of rooftops.
\end{enumerate}

We test our method against baselines on the aforementioned five datasets, considering two primary evaluation metrics: (i) accuracy under communication constraints and (ii) communication overhead required to achieve target accuracy. We establish target accuracy thresholds of 85\%, 60\%, 50\%, 50\%, and 60\% for ActionSense, UCI-HAR, PTB-XL, MELD, and DFC23 datasets, respectively. On this foundation, we conduct extensive experiments comparing our method with baselines under the following four client distribution scenarios:
\begin{itemize}
    \item In the natural distribution setting, we maintain the original client division inherent in the datasets that authentically reflects real-world scenarios, designating entities such as human subjects, speakers, hospitals, and satellites as distinct clients. This setting naturally embodies system-level heterogeneities inherent to individual clients, including variations in noise levels, sensor calibration differences, device age, data collection protocols, and environmental conditions across different deployment sites.
    \item In the IID setting, we shuffle all data and then redistribute these shuffled samples to all clients. In this scenario, individual, group, and system heterogeneities are significantly diminished due to the random and uniform distribution of samples.
    \item In the class non-IID setting, we allocate data to clients according to a Dirichlet distribution Dir($\beta$), where $\beta$ is the concentration parameter controlling the degree of class imbalance. A smaller $\beta$ value results in more heterogeneous class distributions across clients, with each client predominantly receiving samples from a limited subset of classes.
    \item In the modality non-IID setting, we randomly remove modalities from clients at certain probabilities. This leads to each client having a different combination of models while missing some modalities.
\end{itemize}

\subsection{Experiment Setup}

\noindent\textbf{Training Setup.}
To ensure a fair comparison across datasets (i -- iv), we initially reshape all data modalities into a two-dimensional format, i.e., time $\times$ features. For all modalities, we employ a consistent long short-term memory (LSTM) network structure, consisting of a single LSTM layer with 128 hidden units, followed by a fully connected layer. The learning rate $\eta$ for these LSTM models is set to 0.1. For the dataset (v), DFC23, which is composed entirely of images, we use a convolutional neural network (CNN) with one $5\times5$ convolution layer containing 32 channels, followed by a rectified linear unit (ReLU) activation, a $2\times2$ max pooling, and a fully connected layer. The learning rate $\eta$ for these CNN models is set at 0.01. We adopt cross-entropy loss with stochastic gradient descent (SGD) as the optimizer, a batch size of 32, and a local training epoch $E=5$.

\noindent\textbf{Baselines.}
At a high-level, we use two types of baselines for comprehensive comparisons: state-of-the-art (SOTA) MFL frameworks, and ablation studies of our method. We include five SOTA MFL frameworks in our analysis: FL-FD~\cite{qi2023fl}, MMFed~\cite{xiong2022unified}, FedMultimodal~\cite{feng2023fedmultimodal}, FLASH~\cite{salehi2022flash}, and Harmony~\cite{ouyang2023harmony}. Furthermore, to validate the effectiveness of the proposed MFedMC, we conduct comprehensive ablation studies in which the modality selection strategy, client selection strategy, and joint selection strategy are systematically replaced with random selection mechanisms. To ensure a fair systematic comparison within the FL context, we maintain consistent configurations across all methods, including identical base network architectures (LSTM for datasets (i -- iv), CNN for dataset (v)), uniform hyperparameters (learning rate, batch size, optimizer, loss function, local training epoch, etc.), among others. We do not incorporate various specialized techniques present in the original baseline papers, such as co-attention mechanisms, to isolate the impact of the core algorithmic differences. For method-specific configurations unique to individual baselines, we adopt the recommended settings from their original publications when explicitly specified, or otherwise select the most equitable configuration to ensure fair comparison.

\noindent\textbf{Our MFedMC.} For the proposed MFedMC framework, in addition to the configurations mentioned above, these modality encoders do not output logarithmic probabilities but rather provide definitive predicted categories ($\widehat{\mathbb{Y}}$) for the fusion module ($\omega$). We employ Random Forest (RF) as our fusion module, optimizing it to facilitate a more computationally efficient calculation of Shapley values, with the number of trees set at $N_{\omega_k}=10$. To further reduce computational complexity, we subsampling the dataset for each client $k$, constructing a sparse background dataset with samples $|D'_k| = 50$ for all clients $k = 1, 2, \ldots, K$, specifically to compute the Shapley values. The sparse background dataset is created through random sampling from the original dataset, ensuring a representative yet significantly smaller subset of the data.

\definecolor{Gray}{gray}{0.96}
\newcolumntype{a}{>{\columncolor{Gray}}c}

\begin{table*}[t]
\caption{Main Results - Overall Comparison with Baselines: \\ (i) Accuracy under 5 MB Communication Constraint and (ii) Communication Overhead to Achieve Target Accuracy}
\label{Table Overall Comparison}
\centering
\vspace{-2mm}
\resizebox{\linewidth}{!}{
\begin{tabular}{@{}l||ccccc||ccccc@{}}
\toprule
& \multicolumn{5}{c||}{\textbf{(i) Accuracy} (\%) $\uparrow$} & \multicolumn{5}{c}{\textbf{(ii) Communication Overhead} (MB) $\downarrow$} \\
\cmidrule{2-6} \cmidrule{7-11}
& \textbf{Action} & \multirow{2}{*}{\textbf{UCI-HAR}} & \multirow{2}{*}{\textbf{PTB-XL}} & \multirow{2}{*}{\textbf{MELD}} & \multirow{2}{*}{\textbf{DFC23}} & \textbf{Action} & \multirow{2}{*}{\textbf{UCI-HAR}} & \multirow{2}{*}{\textbf{PTB-XL}} & \multirow{2}{*}{\textbf{MELD}} & \multirow{2}{*}{\textbf{DFC23}} \\
& \textbf{Sense} & & & & & \textbf{Sense} & & & & \\
\midrule
& \multicolumn{10}{c}{\cellcolor{gray!15} \textbf{IID Setting}} \\
\cmidrule{2-11}
FL-FD~\cite{qi2023fl} & 39.50 & 35.89 & 22.28 & 50.30 & 62.49 & 48.71 & 50.13 & N/A & 16.89 & 4.54 \\
MMFed~\cite{xiong2022unified} & 44.04 & 21.34 & 13.31 & 51.85 & 61.81 & 82.34 & 152.37 & N/A & 18.09 & 9.04 \\
FedMultimodal~\cite{feng2023fedmultimodal} & 13.45 & 17.52 & 4.07 & 49.39 & 60.24 & 135.08 & 143.70 & N/A & 271.32 & 9.04 \\
FLASH~\cite{salehi2022flash} & 21.41 & 22.88 & 17.44 & 49.10 & 61.67 & N/A & N/A & N/A & N/A & 9.04 \\
Harmony~\cite{ouyang2023harmony} & 42.02 & 21.61 & 15.84 & 45.26 & 57.83 & 110.24 & 145.22 & N/A & N/A & 18.04 \\
\midrule
Ours w/o Modality Sel & 68.73 & 76.92 & 86.50 & 54.89 & 61.16 & 5.99 & \textbf{5.78} & \textbf{0.24} & 2.51 & 9.21 \\
Ours w/o Client Sel. & 81.79 & 70.90 & 85.66 & 55.58 & 63.81 & N/A & 12.96 & \textbf{0.24} & \textbf{0.37} & 4.18 \\
Ours w/o Joint Sel. & 68.00 & 71.36 & 86.90 & 55.44 & \textbf{64.36} & 15.70 & 9.11 & \textbf{0.24} & 2.51 & \textbf{0.84} \\
\midrule
& & & & & & & & & & \\
\multirow{-2}{*}{\textbf{MFedMC (Ours) \textsuperscript{1}}} & \multirow{-2}{*}{\textbf{92.28} \textsuperscript{2}} & \multirow{-2}{*}{\textbf{78.10}} & \multirow{-2}{*}{\textbf{87.04}} & \multirow{-2}{*}{\textbf{55.82}} & \multirow{-2}{*}{62.05} & \multirow{-2}{*}{\textbf{4.27}} & \multirow{-2}{*}{7.18} & \multirow{-2}{*}{\textbf{0.24}} & \multirow{-2}{*}{\textbf{0.37}} & \multirow{-2}{*}{9.20} \\
\midrule
& \multicolumn{10}{c}{\cellcolor{gray!15} \textbf{Natural Distribution}} \\
\cmidrule{2-11}
FL-FD~\cite{qi2023fl} & 51.57 & 35.89 & 22.28 & 51.73 & 62.49 & 48.71 & 50.13 & N/A & 16.89 & 4.54 \\
MMFed~\cite{xiong2022unified} & 47.24 & 21.34 & 13.97 & 51.85 & 61.81 & 82.34 & 152.37 & N/A & 18.09 & 9.04 \\
FedMultimodal~\cite{feng2023fedmultimodal} & 24.81 & 17.70 & 16.40 & 49.39 & 62.40 & 135.08 & 143.70 & N/A & 271.32 & 9.04 \\
FLASH~\cite{salehi2022flash} & 34.77 & 22.88 & 17.44 & 49.10 & 62.13 & N/A & N/A & N/A & N/A & 9.04 \\
Harmony~\cite{ouyang2023harmony} & 50.72 & 21.61 & 11.41 & 46.33 & 63.17 & 110.24 & 145.22 & N/A & N/A & 18.04 \\
\midrule
Ours w/o Modality Sel. & 88.85 & 71.12 & 49.80 & 48.35 & 66.47 & 3.90 & 8.93 & N/A & N/A & \textbf{0.84} \\
Ours w/o Client Sel. & 90.39 & 60.54 & 51.44 & 53.67 & 67.41 & 2.94 & 13.49 & \textbf{0.24} & \textbf{0.37} & \textbf{0.84} \\
Ours w/o Joint Sel. & 89.30 & 60.22 & 54.18 & \textbf{57.43} & 66.69 & 3.86 & 16.64 & \textbf{0.24} & 1.65 & \textbf{0.84} \\
\midrule
& & & & & & & & & & \\
\multirow{-2}{*}{\textbf{MFedMC (Ours)}} & \multirow{-2}{*}{\textbf{98.87}} & \multirow{-2}{*}{\textbf{71.28}} & \multirow{-2}{*}{\textbf{55.09}} & \multirow{-2}{*}{53.31} & \multirow{-2}{*}{\textbf{67.61}} & \multirow{-2}{*}{\textbf{1.01}} & \multirow{-2}{*}{\textbf{7.36}} & \multirow{-2}{*}{0.48} & \multirow{-2}{*}{\textbf{0.37}} & \multirow{-2}{*}{\textbf{0.84}} \\
\bottomrule
\end{tabular}
}

\vspace{2pt}
\raggedright

\noindent{
\textsuperscript{1} \textbf{MFedMC (Ours)} utilize a consistent configuration with $\delta=0.2$, $\gamma=1$, $\alpha_s=1/3$, $\alpha_c=1/3$, $\alpha_r=1/3$ across all datasets. \quad 

\textsuperscript{2} \textbf{Bold} refers to the highest accuracy or the lowest communication overhead.

}
\end{table*}

\begin{figure}[t]
\centering
\includegraphics[width=0.99\linewidth]{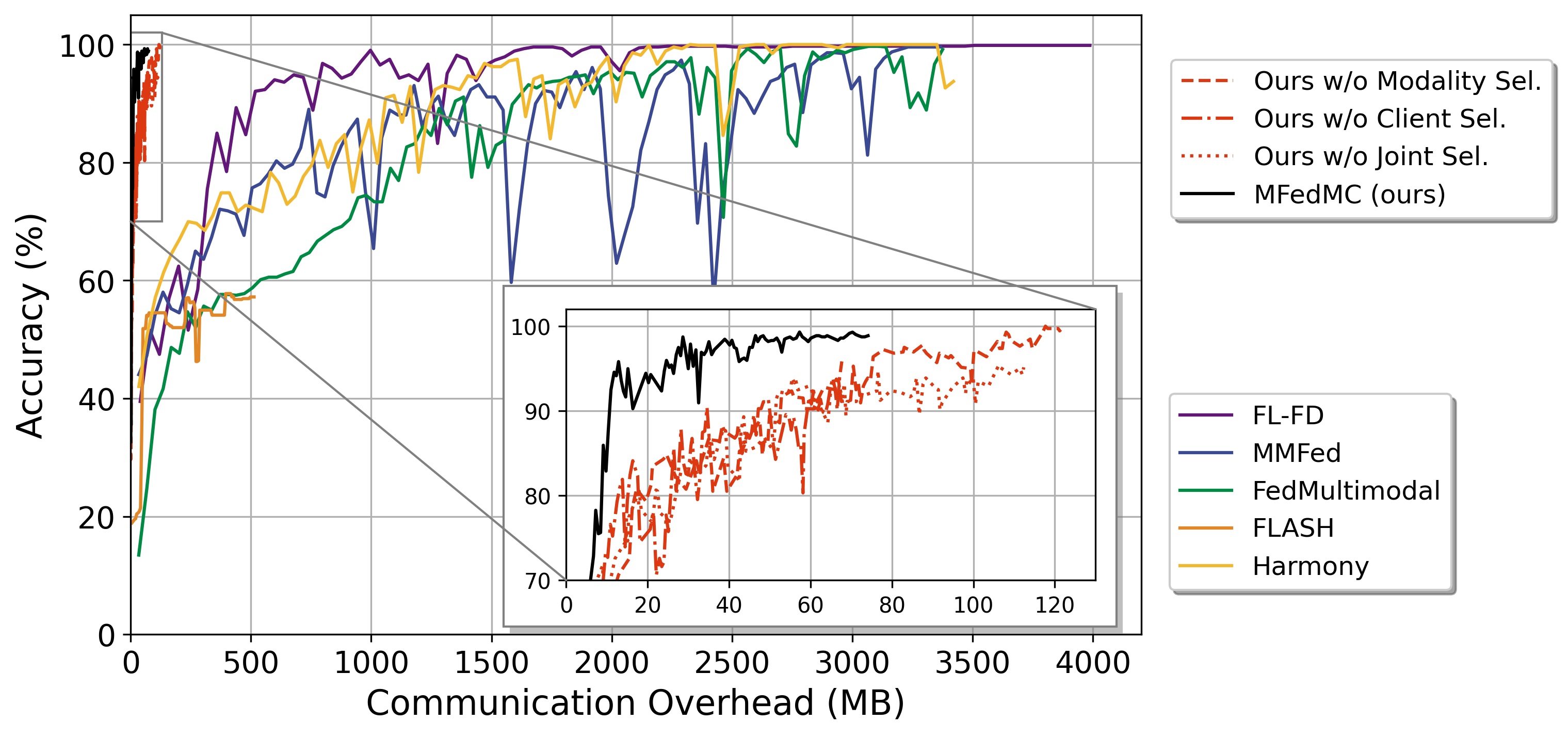}
\caption{Accuracy vs. communication overhead on ActionSense dataset under the natural distribution. MFedMC outperforms SOTA baselines with significantly lower communication overhead through decoupled architecture design and strategic joint selection.}
\label{Fig. Acc_vs_Comm}
\end{figure}

\subsection{Overall Comparison with Baselines}
\label{Sec. Results and Overall Comparison}

The performance obtained by the proposed MFedMC framework, in comparison with eight baselines on the ActionSense dataset, are presented in Table~\ref{Table Overall Comparison} and Fig.~\ref{Fig. Acc_vs_Comm}. 

\noindent\textbf{Comparison with SOTA Baselines.}
The proposed MFedMC, along with its three ablation variants employing random selection strategies, demonstrates substantial performance improvements over four SOTA approaches while significantly reducing communication costs. Notably, even the random selection variants achieve considerable performance gains despite requiring more aggregation iterations due to reduced per-round communication. This accuracy improvement is primarily attributed to our decoupled framework design: clients receive global modality encoders from the server and subsequently perform local fusion module updates. Through server aggregation, modality encoders acquire generalized knowledge from diverse client distributions, while the personalized fusion module fine-tunes locally to optimize multimodal integration based on these updated global encoders. Beyond accuracy improvements, MFedMC reduces communication overhead by nearly an order of magnitude through its selective upload strategy. While random submodel upload strategies~\cite{salehi2022flash} can effectively reduce communication costs to approximately $\frac{1}{M_k+1}$, they lack performance-aware and overhead-aware selection mechanisms, resulting in suboptimal performance-communication trade-offs. In contrast, MFedMC's design is guided by three key takeaways:
\begin{enumerate}[label=(\roman*)]
    \item Different modalities contribute heterogeneously to performance across clients and data distributions.
    \item Selective modality encoder aggregation accelerates convergence compared to aggregating all encoders.
    \item The decoupled architecture enables synergy between generalization (via global encoders) and personalization (via local fusion), which holistic baseline models cannot achieve.
\end{enumerate}

\begin{table}[t]
\caption{Effect of Modality Selection Weights on ActionSense: \\ (i) Accuracy under 25 MB Communication Constraint and \\ (ii) Communication Overhead to Achieve Target of 85\% Accuracy}
\label{Table ActionSense 25 MB}
\centering 
\vspace{-2mm}
\resizebox{\linewidth}{!}{
\begin{tabular}{@{}lcccc|c|c@{}}
\toprule
\multirow{2}{*}{\textbf{Method}} & \multirow{2}{*}{$\gamma$} & \multirow{2}{*}{$\alpha_s$} & \multirow{2}{*}{$\alpha_c$} & \multirow{2}{*}{$\alpha_r$} & \textbf{(i) Accuracy} & \textbf{(ii) Communication} \\
& & & & & (\%) $\uparrow$ & \textbf{Overhead} (MB) $\downarrow$ \\
\midrule
FL-FD~\cite{qi2023fl} & - & - & - & - & 62.43 & 48.71 \\
MMFed~\cite{xiong2022unified} & - & - & - & - & 59.28 & 82.34 \\
FedMultimodal~\cite{feng2023fedmultimodal} & - & - & - & - & 54.69 & 135.08 \\
FLASH~\cite{salehi2022flash} & - & - & - & - & 56.93 & N/A \\
Harmony~\cite{ouyang2023harmony} & - & - & - & - & 69.97 & 110.24 \\
\midrule
\multirow{21}{*}[-5pt]{\textbf{MFedMC ($\delta=1$)}}
& \multirow{7}{*}{1} & 1.0 & 0.0 & 0.0 & 85.14 & 19.54 \\
& & 0.0 & 1.0 & 0.0 & 90.91 & 16.86 \\
& & 0.0 & 0.0 & 1.0 & 86.34 & 17.85 \\
& & 0.0 & 0.5 & 0.5 & 88.95 & 12.23 \\
& & 0.5 & 0.0 & 0.5 & 97.92 & 8.30 \\
& & 0.5 & 0.5 & 0.0 & 98.05 & 11.62 \\
& & \textbf{1/3} & \textbf{1/3} & \textbf{1/3} & \textbf{99.58} & \textbf{3.01} \\
\cmidrule{2-7} 
& \multirow{7}{*}{2} & 1.0 & 0.0 & 0.0 & 88.06 & 20.17 \\
& & 0.0 & 1.0 & 0.0 & 97.36 & 21.33 \\
& & 0.0 & 0.0 & 1.0 & 87.76 & 14.39 \\
& & 0.0 & 0.5 & 0.5 & 86.45 & 22.54 \\
& & 0.5 & 0.0 & 0.5 & 91.94 & 14.00 \\
& & 0.5 & 0.5 & 0.0 & 97.92 & 13.38 \\
& & 0.3 & 0.3 & 0.3 & 97.08 & 4.39 \\
\cmidrule{2-7} 
& \multirow{7}{*}{3} & 1.0 & 0.0 & 0.0 & 93.45 & 17.65 \\
& & 0.0 & 1.0 & 0.0 & 88.97 & 25.62 \\
& & 0.0 & 0.0 & 1.0 & 87.91 & 17.85 \\
& & 0.0 & 0.5 & 0.5 & 87.28 & 17.55 \\
& & 0.5 & 0.0 & 0.5 & 91.66 & 12.27 \\
& & 0.5 & 0.5 & 0.0 & 94.44 & 5.83 \\
& & 0.3 & 0.3 & 0.3 & 94.71 & 14.18 \\
\bottomrule
\end{tabular}
}
\end{table}

\begin{table}[t]
\caption{Effect of Modality Selection Weights on UCI-HAR: \\ (i) Accuracy under 25 MB Communication Constraint and \\ (ii) Communication Overhead to Achieve Target of 60\% Accuracy}
\label{Table UCI-HAR 25 MB}
\centering 
\vspace{-2mm}
\resizebox{\linewidth}{!}{
\begin{tabular}{@{}lcccc|c|c@{}}
\toprule
\multirow{2}{*}{\textbf{Method}} & \multirow{2}{*}{$\gamma$} & \multirow{2}{*}{$\alpha_s$} & \multirow{2}{*}{$\alpha_c$} & \multirow{2}{*}{$\alpha_r$} & \textbf{(i) Accuracy} & \textbf{(ii) Communication} \\
& & & & & (\%) $\uparrow$ & \textbf{Overhead} (MB) $\downarrow$ \\
\midrule
FL-FD~\cite{qi2023fl} & - & - & - & - & 71.90 & 50.13 \\
MMFed~\cite{xiong2022unified} & - & - & - & - & 43.85 & 152.37 \\
FedMultimodal~\cite{feng2023fedmultimodal} & - & - & - & - & 37.36 & 143.70 \\
FLASH~\cite{salehi2022flash} & - & - & - & - & 48.54 & N/A \\
Harmony~\cite{ouyang2023harmony} & - & - & - & - & 45.52 & 145.22 \\
\midrule
\multirow{14}{*}[-2.5pt]{\textbf{MFedMC ($\delta=1$)}}
& \multirow{7}{*}{\textbf{1}} & 1.0 & 0.0 & 0.0 & 74.51 & \textbf{35.90} \\
& & 0.0 & 1.0 & 0.0 & 37.61 & N/A \\
& & 0.0 & 0.0 & 1.0 & 56.97 & 87.57 \\
& & 0.0 & 0.5 & 0.5 & 57.49 & 93.70 \\
& & 0.5 & 0.0 & 0.5 & 74.22 & 41.16 \\
& & \textbf{0.5} & \textbf{0.5} & \textbf{0.0} & \textbf{75.58} & 39.41 \\
& & 0.3 & 0.3 & 0.3 & 75.57 & 37.66 \\
\cmidrule{2-7} 
& \multirow{7}{*}{2} & 1.0 & 0.0 & 0.0 & 53.81 & 103.34 \\
& & 0.0 & 1.0 & 0.0 & 54.53 & 99.83 \\
& & 0.0 & 0.0 & 1.0 & 54.36 & 92.83 \\
& & 0.0 & 0.5 & 0.5 & 54.34 & 96.33 \\
& & 0.5 & 0.0 & 0.5 & 56.93 & N/A \\
& & 0.5 & 0.5 & 0.0 & 50.55 & 96.33 \\
& & 0.3 & 0.3 & 0.3 & 43.29 & N/A \\
\bottomrule
\end{tabular}
}
\end{table}

\noindent\textbf{Comparison with Ablation Baselines.} Compared to the three ablation baselines implemented through random selection, the proposed MFedMC generally achieves a similar or even lower communication overhead, while outperforming in most scenarios. This is attributed to the selective modality and client communication strategy, where not only does the lower communication cost promote more frequent aggregation, but also the Shapley value and client selection based on local loss can identify the clients with a greater impact in each modality. Due to the inherent randomness, random modality and client selections in some scenarios might marginally outperform MFedMC. This is because randomness, to some extent, ensures that all modalities for all clients have the same probability of being selected. However, this is not the optimal solution as excessive randomness and instability are not preferable.

\noindent\textbf{Comparison between Natural Distribution and IID.} 
The results indicate that IID settings do not always provide clear advantages in model generalization. Both IID and natural distribution settings present unique challenges due to client heterogeneity. In natural distribution scenarios, each client's training and testing data typically share similar yet biased distributions. For example, a hospital in Asia is unlikely to receive patients from North America during deployment. In such cases, the personalized local fusion module adapts effectively to region-specific patterns (e.g., Asian patient characteristics) by fine-tuning based on global modality encoders. Conversely, in IID settings where data is uniformly distributed across clients, the fusion module's personalization capability is underutilized, as local data does not exhibit distinctive patterns requiring specialized adaptation.Furthermore, natural distributions often follow skewed sample distributions. In the PTB-XL dataset, three hospital sites hold 93.54\% of the data, with 34 out of 39 sites having fewer than 100 samples each. Similarly, in MELD, six speakers possess 92.68\% of the data, with 36 out of 42 speakers having fewer than 100 samples. Clients with limited samples are prone to overfitting, resulting in artificially low local losses that can mislead our client selection strategy. We further investigate the effect of long-tail distribution in Section~\ref{Sec. Effect of Long-Tail Distribution}.

\begin{figure}[t]
\centering
\subfloat[ActionSense Dataset with Six Modalities\label{Fig. Impact ActionSense}]{\includegraphics[width=1\linewidth]{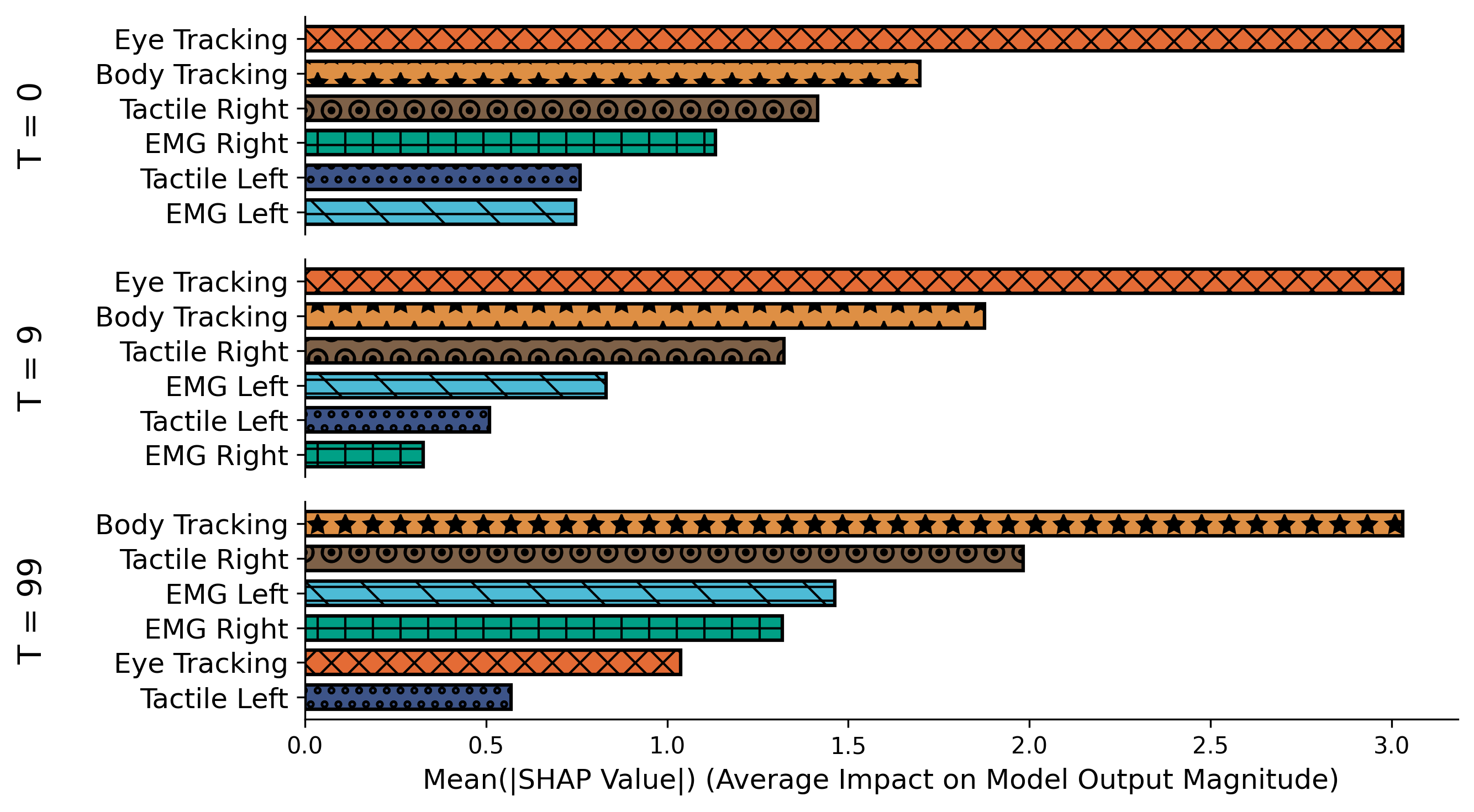}}\\
\subfloat[UCI-HAR with Two Modalities\label{Fig. Impact UCI-HAR}]{\includegraphics[width=1\linewidth]{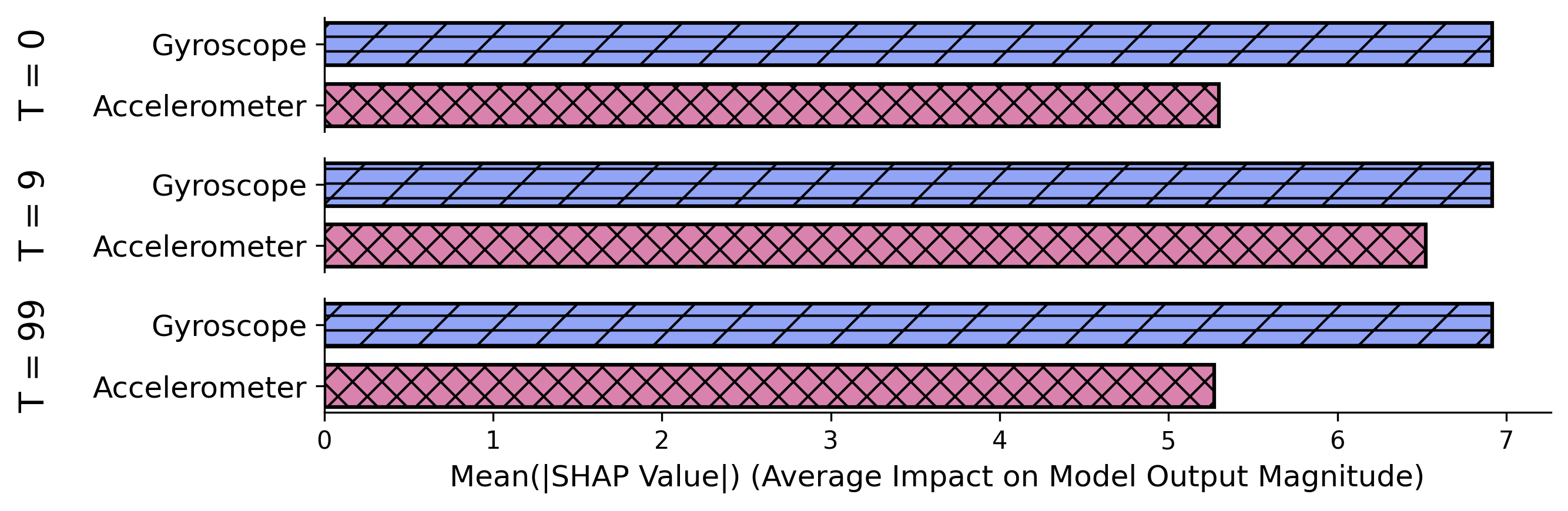}}
\caption{The mean Shapley value (i.e., impact) of modality encoders throughout the MFedMC iteration. MFedMC demonstrates adaptive modality prioritization: (a) on ActionSense, contributions shift from Eye Tracking (initially dominant due to efficiency) to Body Tracking (information-rich as encoders mature), and (b) on UCI-HAR, Gyroscope exhibits higher impact for capturing rotational movements in activity recognition.}
\label{Fig. Impact}
\end{figure}

\begin{table*}[t]
\caption{Effect of Client Selection on ActionSense under 5 MB Communication Constraint}
\label{Table ActionSense 5 MB}
\begin{minipage}[t]{0.80\linewidth}
\vspace{-6mm}
\centering 
\resizebox{0.95\linewidth}{!}{
\begin{tabular}{@{}lcccc||ccc||ccc@{}}
\toprule
& & & & & \multicolumn{3}{c||}{\textbf{Higher Loss}} & \multicolumn{3}{c}{\textbf{Lower Loss}} \\
\cmidrule{6-11}
\multirow{2}{*}{$\delta$} & \multirow{2}{*}{$\gamma$} & \multirow{2}{*}{$\alpha_s$} & \multirow{2}{*}{$\alpha_c$} & \multirow{2}{*}{$\alpha_r$} & \textbf{Accuracy} & \textbf{Communication} & \textbf{Comm.} &  \textbf{Accuracy} & \textbf{Communication} & \textbf{Comm.} \\
& & & & & (\%) $\uparrow$ & \textbf{Overhead} (MB) $\downarrow$ & \textbf{Round} & (\%) $\uparrow$ & \textbf{Overhead} (MB) $\downarrow$ & \textbf{Round} \\
\midrule

\multirow{21}{*}{0.2} & \multirow{7}{*}{1} & 1.0 & 0.0 & 0.0 & \cellcolor{StrongUp}81.97(+29.16\%) & \cellcolor{StrongUp}0.05(-80.77\%) & 99(+80) & \cellcolor{StrongUp}98.30(+54.88\%) & \cellcolor{StrongUp}0.09(-65.48\%) & 57(+38) \\
& & 0.0 & 1.0 & 0.0 & \cellcolor{Down}74.04(-10.00\%) & 0.06(0.00\%) & 84(0) & \cellcolor{StrongUp}98.87(+20.20\%) & 0.06(0.00\%) & 84(0) \\
& & 0.0 & 0.0 & 1.0 & \cellcolor{Up}88.17(+3.95\%) & 0.13(0.00\%) & 38(0) & \cellcolor{Down}84.68(-0.17\%) & 0.13(0.00\%) & 38(0) \\
& & 0.0 & 0.5 & 0.5 & \cellcolor{Down}85.30(-2.90\%) & 0.07(0.00\%) & 76(0) & \cellcolor{Up}88.59(+0.84\%) & 0.07(0.00\%) & 76(0) \\
& & 0.5 & 0.0 & 0.5 & \cellcolor{Down}87.59(-6.27\%) & \cellcolor{Down}0.05(+1.37\%) & 92(0) & \cellcolor{Up}98.87(+5.80\%) & \cellcolor{StrongDown}0.06(+12.25\%) & 83(-9) \\
& & 0.5 & 0.5 & 0.0 & \cellcolor{Up}90.62(+5.70\%) & \cellcolor{Down}0.08(+9.40\%) & 61(-5) & \cellcolor{StrongUp}92.88(+34.04\%) & \cellcolor{StrongUp}0.14(-64.29\%) & 35(+23) \\
& & 1/3 & 1/3 & 1/3 & \cellcolor{Down}85.06(-8.82\%) & \cellcolor{Up}0.07(-9.48\%) & 76(+7) & \cellcolor{StrongUp}98.87(+14.52\%) & \cellcolor{StrongUp}0.06(-15.66\%) & 79(+13) \\
\cmidrule{2-11}
& \multirow{7}{*}{2} & 1.0 & 0.0 & 0.0 & \cellcolor{StrongUp}82.24(+18.69\%) & \cellcolor{StrongUp}0.09(-75.98\%) & 53(+41) & \cellcolor{Up}98.89(+6.00\%) & \cellcolor{Up}0.06(-9.98\%) & 77(+8) \\
& & 0.0 & 1.0 & 0.0 & \cellcolor{Down}82.50(-7.70\%) & 0.12(0.00\%) & 41(0) & \cellcolor{Up}94.55(+5.77\%) & 0.12(0.00\%) & 41(0) \\
& & 0.0 & 0.0 & 1.0 & \cellcolor{Down}86.39(-0.95\%) & 0.26(0.00\%) & 19(0) & \cellcolor{Down}86.11(-1.27\%) & 0.26(0.00\%) & 19(0) \\
& & 0.0 & 0.5 & 0.5 & \cellcolor{Down}82.55(-5.27\%) & 0.13(0.00\%) & 37(0) & \cellcolor{Up}89.66(+2.89\%) & 0.13(0.00\%) & 37(0) \\
& & 0.5 & 0.0 & 0.5 & \cellcolor{Down}87.87(-2.03\%) & \cellcolor{StrongUp}0.13(-19.20\%) & 38(+7) & \cellcolor{Up}91.48(+2.01\%) & \cellcolor{Down}0.16(+1.38\%) & 30(-1) \\
& & 0.5 & 0.5 & 0.0 & \cellcolor{Up}86.17(+0.16\%) & \cellcolor{Down}0.14(+0.68\%) & 34(-1) & \cellcolor{StrongUp}98.60(+14.61\%) & \cellcolor{Down}0.14(+0.59\%) & 35(0) \\
& & 1/3 & 1/3 & 1/3 & \cellcolor{Down}86.47(-2.82\%) & \cellcolor{Up}0.13(-1.71\%) & 38(+1) & \cellcolor{Up}97.05(+9.07\%) & \cellcolor{Down}0.14(+5.67\%) & 35(-2) \\
\cmidrule{2-11}
& \multirow{7}{*}{3} & 1.0 & 0.0 & 0.0 & \cellcolor{StrongUp}86.75(+18.81\%) & \cellcolor{StrongUp}0.17(-68.80\%) & 30(+21) & \cellcolor{StrongUp}89.25(+22.24\%) & \cellcolor{StrongUp}0.21(-61.33\%) & 24(+15) \\
& & 0.0 & 1.0 & 0.0 & \cellcolor{Down}79.59(-7.19\%) & 0.18(0.00\%) & 27(0) & \cellcolor{Up}93.30(+8.80\%) & 0.18(0.00\%) & 27(0) \\
& & 0.0 & 0.0 & 1.0 & \cellcolor{Up}86.77(+5.03\%) & 0.39(0.00\%) & 12(0) & \cellcolor{Up}86.67(+4.90\%) & 0.39(0.00\%) & 12(0) \\
& & 0.0 & 0.5 & 0.5 & \cellcolor{Down}83.65(-7.01\%) & 0.19(0.00\%) & 25(0) & \cellcolor{Up}91.21(+1.41\%) & 0.19(0.00\%) & 25(0) \\
& & 0.5 & 0.0 & 0.5 & \cellcolor{Down}89.55(-2.39\%) & \cellcolor{Up}0.21(-2.11\%) & 24(+1) & \cellcolor{Down}90.20(-1.68\%) & \cellcolor{StrongDown}0.26(+20.91\%) & 19(-4) \\
& & 0.5 & 0.5 & 0.0 & \cellcolor{Down}81.83(-4.73\%) & \cellcolor{Up}0.18(-9.07\%) & 27(+3) & \cellcolor{Up}93.00(+8.28\%) & \cellcolor{Down}0.21(+2.48\%) & 24(0) \\
& & 1/3 & 1/3 & 1/3 & \cellcolor{Down}81.97(-7.74\%) & \cellcolor{Up}0.19(-2.31\%) & 26(0) & \cellcolor{Up}93.44(+5.16\%) & \cellcolor{Down}0.21(+8.09\%) & 24(-2) \\  
\bottomrule
\end{tabular}
}
\end{minipage}
\begin{minipage}[t]{0.1\linewidth}

\renewcommand{\arraystretch}{1.2}
\setlength{\aboverulesep}{0pt}
\setlength{\belowrulesep}{0pt}
\vspace{-6mm}
\scriptsize
\centering 
\begin{tabular}{!{\vrule width 0.25ex}c!{\vrule width 0.25ex}}
\toprule
Improvement over Baseline \\
(Random Client Selection) \\
\toprule
\cellcolor{StrongUp}$\geq$ 10\% \\
\cellcolor{Up}$0\%$ to 10\% \\
\cellcolor{Down}$-10\%$ to 0\% \\
\cellcolor{StrongDown}$\leq$ -10\% \\
\toprule
\end{tabular}
\end{minipage}
\end{table*}

\begin{table*}[t]

\caption{Effect of Client Selection on UCI-HAR under 5 MB Communication Constraint}
\label{Table UCI-HAR 5 MB}
\begin{minipage}[t]{0.80\linewidth}
\vspace{-6mm}
\centering 
\resizebox{0.95\linewidth}{!}{
\begin{tabular}{@{}lcccc||ccc||ccc@{}}
\toprule
& & & & & \multicolumn{3}{c||}{\textbf{Higher Loss}} & \multicolumn{3}{c}{\textbf{Lower Loss}} \\
\cmidrule{6-11}
\multirow{2}{*}{$\delta$} & \multirow{2}{*}{$\gamma$} & \multirow{2}{*}{$\alpha_s$} & \multirow{2}{*}{$\alpha_c$} & \multirow{2}{*}{$\alpha_r$} & \textbf{Accuracy} & \textbf{Communication} & \textbf{Comm.} &  \textbf{Accuracy} & \textbf{Communication} & \textbf{Comm.} \\
& & & & & (\%) $\uparrow$ & \textbf{Overhead} (MB) $\downarrow$ & \textbf{Round} & (\%) $\uparrow$ & \textbf{Overhead} (MB) $\downarrow$ & \textbf{Round} \\
\midrule
\multirow{14}{*}{0.2} & \multirow{7}{*}{1} & 1.0 & 0.0 & 0.0 & \cellcolor{StrongDown}47.78(-22.79\%) & 0.05(0.00\%) & 95(0) & \cellcolor{StrongUp}68.77(+11.13\%) & 0.05(0.00\%) & 95(0) \\
& & 0.0 & 1.0 & 0.0 & \cellcolor{StrongDown}33.80(-39.88\%) & 0.05(0.00\%) & 95(0) & \cellcolor{Down}55.77(-0.80\%) & 0.05(0.00\%) & 95(0) \\
& & 0.0 & 0.0 & 1.0 & \cellcolor{StrongDown}40.31(-18.33\%) & 0.05(0.00\%) & 95(0) & \cellcolor{StrongUp}68.27(+38.30\%) & 0.05(0.00\%) & 95(0) \\
& & 0.0 & 0.5 & 0.5 & \cellcolor{StrongDown}37.29(-23.51\%) & 0.05(0.00\%) & 95(0) & \cellcolor{StrongUp}67.60(+38.67\%) & 0.05(0.00\%) & 95(0) \\
& & 0.5 & 0.0 & 0.5 & \cellcolor{StrongDown}42.39(-34.28\%) & 0.05(0.00\%) & 95(0) & \cellcolor{Up}70.62(+9.50\%) & 0.05(0.00\%) & 95(0) \\
& & 0.5 & 0.5 & 0.0 & \cellcolor{StrongDown}47.70(-14.81\%) & 0.05(0.00\%) & 95(0) & \cellcolor{StrongUp}65.16(+16.36\%) & 0.05(0.00\%) & 95(0) \\
& & 1/3 & 1/3 & 1/3 & \cellcolor{StrongDown}39.74(-24.53\%) & 0.05(0.00\%) & 95(0) & \cellcolor{StrongUp}71.28(+35.37\%) & 0.05(0.00\%) & 95(0) \\
\cmidrule{2-11}
& \multirow{7}{*}{2} & 1.0 & 0.0 & 0.0 & \cellcolor{StrongDown}36.74(-22.25\%) & 0.11(0.00\%) & 47(0) & \cellcolor{StrongUp}65.77(+39.19\%) & 0.11(0.00\%) & 47(0) \\
& & 0.0 & 1.0 & 0.0 & \cellcolor{StrongDown}35.43(-29.30\%) & 0.11(0.00\%) & 47(0) & \cellcolor{StrongUp}67.37(+34.43\%) & 0.11(0.00\%) & 47(0) \\
& & 0.0 & 0.0 & 1.0 & \cellcolor{StrongDown}41.61(-15.11\%) & 0.11(0.00\%) & 47(0) & \cellcolor{StrongUp}68.22(+39.17\%) & 0.11(0.00\%) & 47(0) \\
& & 0.0 & 0.5 & 0.5 & \cellcolor{StrongDown}37.24(-27.29\%) & 0.11(0.00\%) & 47(0) & \cellcolor{StrongUp}65.35(+27.58\%) & 0.11(0.00\%) & 47(0) \\
& & 0.5 & 0.0 & 0.5 & \cellcolor{StrongDown}39.52(-23.88\%) & 0.11(0.00\%) & 47(0) & \cellcolor{StrongUp}66.99(+29.02\%) & 0.11(0.00\%) & 47(0) \\
& & 0.5 & 0.5 & 0.0 & \cellcolor{StrongDown}37.60(-20.59\%) & 0.11(0.00\%) & 47(0) & \cellcolor{StrongUp}69.74(+47.28\%) & 0.11(0.00\%) & 47(0) \\
& & 1/3 & 1/3 & 1/3 & \cellcolor{StrongDown}36.02(-26.67\%) & 0.11(0.00\%) & 47(0) & \cellcolor{StrongUp}67.83(+38.10\%) & 0.11(0.00\%) & 47(0) \\
\bottomrule
\end{tabular}
}
\end{minipage}
\begin{minipage}[t]{0.1\linewidth}

\renewcommand{\arraystretch}{1.2}
\setlength{\aboverulesep}{0pt}
\setlength{\belowrulesep}{0pt}
\vspace{-6mm}
\scriptsize
\centering 
\begin{tabular}{!{\vrule width 0.25ex}c!{\vrule width 0.25ex}}
\toprule
Improvement over Baseline \\
(Random Client Selection) \\
\toprule
\cellcolor{StrongUp}$\geq$ 10\% \\
\cellcolor{Up}$0\%$ to 10\% \\
\cellcolor{Down}$-10\%$ to 0\% \\
\cellcolor{StrongDown}$\leq$ -10\% \\
\toprule
\end{tabular}
\end{minipage}
\end{table*}

\subsection{Ablation Study: Modality Selection} 
\label{Sec. Ablation Study: Modality Selection} 

\subsubsection{Trade-Off Analysis}
\label{Sec. Trade-Off Analysis}

Tables~\ref{Table ActionSense 25 MB} and~\ref{Table UCI-HAR 25 MB} present ablation results on ActionSense and UCI-HAR without client selection, examining the impact of modality selection parameters $\gamma$, $\alpha_s$, $\alpha_c$, and $\alpha_r$ on performance and communication overhead. Optimal performance requires balancing modality informativeness ($\alpha_s$) and communication cost ($\alpha_c$). Increasing $\gamma$ does not always improve accuracy and can substantially increase communication overhead. Selectively uploading only the most informative modalities often yields better performance than uploading all encoders. However, relying solely on Shapley value can lead to a \textit{single modality optimization trap}: if a compact modality contains rich information, it achieves high Shapley values and is repeatedly selected. Server aggregation further refines this encoder, creating a feedback loop where the framework focuses exclusively on one modality while neglecting others. The recency term ($\alpha_r$) mitigates this by introducing temporal diversity, encouraging the framework to explore different modalities over time rather than prematurely converging to a single dominant modality.

The effectiveness of these selection criteria depends on encoder architecture. In UCI-HAR, both modalities have identical encoder sizes (0.26 MB), making communication overhead ineffective for differentiation. Moreover, with only two modalities, recency causes cyclic selection, diminishing Shapley-based prioritization. Conversely, MFedMC performs best on datasets like ActionSense with heterogeneous modality dimensions, where diverse encoder sizes enable nuanced selection that effectively balances informativeness, communication cost, and temporal diversity.

\subsubsection{Analytics on Modality Impact}
\label{Sec. Analytic on Modality Impact}

Beyond guiding modality selection, Shapley values provide an interpretable metric to quantify each modality's contribution to the fusion module's predictions. Fig. \ref{Fig. Impact} illustrates the dynamics of modality impact across communication rounds. In early MFL stages, modalities with simpler features exhibit higher impact as their encoders quickly capture discriminative patterns. As training progresses, the selection dynamics evolve based on the interplay between information richness, encoder complexity, and communication overhead. Modalities that balance informativeness with lower communication costs, such as Body Tracking in ActionSense, emerge as primary contributors, while complex modalities with marginal information gains become less favored despite their potential accuracy improvements.

Fig. \ref{Fig. Impact ActionSense} demonstrates dynamic modality selection on ActionSense. Initially, Eye Tracking dominates due to its simple structure and low communication overhead, enabling frequent uploads and rapid convergence. However, as encoders mature, its contribution diminishes as simpler modalities reach saturation. Body Tracking gradually becomes the primary contributor, offering rich information with moderate communication cost. Tactile Right maintains a secondary position despite high complexity due to its information richness, while Tactile Left remains disadvantaged, likely reflecting limited diversity in left-hand actions. This evolution reflects MFedMC's adaptive prioritization: favoring cost-effective modalities that provide substantial information gains. In contrast, Fig. \ref{Fig. Impact UCI-HAR} shows subtle changes on UCI-HAR, where both modalities have identical feature dimensions and encoder sizes. Modality selection depends primarily on Shapley values, with recency introducing cyclic patterns that diminish differentiation. The Gyroscope exhibits slightly higher impact than the Accelerometer, as it more effectively captures rotational movements and angular velocities critical for distinguishing activities like running versus climbing stairs.

\subsection{Ablation Study: Client Selection}

\subsubsection{Comparison of Loss-Based Selection Criteria}

Recall that our primary objective is to maximize model performance with limited communication resources. In MFL settings, particularly in the MFedMC framework after modality selection, client selection presents a unique challenge: how to compare the quality of different modality encoders across clients without additional assumptions (e.g., access to a public validation dataset or uploading all candidate encoders to the server). Local loss provides a practical solution by enabling cross-modality comparison: a lower loss for a modality encoder indicates more effective learning from local data and suggests higher reliability for global aggregation.

Tables~\ref{Table ActionSense 5 MB} and \ref{Table UCI-HAR 5 MB} compare different client selection strategies on the ActionSense and UCI-HAR datasets across various configurations. The results consistently demonstrate that selecting clients with lower local losses outperforms both random selection and selecting clients with higher losses. This finding contrasts with some single-modality FL approaches, such as~\cite{cho2020client}, which propose selecting higher-loss clients to prioritize difficult samples. However, such strategies' objectives fundamentally differ from ours: they prioritize exploration of difficult samples to improve model generalization, whereas our communication-constrained setting prioritizes fast convergence to minimize communication overhead. In our MFedMC framework, convergence speed directly determines the total communication cost required to reach target accuracy. Prioritizing well-trained encoders (i.e., lower-loss clients) ensures stable global aggregation by consistently aggregating high-quality model updates, prevents poorly-trained encoders from degrading the global model, and accelerates convergence to reduce the communication overhead needed.

\begin{figure}[t]
\centering
\captionsetup[subfloat]{font=scriptsize}
\subfloat[ActionSense - Higher Loss]{\includegraphics[width=0.5\linewidth]{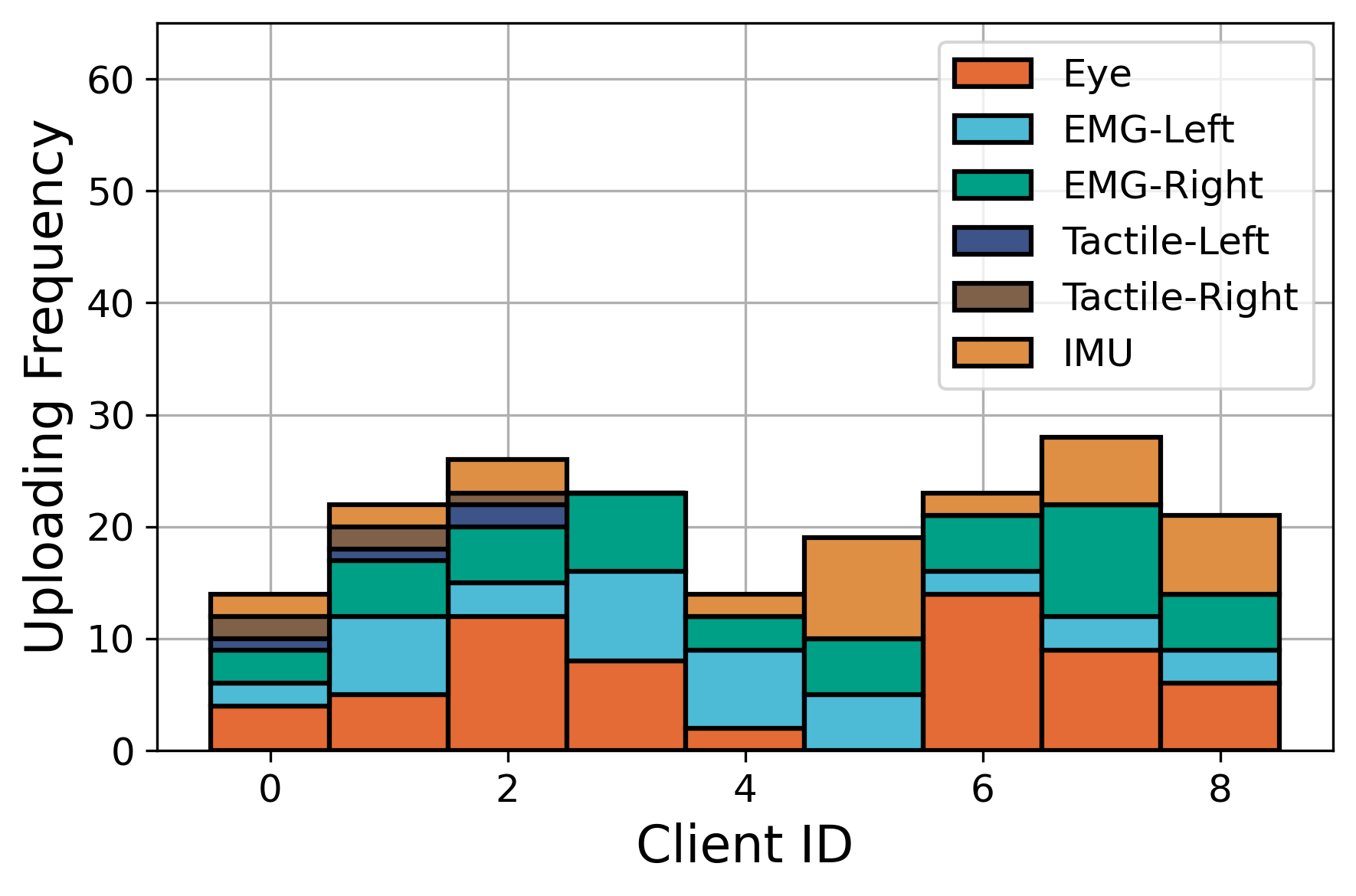}}
\subfloat[ActionSense - Lower Loss]{\includegraphics[width=0.5\linewidth]{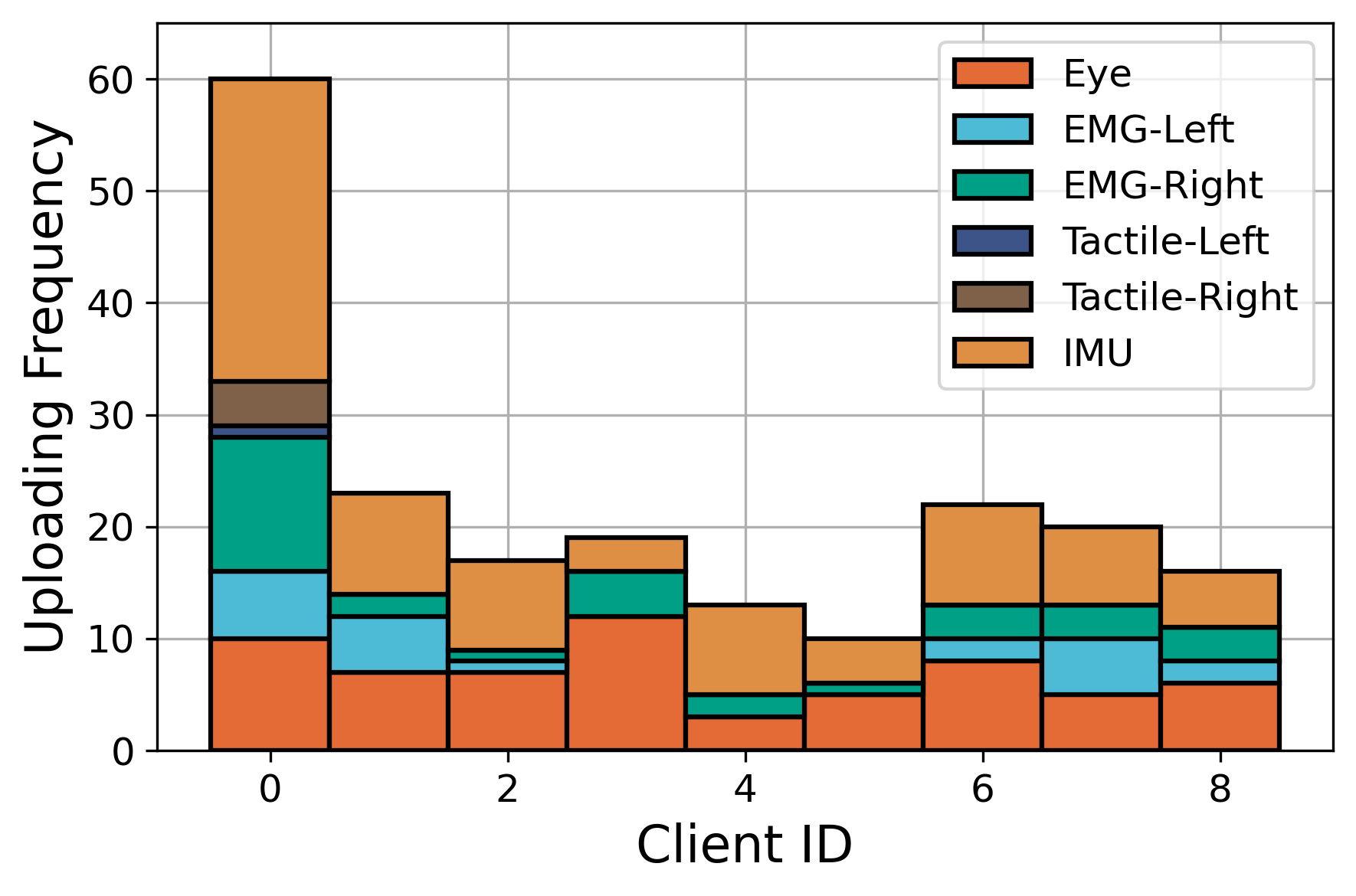}} \\
\subfloat[UCI-HAR - Higher Loss]{\includegraphics[width=0.5\linewidth]{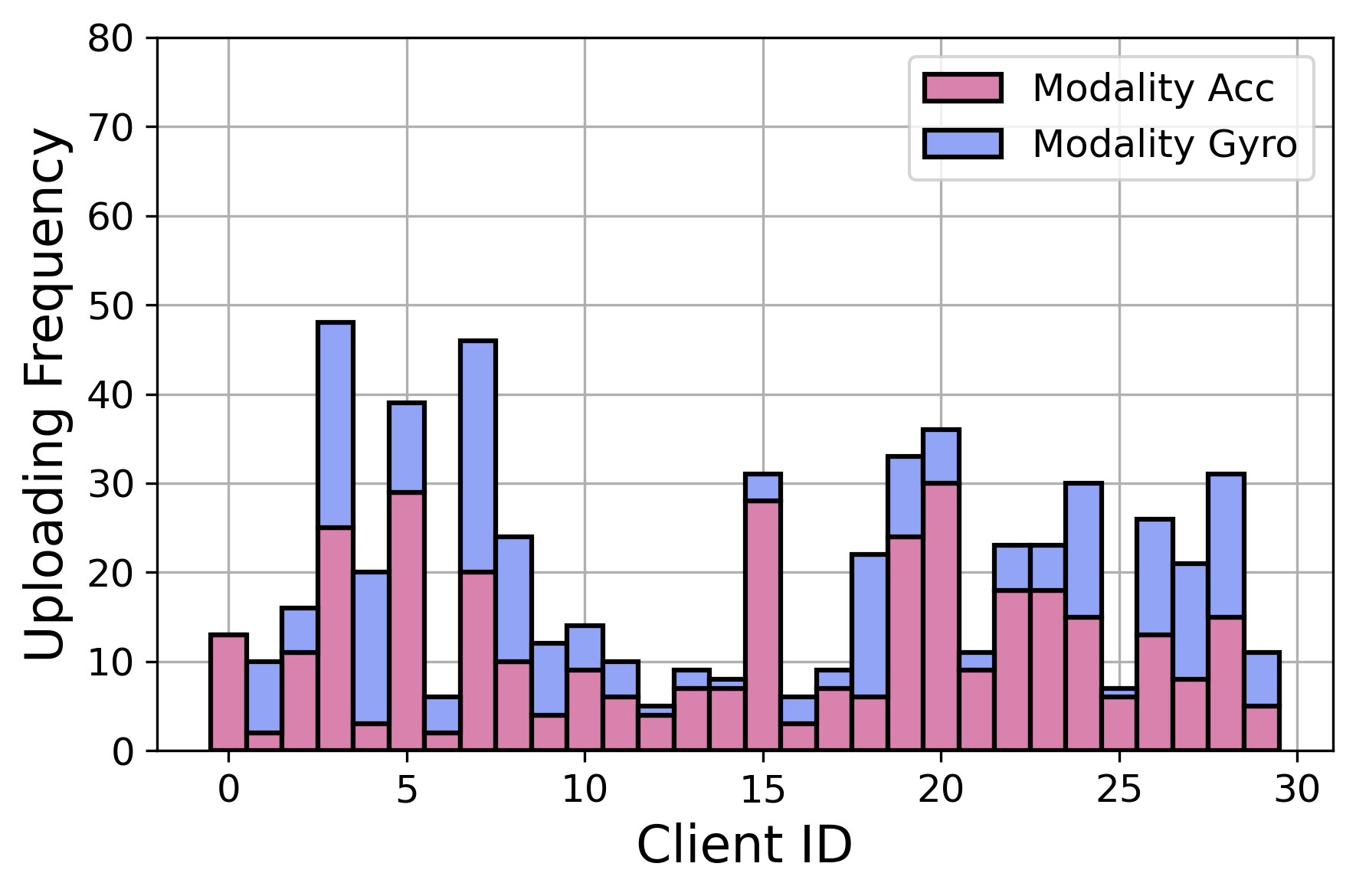}}
\subfloat[UCI-HAR - Lower Loss]{\includegraphics[width=0.5\linewidth]{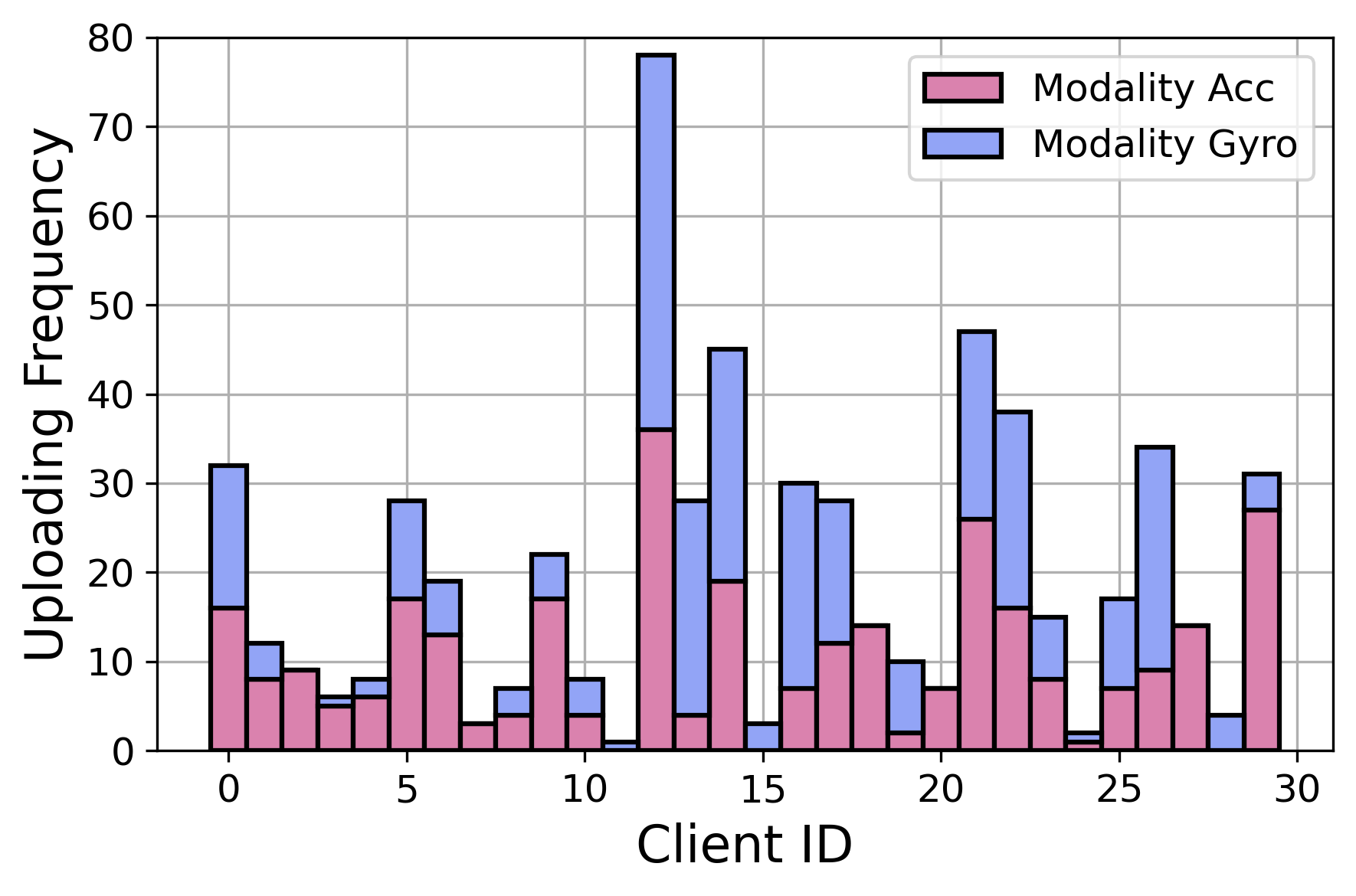}}
\caption{Histograms of the client selection frequency based on the criteria of higher and lower loss.}
\label{Fig. Histogram} 
\end{figure}

\subsubsection{Comparison of Client Selection Frequency}

Fig.~\ref{Fig. Histogram} illustrates the frequency distribution of client selection under the higher-loss and lower-loss strategies, revealing their distinct preferences for certain clients. The higher-loss strategy results in a more uniform distribution, where each client is selected with relatively equal frequency. In contrast, the lower-loss strategy exhibits a clear preference for a subset of clients, leading to a more skewed distribution. This difference stems from the nature of local loss, which reflects not only the information richness of client data but also factors such as client heterogeneity, model optimization dynamics, and data quality. Clients that deviate significantly from the majority due to heterogeneity, high data noise, erroneous labels, or even malicious attacks (e.g., data poisoning) tend to exhibit higher local losses. Consequently, while the lower-loss strategy may favor clients with more homogeneous data, the higher-loss strategy is more susceptible to selecting clients that deviate from the main client data distribution.

\subsection{Effect of Class and Modality non-IID Settings} 

\begin{figure*}[t]
\centering
\subfloat[Class Non-IID Setting \label{Fig. Class Non-IID}]{\includegraphics[height=4.8cm]{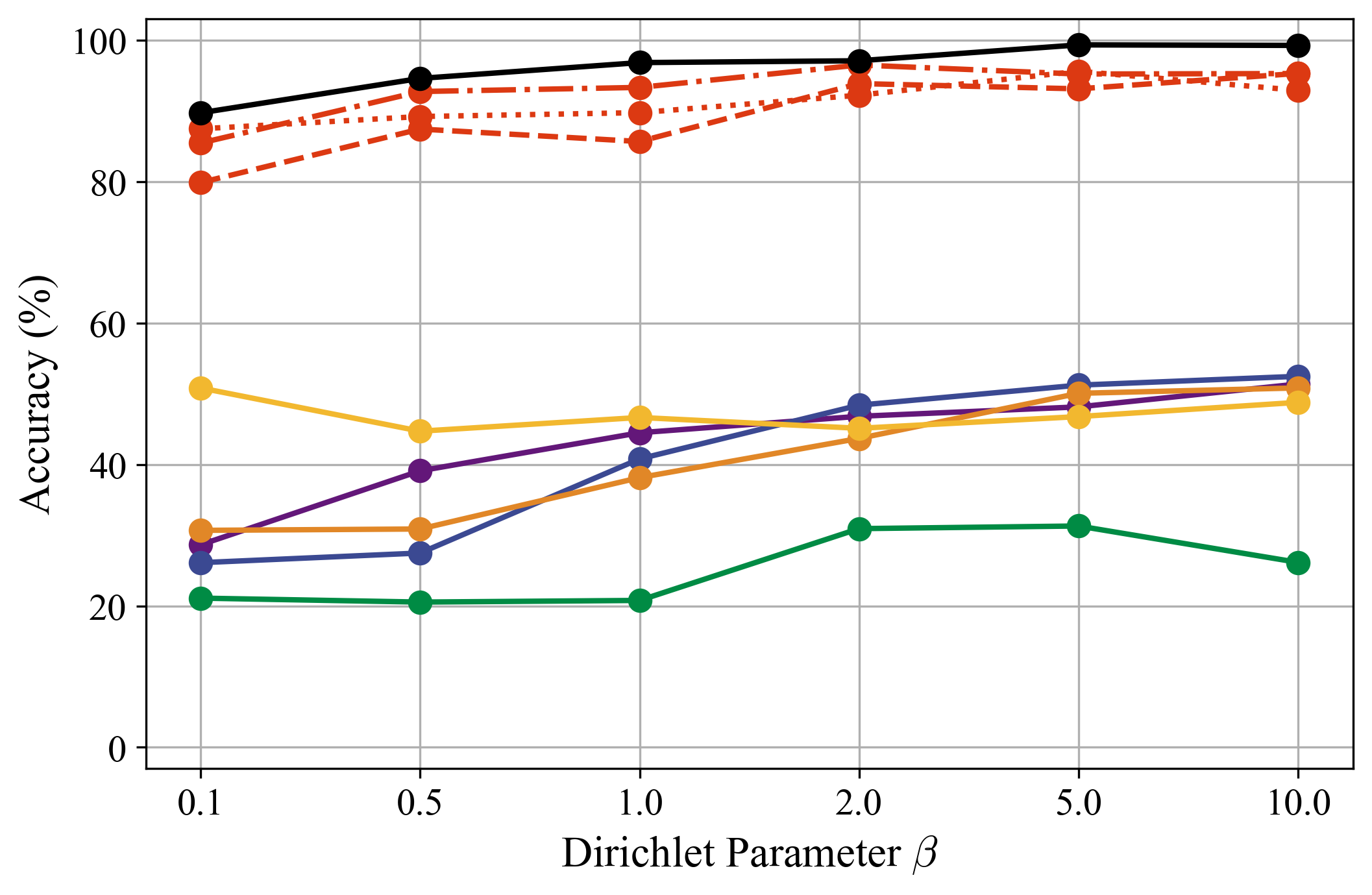}}
\subfloat[Modality Non-IID Setting \label{Fig. Modality Non-IID}]{\includegraphics[height=4.8cm]{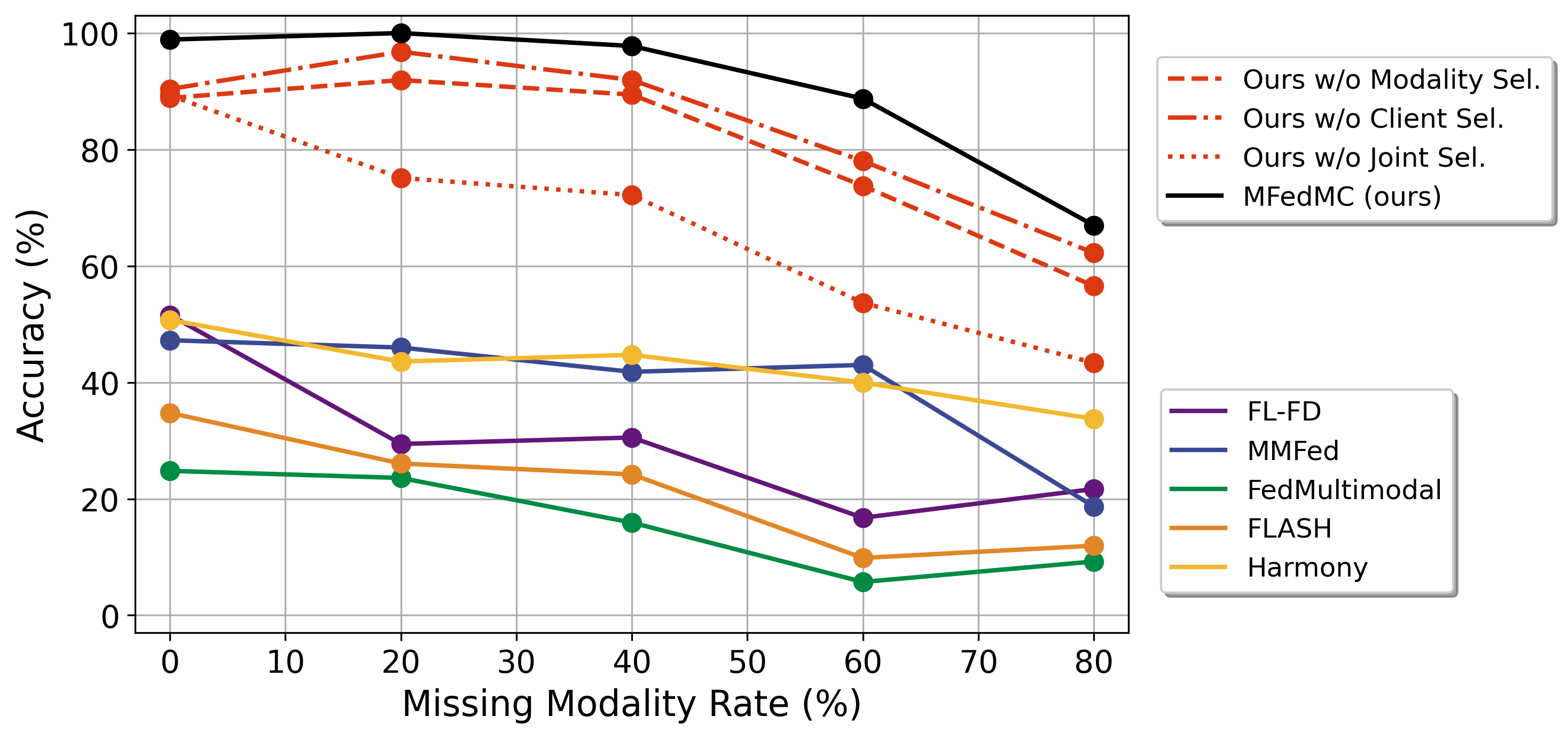}}
\caption{Effect of class and modality non-IID settings on ActionSense dataset. MFedMC demonstrates robust performance under both non-IID conditions: (a) maintaining superior accuracy across varying Dirichlet parameters through local fusion module personalization, and (b) outperforming baselines under extreme missing modality rates.}
\label{Fig. Effect of non-IID} 
\end{figure*}

\noindent\textbf{Class Non-IID Setting.} To simulate class non-IID setting, we redistribute client datasets through a Dirichlet distribution Dir($\beta$) in the ActionSense dataset, with results shown in Fig. \ref{Fig. Class Non-IID}. It can be observed that our method consistently outperforms all baselines at any $\beta$ value, and the impact of non-IIDness on our method is minimal, whether for our full approach or our ablation baselines with random selection. This advantage primarily stems from our strategy of separating modality encoders and fusion modules while keeping fusion modules local as personalization implementations. This optimizes fusion modules to learn local data distributions without interference from other clients' data distributions. The baseline methods lack such unique personalization configurations, resulting in slower convergence under non-IID conditions.

\noindent\textbf{Modality Non-IID Setting.} To simulate scenarios where clients do not possess all modalities (i.e., have missing modalities), we evaluate the performance of our system by randomly removing modalities from clients at certain probabilities, as illustrated in Fig. \ref{Fig. Modality Non-IID}. Thus, a missing modality rate of $0 \%$ corresponds to the case shown in Fig. \ref{Fig. Acc_vs_Comm}. A missing modality rate of $80 \%$ corresponds to the case where each client has at least two modalities (i.e., $M_k \geq 2$ for all clients $k$). Also, recall that even with the missing modality rate of $0 \%$, the ActionSense dataset inherently has missing modalities, with subjects 06 through 09 missing both left and right tactile data. We see that our MFedMC framework still demonstrates superior performance compared to baselines across missing modalities. Under extreme modality non-IID (missing modality rate of $80 \%$), the MFedMC framework inevitably experiences performance degradation due to the significant loss of information from the missing modalities. Interestingly, the performance of the MFedMC framework under extreme modality non-IID (missing modality rate of $80 \%$) still surpasses that of the baselines under no modality non-IID (missing modality rate of $0 \%$). This underscores the advantages provided by MFedMC: the fusion module enhances accuracy through personalization, while joint selection identifies the most informative modalities, optimizes client models, and reduces communication overhead.

\subsection{Effect of Heterogeneous Network} 

In practice, different clients possess varying available network bandwidth, inhibiting their ability to upload all model parameters. Through modality selection, MFedMC can adaptively select which modality encoders to transmit based on bandwidth limitations, ensuring efficient use of available communication channels and minimizing bottlenecks due to bandwidth constraints. To demonstrate this, we conduct experiments in a setting where clients have different uplink communication restrictions for uploading modality encoders to the server, due to heterogeneous transmit powers and uplink channels. Specifically, using the ActionSense dataset, we classify the clients into three levels based on their communication capabilities:
\begin{itemize}
    \item \textbf{Unrestricted Communication:} Clients 1-2 are able to upload all modality encoders.
    \item \textbf{Moderate Communication Restrictions:} Clients 3-5 are able to upload the Eye Tracking, EMG Left, EMG Right, and Body Tracking models.
    \item \textbf{Severe Communication Restrictions:} Clients 6-9 are limited to uploading the Eye Tracking, EMG Left, and EMG Right models due to their manageable sizes.
\end{itemize}
For the SOTA baselines, FLASH~\cite{salehi2022flash} and Harmony~\cite{ouyang2023harmony} allow for the segmentation of models, enabling each client to participate in the FL process. In contrast, the other three baselines, FL-FD~\cite{qi2023fl}, MMFed~\cite{xiong2022unified}, and FedMultimodal~\cite{feng2023fedmultimodal}, utilize end-to-end models, which means that under communication restrictions, only clients 1-2 are able to engage in FL.

\begin{figure}[t]
\centering
\includegraphics[width=0.99\linewidth]{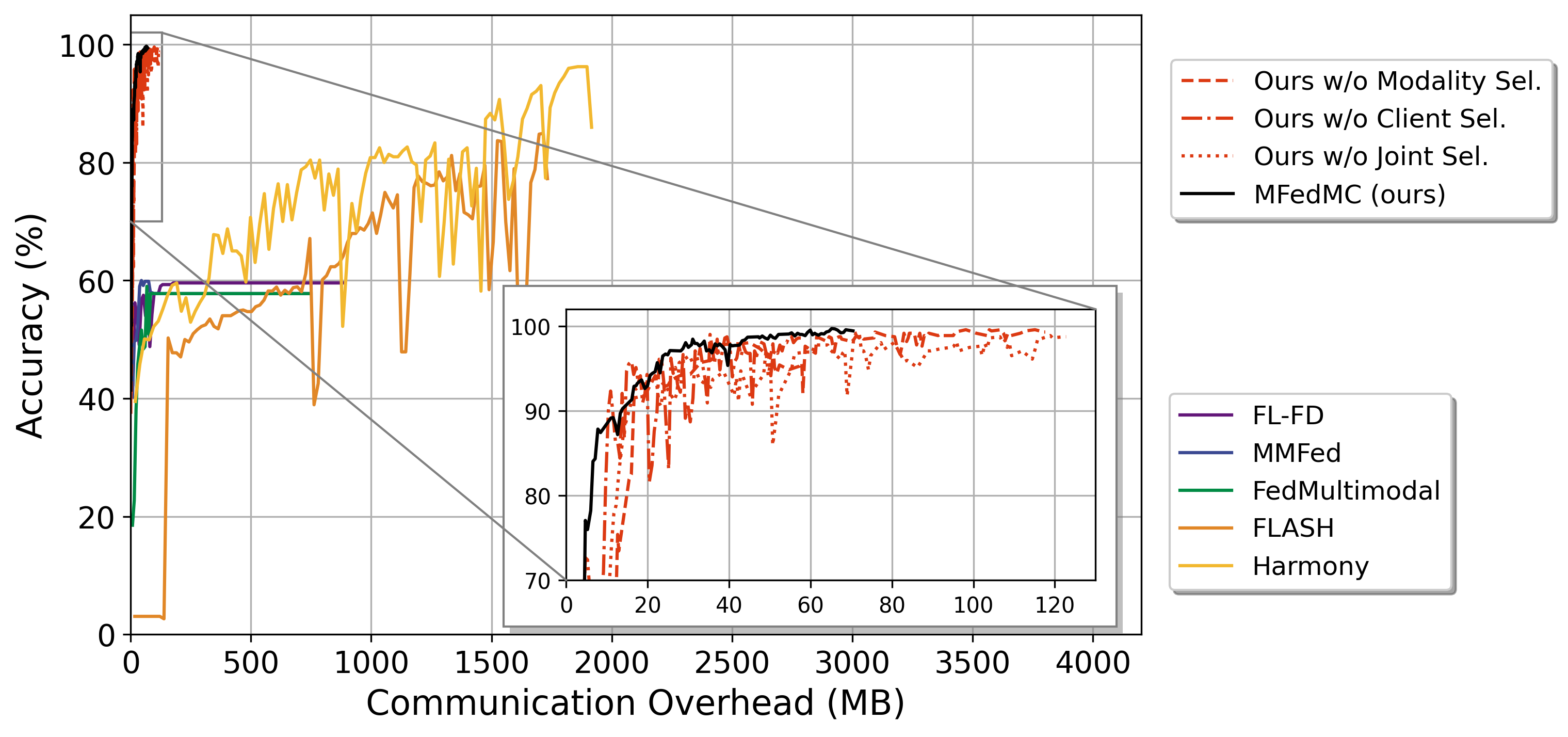}
\caption{Effect of network heterogeneity. MFedMC enables all clients to participate regardless of bandwidth constraints through strategic modality and client selection, achieving comparable final accuracy to homogeneous settings while most baselines either converge to low accuracy or exhibit significantly slower improvement due to inability to handle heterogeneous network conditions.}
\label{Fig. Effect_Heterogeneous_Network} 
\end{figure}

Fig. \ref{Fig. Effect_Heterogeneous_Network} shows the result of the proposed MFedMC and eight baselines in the heterogeneous network scenario. Compared to the homogeneous communication scenario depicted in Fig. \ref{Fig. Acc_vs_Comm}, we see that while communication constraints lead to a drop in accuracy in the early stages of the MFedMC framework, our methodology ultimately reaches roughly the same accuracy. This demonstrates the capability of the MFedMC framework to handle heterogeneous network conditions, ensuring that all clients, regardless of their individual restrictions, contribute to and benefit from the FL process. It is noteworthy that most SOTA baseline methods are not designed to operate effectively under heterogeneous network conditions. If clients have varying network bandwidths, which is common, the majority of clients (in our setup, Clients 3-9) will be unable to upload complete models to the server, thereby being unable to contribute to FL due to bandwidth limitations. As shown in Fig. \ref{Fig. Effect_Heterogeneous_Network}, this limitation causes some SOTA baselines (FL-FD, MMFed, and FedMultimodal) to converge to a low accuracy level, reflecting the maximum performance achievable using data from only Clients 1-2. Other SOTA baselines (FLASH and Harmony) allow uploading parts of the complete model; thus, although they can continue to improve accuracy, the improvement is much slower compared to our MFedMC algorithm that selects modalities and clients. In contrast, our modality selection strategically prioritizes and optimizes data transmission, enabling all clients to continuously participate in the FL network without being hindered by their individual bandwidth restrictions.

\subsection{Effect of Long-Tail Distribution}
\label{Sec. Effect of Long-Tail Distribution}

To investigate the robustness of MFedMC under imbalanced client data distributions, particularly considering that lower loss-based client selection might favor clients with smaller or simpler datasets, we conduct experiments with varying degrees of long-tail distribution across clients. Following prior work [57], we control the client sample distribution through the Imbalance Factor (IF), where higher IF values represent more severe long-tail distributions with greater disparity in sample sizes across clients. Furthermore, we explore whether incorporating the recency term from modality selection into client selection could mitigate potential bias caused by loss-based selection. We compare five client selection strategies with different loss-recency weight combinations: pure loss-based selection, hybrid strategies that balance loss and recency considerations, and pure recency-based selection that prioritizes clients not recently selected.

\begin{figure}[t]
\centering
\includegraphics[width=\linewidth]{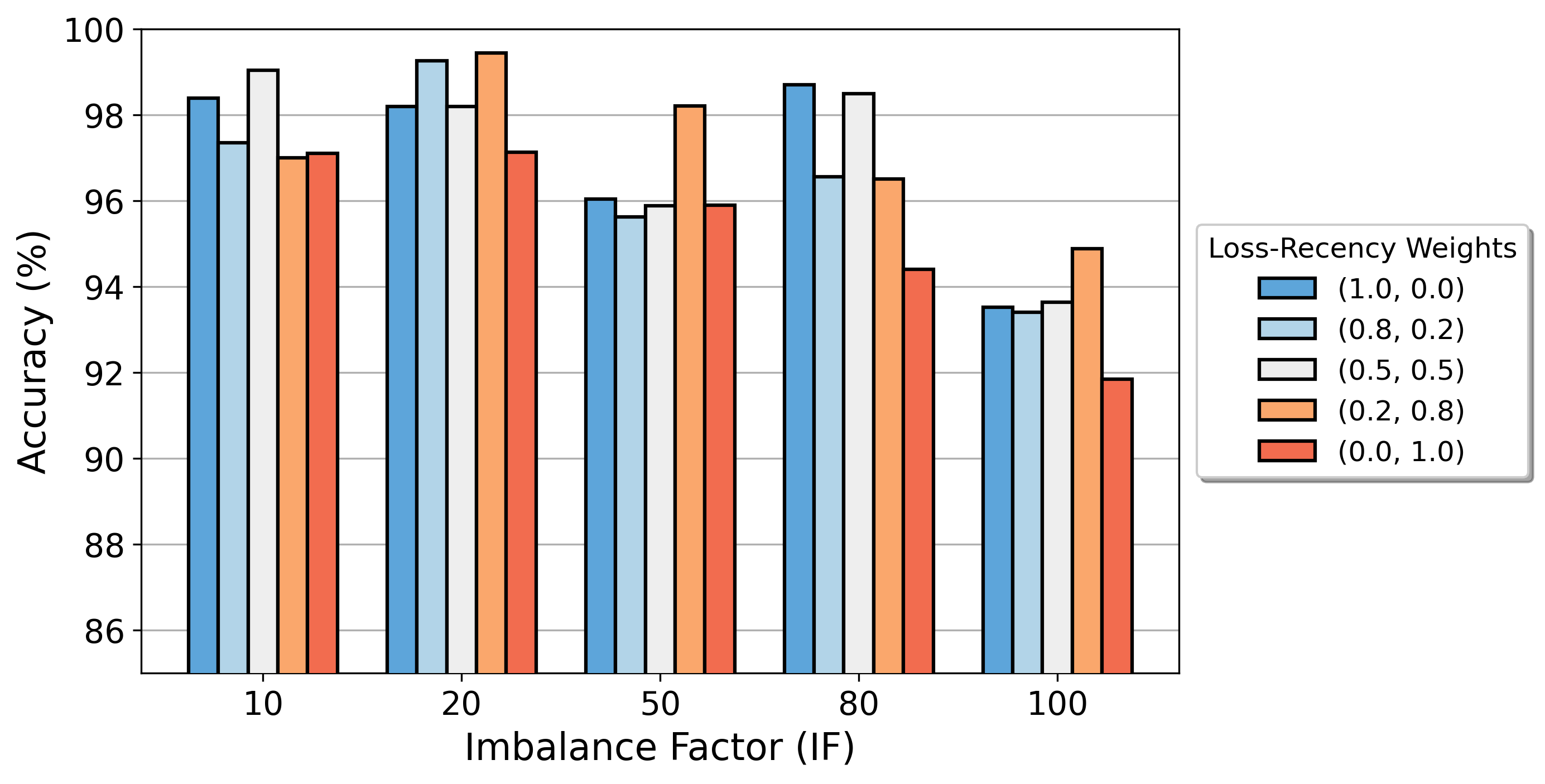}
\caption{Effect of imbalance data distribution and loss-recency weight combinations. Pure loss-based selection (1.0, 0.0) maintains competitive performance across all imbalance levels, while hybrid strategies incorporating recency show marginal improvements with varying optimal configurations across different imbalance factors.}
\label{Fig. Effect_Longtail} 
\end{figure}

As shown in Fig. \ref{Fig. Effect_Longtail}, several key observations emerge from our experiments. First, pure lower loss-based selection (weight configuration $(1.0, 0.0)$) maintains acceptable performance across all imbalance levels, remaining competitive even under extreme conditions (IF=100). This suggests that while selecting clients solely based on lower local loss may have theoretical limitations, its practical impact appears more limited than initially anticipated. Moreover, the rationale for lower loss-based selection is fundamentally aligned with MFedMC's core objective of communication efficiency. In communication-constrained FL, convergence speed directly determines the total communication cost required to reach target accuracy. By prioritizing clients with lower losses, this selection strategy consistently aggregates encoder updates, thereby accelerating convergence and reducing the communication overhead needed to achieve satisfactory performance. Second, while incorporating recency provides some improvements in certain scenarios (e.g., $(0.2, 0.8)$ performs best at IF=20, 50, 100), the benefits are limited and inconsistent. Compared with pure lower loss-based selection, improvements are often marginal, optimal configurations vary significantly across imbalance levels (e.g., $(0.2, 0.8)$ degrades at IF=10, 80), and the performance gap remains small across all tested factors.

These findings indicate that while incorporating recency has conceptual merit, it introduces hyperparameter sensitivity without consistent improvements and may compromise communication efficiency. Notably, the role of recency in client selection fundamentally differs from its role in modality selection. In modality selection, incorporating recency to diversify modality choices is beneficial because multiple modality encoders jointly contribute to predictions through the fusion module, which can adaptively weigh and compensate for weaker modality predictions. In contrast, client selection aims to identify clients with the best-trained modality encoders for global aggregation, where selecting poorly-trained encoders (even for diversity) directly slows convergence and increases the communication overhead required to reach target accuracy. Therefore, we maintain the pure lower loss-based client selection strategy, which prioritizes high-quality encoders to maximize aggregation effectiveness and communication efficiency.

\begin{figure*}[t]
\centering
\subfloat[Accuracy vs. Communication Overhead (50\% Availability)]{\includegraphics[height=4.8cm]{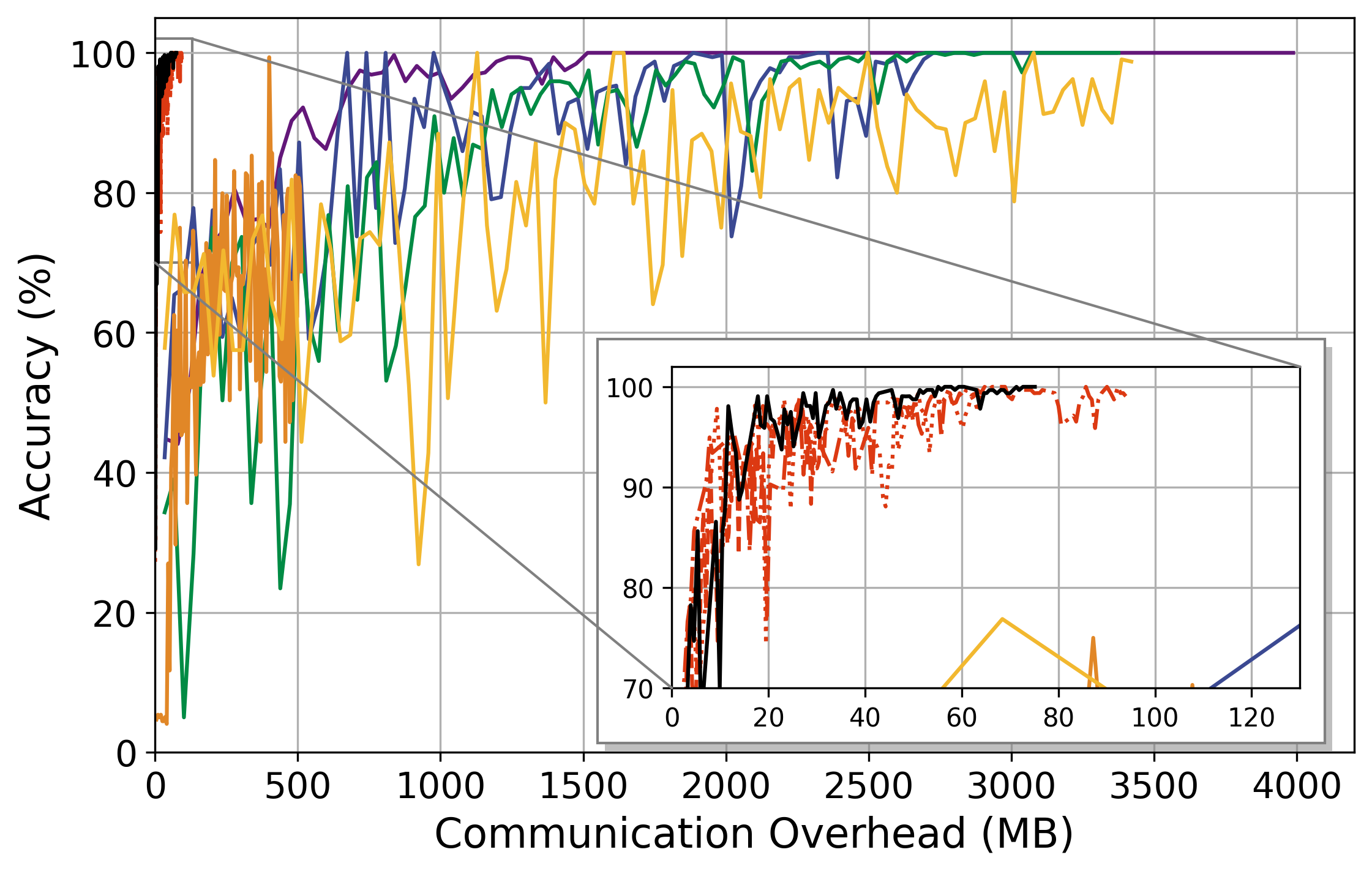}}
\subfloat[Accuracy vs. Client Availability Rate]{\includegraphics[height=4.8cm]{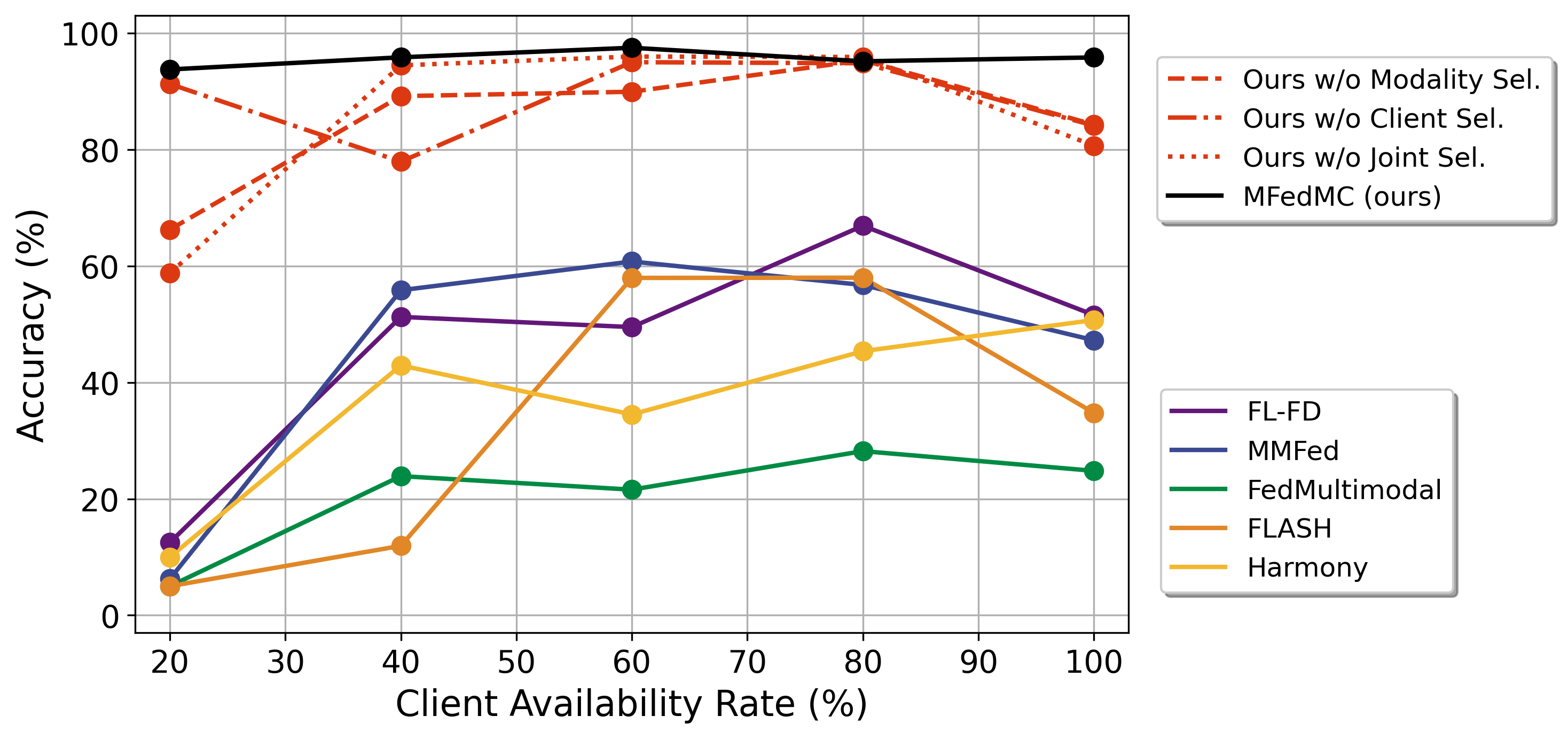}}
\caption{Effect of client availability. MFedMC achieves faster convergence with lower communication overhead and maintains superior accuracy across different client availability rates compared to baseline methods.}
\label{Fig. Effect_Client_Availability}
\end{figure*}

\subsection{Effect of Client Dynamics and Availability}

Fig. \ref{Fig. Effect_Client_Availability} demonstrates the robustness of our framework under client dynamics caused by stragglers and client churn, where MFedMC and baselines can only perform federated training on a subset of available clients. We observe that MFedMC and its ablation variants maintain superior performance even when a large proportion of clients are unavailable. In contrast, client availability has a more pronounced impact on SOTA baselines, particularly under lower availability rates. This is because when client availability decreases, baseline methods struggle to acquire sufficient global representations as they require all clients to upload complete models in each round. Our method, however, not only obtains richer global representations through reduced communication overhead (enabling more frequent aggregation rounds), but also compensates for incomplete global information through the personalized fusion module strategy. Specifically, the local fusion module can adaptively weight and combine the available modality encoders based on local data characteristics, effectively mitigating the negative impact of missing updates from unavailable clients. This design makes MFedMC naturally resilient to client churn and stragglers, as the personalized fusion module serves as a robust buffer that maintains performance stability even when global modality encoders receive updates from only a subset of clients.

\begin{figure}[t]
\centering
\includegraphics[width=1\linewidth]{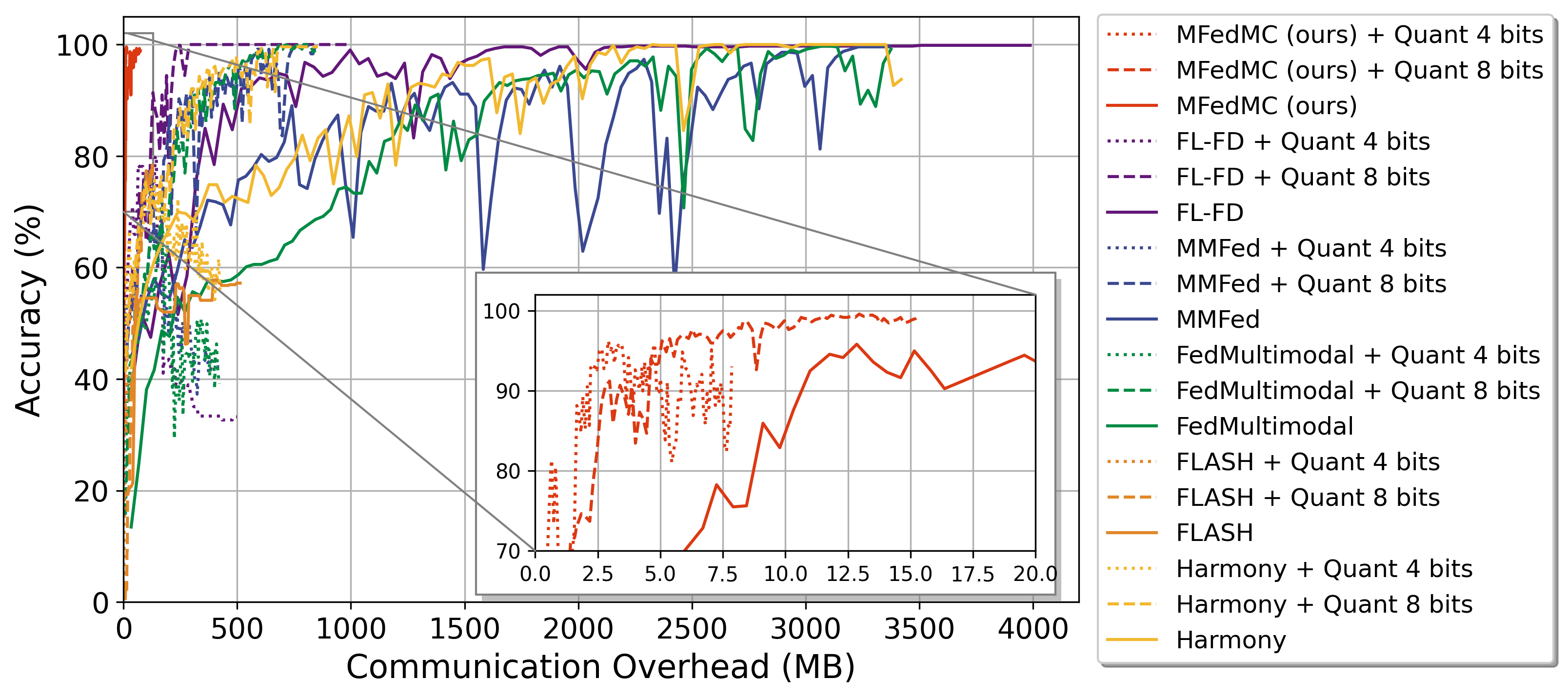}
\caption{Effect of quantization precision. MFedMC is compatible with different quantization levels (4-bit and 8-bit) and consistently achieves faster convergence with lower communication overhead compared to baseline methods.}
\label{Fig. Effect_Quant} 
\end{figure}

\subsection{Integration of Communication Compression} 

Our MFedMC framework reduces communication overhead through joint modality and client selection, and also can integrate with communication compression algorithms to further minimize communication costs and bandwidth requirements. As illustrated in Fig. \ref{Fig. Effect_Quant}, comparative analysis reveals that our method consistently outperforms baseline approaches when combined with quantization techniques, demonstrating accelerated convergence and reduced communication costs across all precision levels. Particularly noteworthy is the performance differentiation at varying quantization levels: while all methods maintain normal convergence at 8-bit quantization, only our approach sustains acceptable accuracy at 4-bit quantization, whereas baseline methods fail to converge entirely. This robust performance can be attributed to our personalized local fusion module, which effectively mitigates quantization errors. Conversely, baseline methods amplify these errors as they propagate through the fusion module directly connected to the task head. Consequently, the distinctive architecture of localized fusion modules within the MFedMC framework demonstrates remarkable compatibility with quantization-induced perturbations, enabling effective integration with additional communication compression algorithms to achieve further reductions in transmission overhead.

\begin{table}[t]
\centering
\caption{End-to-End System-Level Runtime Comparison}
\label{Table System Time}
\vspace{-2mm}
\resizebox{\linewidth}{!}{
\begin{tabular}{@{}lccc@{}}
\toprule
\multirow{2}{*}{\textbf{Method}} & \textbf{Training} & \textbf{Communication} & \textbf{Total Time} \\
& \textbf{Time (mins)} & \textbf{Time (mins)} & \textbf{(mins)} \\
\midrule
FL-FD~\cite{qi2023fl} & 31.11 & 100.29 & 131.40 \\
MMFed~\cite{xiong2022unified} & 27.03 & 130.15 & 157.18 \\
FedMultimodal~\cite{feng2023fedmultimodal} & 29.44 & 130.36 & 159.80 \\
FLASH~\cite{salehi2022flash} & 27.77 & 130.15 & 157.91 \\
Harmony~\cite{ouyang2023harmony} & 37.60 & 130.15 & 167.75 \\
\midrule
Ours w/o Modality Sel. & 19.52 & 1.75 & 21.27 \\
Ours w/o Client Sel. & 23.57 & 1.63 & 25.20 \\
Ours w/o Joint Sel. & 19.13 & 1.87 & 21.00 \\
\midrule
& & & \\
\multirow{-2}{*}{MFedMC (Ours)} & \multirow{-2}{*}{23.56} & \multirow{-2}{*}{2.10} & \multirow{-2}{*}{25.67} \\
\bottomrule
\end{tabular}
}
\end{table}

\subsection{Computational Efficiency Analysis}

\textbf{End-to-End System-Level Time Analysis.} We measure the actual wall-clock training time for all methods on the ActionSense dataset over 100 communication rounds. The total time comprises two components: (1) \textit{Training time}: the actual wall-clock time for local model training, and (2) \textit{Communication time}: the simulated transmission time based on realistic network conditions with an uplink bandwidth of 10 Mbps (representative of typical edge/IoT devices), a protocol encoding overhead of 1.2$\times$, and a forward error correction (FEC) overhead of 1.5$\times$, calculated as $T_{\text{comm}} = \text{Model Size (bytes)} \times 1.2 \times 1.5 \div (10 \times 10^6 / 8)$ seconds. As shown in Table \ref{Table System Time}, MFedMC achieves a comparable or lower training time relative to baselines (23.56 mins vs. 27-37 mins) while demonstrating a significant reduction in communication time (2.10 mins vs. 100-130 mins). Although MFedMC requires slightly more training time than some of our ablation variants due to Shapley value computation, this modest computational overhead is well justified by the accuracy improvements (4-10\% as shown in Table 2) and the 50-60$\times$ reduction in communication time. These results confirm that communication is the dominant bottleneck, accounting for 76-84\% of the total time for baseline methods. Our joint selection strategy effectively addresses this challenge, achieving an overall 5-6$\times$ speedup in end-to-end training time.

\begin{figure}[t]
    \centering
    \subfloat[Effect of Modality Number]{\includegraphics[width=0.5\linewidth]{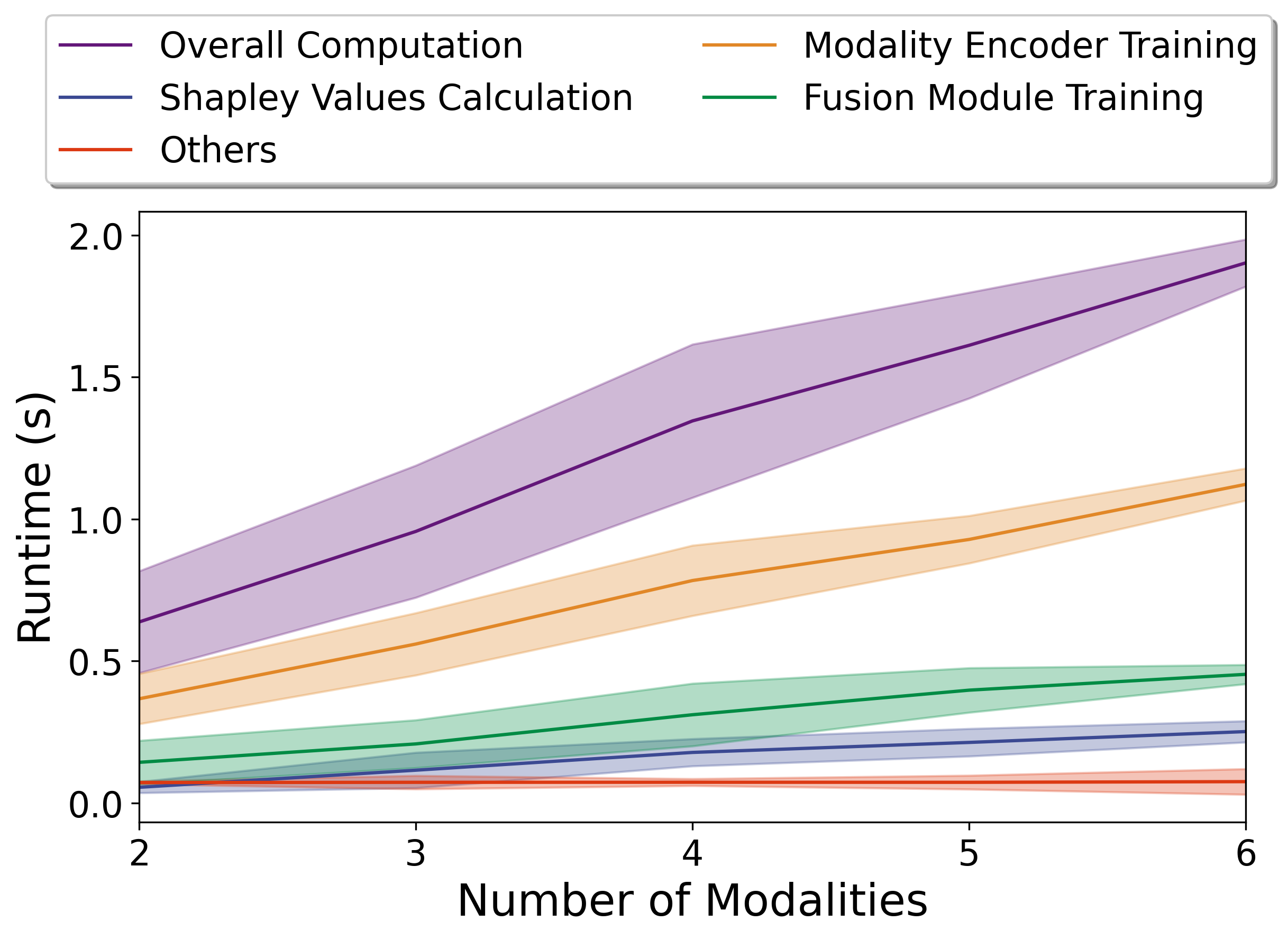}\label{Fig. Runtime Modality}}
    \subfloat[Effect of Subsampling]{\includegraphics[width=0.5\linewidth]{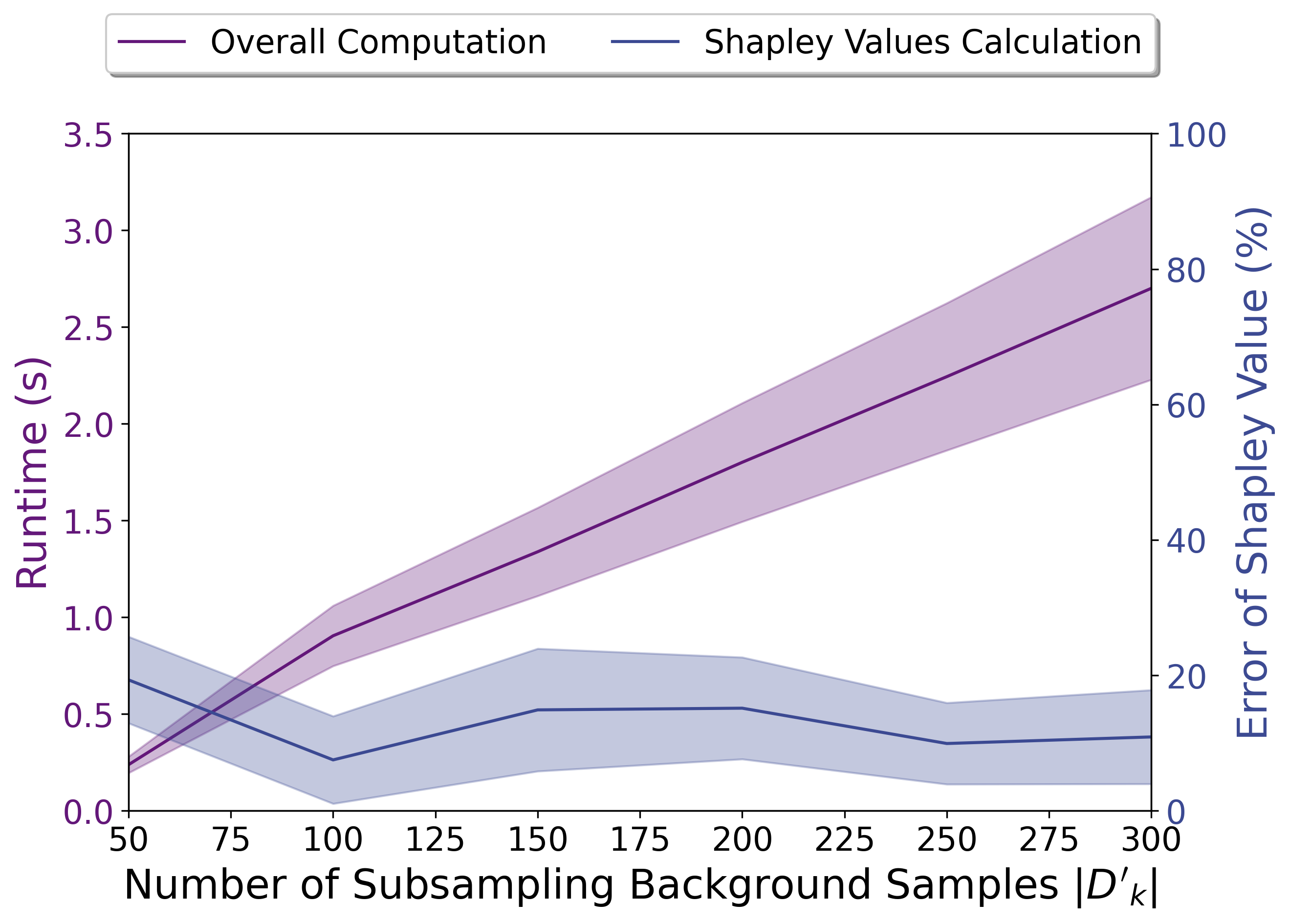}\label{Fig. Runtime Subsampling}}
    \caption{Runtime analysis of Shapley value computation. Solid lines represent mean values, while shaded regions indicate standard deviation.}
\label{Fig. Runtime}
\end{figure}

\noindent\textbf{Shapley Value Computation Overhead.} Fig. \ref{Fig. Runtime Modality} illustrates the impact of varying the number of modalities on runtime performance. The computational overhead of Shapley value calculation is independent of the number of parameters in the modality encoders. According to the computational complexity of $\mathcal{O}(N_{\omega_k} L_{\omega_k} H_{\omega_k} |D'_k|)$ derived in Sec. \ref{Sec. Server Aggregation and Local Retention}, the runtime associated with Shapley value computation remains independent of the number of modalities, exhibiting only a marginal increase attributable to auxiliary data processing operations such as normalization. Relative to the dominant computational components, modality encoder training and fusion module training, the runtime for Shapley value calculation is substantially lower, demonstrating that our approach introduces no significant scalability challenges across varying numbers of modalities. Fig. \ref{Fig. Runtime Subsampling} presents the relationship between subsampling size and both runtime performance and Shapley value estimation error. As the number of subsampled points $|D'_k|$ increases, runtime exhibits linear growth consistent with the complexity bound $\mathcal{O}(N_{\omega_k} L_{\omega_k} H_{\omega_k} |D'_k|)$. Notably, the estimation error decreases only marginally from 19.33\% to 10.91\% as the background sample size increases from $|D'_k| = 50$ to $|D'_k| = 300$, demonstrating a favorable trade-off between computational efficiency and approximation fidelity.

\section{Conclusion}
\label{Sec. Conclusion}

In this paper, we introduced the MFedMC framework that minimizes communication overhead and enhances learning efficiency through joint modality and client selection strategies. By optimizing modality selection with Shapley values, modality encoder sizes, and recency, as well as favoring clients with lower local loss for client selection, we achieved considerable recognition accuracy and up to 20$\times$ increase in communication efficiency. The proposed MFedMC is flexible, suitable for heterogeneous clients, offers modular modality encoders that can be detached, and provides impact assessment for data modalities. Extensive experiments across a diverse range of real-world datasets, including wearable sensors, healthcare, NLP, and remote sensing satellite datasets, demonstrated the outstanding performance of the MFedMC framework across various data modalities.

Our future work will focus on enhancing the adaptability of MFedMC through dynamic configuration. For modality selection, we could adjust the weight of communication overhead based on the dynamically available communication resources in the real world (such as higher bandwidth at night), allowing for the uploading of more modality encoders. For client selection, we may consider a dynamic strategy based on local loss, employing selection of higher local losses at the initial stages of FL to speed up convergence, and transitioning to the proposed lower local loss selection towards the end to optimize local minima. 

\bibliographystyle{IEEEtran}
\small\bibliography{reference}

@article{lin2023federated,
  title={Federated learning on multimodal data: A comprehensive survey},
  author={Lin, Yi-Ming and Gao, Yuan and Gong, Mao-Guo and Zhang, Si-Jia and Zhang, Yuan-Qiao and Li, Zhi-Yuan},
  journal={Machine Intelligence Research},
  volume={20},
  number={4},
  pages={539--553},
  year={2023},
  publisher={Springer}
}

@inproceedings{chen2024fedmbridge,
  title={FedMBridge: Bridgeable multimodal federated learning},
  author={Chen, Jiayi and Zhang, Aidong},
  booktitle={Forty-first International Conference on Machine Learning},
  year={2024}
}

@article{fang2025federated,
  title={Federated Sketching LoRA: On-Device Collaborative Fine-Tuning of Large Language Models},
  author={Fang, Wenzhi and Han, Dong-Jun and Yuan, Liangqi and Hosseinalipour, Seyyedali and Brinton, Christopher G},
  journal={arXiv preprint arXiv:2501.19389},
  year={2025}
}

@article{young1985monotonic,
  title={Monotonic solutions of cooperative games},
  author={Young, H Peyton},
  journal={International Journal of Game Theory},
  volume={14},
  number={2},
  pages={65--72},
  year={1985},
  publisher={Springer}
}

@article{shapley1953value,
  title={A value for n-person games},
  author={Shapley, Lloyd S and others},
  year={1953},
  publisher={Princeton University Press Princeton}
}

@inproceedings{xiong2023client,
  title={Client-adaptive cross-model reconstruction network for modality-incomplete multimodal federated learning},
  author={Xiong, Baochen and Yang, Xiaoshan and Song, Yaguang and Wang, Yaowei and Xu, Changsheng},
  booktitle={Proceedings of the 31st ACM International Conference on Multimedia},
  pages={1241--1249},
  year={2023}
}

@article{le2024cross,
  title={Cross-modal prototype based multimodal federated learning under severely missing modality},
  author={Le, Huy Q and Thwal, Chu Myaet and Qiao, Yu and Tun, Ye Lin and Nguyen, Minh NH and Hong, Choong Seon},
  journal={arXiv preprint arXiv:2401.13898},
  year={2024}
}

@inproceedings{yuan2025local,
  title={Local-Cloud Inference Offloading for LLMs in Multi-Modal, Multi-Task, Multi-Dialogue Settings},
  author={Yuan, Liangqi and Han, Dong-Jun and Wang, Shiqiang and Brinton, Christopher},
  booktitle={Proceedings of the Twenty-Sixth International Symposium on Theory, Algorithmic Foundations, and Protocol Design for Mobile Networks and Mobile Computing},
  pages={201--210},
  year={2025}
}

@inproceedings{chen2022towards,
  title={Towards optimal multi-modal federated learning on non-IID data with hierarchical gradient blending},
  author={Chen, Sijia and Li, Baochun},
  booktitle={IEEE INFOCOM 2022-IEEE conference on computer communications},
  pages={1469--1478},
  year={2022},
  organization={IEEE}
}

@article{yu2023multimodal,
  title={Multimodal federated learning via contrastive representation ensemble},
  author={Yu, Qiying and Liu, Yang and Wang, Yimu and Xu, Ke and Liu, Jingjing},
  journal={arXiv preprint arXiv:2302.08888},
  year={2023}
}

@article{bao2023multimodal,
  title={Multimodal Federated Learning with Missing Modality via Prototype Mask and Contrast},
  author={Bao, Guangyin and Zhang, Qi and Miao, Duoqian and Gong, Zixuan and Hu, Liang},
  journal={arXiv preprint arXiv:2312.13508},
  year={2023}
}

@inproceedings{zheng2023autofed,
  title={Autofed: Heterogeneity-aware federated multimodal learning for robust autonomous driving},
  author={Zheng, Tianyue and Li, Ang and Chen, Zhe and Wang, Hongbo and Luo, Jun},
  booktitle={Proceedings of the 29th Annual International Conference on Mobile Computing and Networking},
  pages={1--15},
  year={2023}
}

@inproceedings{mcmahan2017communication,
  title={Communication-efficient learning of deep networks from decentralized data},
  author={McMahan, Brendan and Moore, Eider and Ramage, Daniel and Hampson, Seth and y Arcas, Blaise Aguera},
  booktitle={Artificial intelligence and statistics},
  pages={1273--1282},
  year={2017},
  organization={PMLR}
}

@article{xiong2022unified,
  title={A unified framework for multi-modal federated learning},
  author={Xiong, Baochen and Yang, Xiaoshan and Qi, Fan and Xu, Changsheng},
  journal={Neurocomputing},
  volume={480},
  pages={110--118},
  year={2022},
  publisher={Elsevier}
}

@article{qi2023fl,
  title={{FL-FD: Federated learning-based fall detection with multimodal data fusion}},
  author={Qi, Pian and Chiaro, Diletta and Piccialli, Francesco},
  journal={Information Fusion},
  pages={101890},
  year={2023},
  publisher={Elsevier}
}

@inproceedings{zhao2022multimodal,
  title={Multimodal federated learning on iot data},
  author={Zhao, Yuchen and Barnaghi, Payam and Haddadi, Hamed},
  booktitle={2022 IEEE/ACM Seventh International Conference on Internet-of-Things Design and Implementation (IoTDI)},
  pages={43--54},
  year={2022},
  organization={IEEE}
}

@inproceedings{salehi2022flash,
  title={Flash: Federated learning for automated selection of high-band mmwave sectors},
  author={Salehi, Batool and Gu, Jerry and Roy, Debashri and Chowdhury, Kaushik},
  booktitle={IEEE INFOCOM 2022-IEEE Conference on Computer Communications},
  pages={1719--1728},
  year={2022},
  organization={IEEE}
}

@article{feng2023fedmultimodal,
  title={FedMultimodal: A Benchmark For Multimodal Federated Learning},
  author={Feng, Tiantian and Bose, Digbalay and Zhang, Tuo and Hebbar, Rajat and Ramakrishna, Anil and Gupta, Rahul and Zhang, Mi and Avestimehr, Salman and Narayanan, Shrikanth},
  journal={arXiv preprint arXiv:2306.09486},
  year={2023}
}

@inproceedings{ouyang2023harmony,
  title={Harmony: Heterogeneous Multi-Modal Federated Learning through Disentangled Model Training},
  author={Ouyang, Xiaomin and Xie, Zhiyuan and Fu, Heming and Cheng, Sitong and Pan, Li and Ling, Neiwen and Xing, Guoliang and Zhou, Jiayu and Huang, Jianwei},
  booktitle={Proceedings of the 21st Annual International Conference on Mobile Systems, Applications and Services},
  pages={530--543},
  year={2023}
}

@inproceedings{chen2022fedmsplit,
  title={FedMSplit: Correlation-adaptive federated multi-task learning across multimodal split networks},
  author={Chen, Jiayi and Zhang, Aidong},
  booktitle={Proceedings of the 28th ACM SIGKDD Conference on Knowledge Discovery and Data Mining},
  pages={87--96},
  year={2022}
}

@article{delpreto2022actionsense,
  title={ActionSense: A multimodal dataset and recording framework for human activities using wearable sensors in a kitchen environment},
  author={DelPreto, Joseph and Liu, Chao and Luo, Yiyue and Foshey, Michael and Li, Yunzhu and Torralba, Antonio and Matusik, Wojciech and Rus, Daniela},
  journal={Advances in Neural Information Processing Systems},
  volume={35},
  pages={13800--13813},
  year={2022}
}

@article{chellapandi2023federated,
  title={Federated Learning for Connected and Automated Vehicles: A Survey of Existing Approaches and Challenges},
  author={Chellapandi, Vishnu Pandi and Yuan, Liangqi and Brinton, Christopher G. and Zak, Stanislaw H and Wang, Ziran},
  journal={IEEE Transactions on Intelligent Vehicles},
  volume={9},
  number={1},
  pages={119 - 137},
  month={November},
  year={2023},
  publisher={IEEE}
}

@inproceedings{chellapandi2023survey,
  title={A Survey of Federated Learning for Connected and Automated Vehicles},
  author={Chellapandi, Vishnu Pandi and Yuan, Liangqi and {\.Z}ak, Stanislaw H and Wang, Ziran},
  booktitle={2023 IEEE 26th International Conference on Intelligent Transportation Systems (ITSC)},
  pages={2485--2492},
  year={2023},
  organization={IEEE}
}

@article{yuan2024decentralized,
  title={Decentralized Federated Learning: A Survey and Perspective},
  author={Yuan, Liangqi and Wang, Ziran and Sun, Lichao and Yu, Philip S. and Brinton, Christopher G.},
  journal={IEEE Internet of Things Journal}, 
  volume={11},
  number={21},
  pages={34617 - 34638},
  month={May},
  year={2024},
  publisher={IEEE}
}

@inproceedings{yuan2023peer,
  title={Peer-to-Peer Federated Continual Learning for Naturalistic Driving Action Recognition},
  author={Yuan, Liangqi and Ma, Yunsheng and Su, Lu and Wang, Ziran},
  booktitle={Proceedings of the IEEE/CVF Conference on Computer Vision and Pattern Recognition},
  pages={5249--5258},
  year={2023}
}

@article{yuan2023federated,
  title={Federated Transfer-Ordered-Personalized Learning for Driver Monitoring Application},
  author={Yuan, Liangqi and Su, Lu and Wang, Ziran},
  journal={IEEE Internet of Things Journal},
  volume={10},
  number={20},
  pages={18292--18301},
  month={May},
  year={2023},
  publisher={IEEE}
}

@article{lundberg2017unified,
  title={A unified approach to interpreting model predictions},
  author={Lundberg, Scott M and Lee, Su-In},
  journal={Advances in neural information processing systems},
  volume={30},
  year={2017}
}

@article{lundberg2020local2global,
  title={From local explanations to global understanding with explainable AI for trees},
  author={Lundberg, Scott M. and Erion, Gabriel and Chen, Hugh and DeGrave, Alex and Prutkin, Jordan M. and Nair, Bala and Katz, Ronit and Himmelfarb, Jonathan and Bansal, Nisha and Lee, Su-In},
  journal={Nature Machine Intelligence},
  volume={2},
  number={1},
  pages={2522--5839},
  year={2020},
  publisher={Nature Publishing Group}
}

@inproceedings{sundararajan2020many,
  title={The many Shapley values for model explanation},
  author={Sundararajan, Mukund and Najmi, Amir},
  booktitle={International conference on machine learning},
  pages={9269--9278},
  year={2020},
  organization={PMLR}
}

@article{imteaj2021survey,
  title={A survey on federated learning for resource-constrained IoT devices},
  author={Imteaj, Ahmed and Thakker, Urmish and Wang, Shiqiang and Li, Jian and Amini, M Hadi},
  journal={IEEE Internet of Things Journal},
  volume={9},
  number={1},
  pages={1--24},
  year={2021},
  publisher={IEEE}
}

@inproceedings{fang2024submodel,
  title={Submodel partitioning in hierarchical federated learning: Algorithm design and convergence analysis},
  author={Fang, Wenzhi and Han, Dong-Jun and Brinton, Christopher G},
  booktitle={ICC 2024-IEEE International Conference on Communications},
  pages={268--273},
  year={2024},
  organization={IEEE}
}

@article{wang2023device,
  title={Device sampling and resource optimization for federated learning in cooperative edge networks},
  author={Wang, Su and Morabito, Roberto and Hosseinalipour, Seyyedali and Chiang, Mung and Brinton, Christopher G},
  journal={arXiv preprint arXiv:2311.04350},
  year={2023}
}

@article{cho2020client,
  title={Client selection in federated learning: Convergence analysis and power-of-choice selection strategies},
  author={Cho, Yae Jee and Wang, Jianyu and Joshi, Gauri},
  journal={arXiv preprint arXiv:2010.01243},
  year={2020}
}

@inproceedings{anguita2013public,
  title={A public domain dataset for human activity recognition using smartphones.},
  author={Anguita, Davide and Ghio, Alessandro and Oneto, Luca and Parra, Xavier and Reyes-Ortiz, Jorge Luis and others},
  booktitle={Esann},
  volume={3},
  pages={3},
  year={2013}
}

@article{wagner2020ptb,
  title={PTB-XL, a large publicly available electrocardiography dataset},
  author={Wagner, Patrick and Strodthoff, Nils and Bousseljot, Ralf-Dieter and Kreiseler, Dieter and Lunze, Fatima I and Samek, Wojciech and Schaeffter, Tobias},
  journal={Scientific data},
  volume={7},
  number={1},
  pages={154},
  year={2020},
  publisher={Nature Publishing Group UK London}
}

@article{strodthoff2020deep,
  title={Deep learning for ECG analysis: Benchmarks and insights from PTB-XL},
  author={Strodthoff, Nils and Wagner, Patrick and Schaeffter, Tobias and Samek, Wojciech},
  journal={IEEE Journal of Biomedical and Health Informatics},
  volume={25},
  number={5},
  pages={1519--1528},
  year={2020},
  publisher={IEEE}
}

@inproceedings{poria-etal-2019-meld,
    title = "{MELD}: A Multimodal Multi-Party Dataset for Emotion Recognition in Conversations",
    author = "Poria, Soujanya  and
      Hazarika, Devamanyu  and
      Majumder, Navonil  and
      Naik, Gautam  and
      Cambria, Erik  and
      Mihalcea, Rada",
    booktitle = "Proceedings of the 57th Annual Meeting of the Association for Computational Linguistics",
    year = "2019",
    pages = "527--536",
}

@data{mrnt-8w27-22,
  doi = {10.21227/mrnt-8w27},
  url = {https://dx.doi.org/10.21227/mrnt-8w27},
  author = {Claudio Persello and Ronny Hänsch and Gemine Vivone and Kaiqiang Chen and Zhiyuan Yan and Deke Tang and Hai Huang and Michael Schmitt and Xian Sun},
  publisher = {IEEE Dataport},
  title = {2023 IEEE GRSS Data Fusion Contest: Large-Scale Fine-Grained Building Classification for Semantic Urban Reconstruction},
  year = {2022} 
}

@inproceedings{yuan2024fedmfs,
  title={Fedmfs: Federated multimodal fusion learning with selective modality communication},
  author={Yuan, Liangqi and Han, Dong-Jun and Chellapandi, Vishnu Pandi and Zak, Stanislaw H and Brinton, Christopher G},
  booktitle={ICC 2024-IEEE International Conference on Communications},
  pages={287--292},
  year={2024},
  organization={IEEE}
}

@article{yu2021fedhar,
  title={FedHAR: Semi-supervised online learning for personalized federated human activity recognition},
  author={Yu, Hongzheng and Chen, Zekai and Zhang, Xiao and Chen, Xu and Zhuang, Fuzhen and Xiong, Hui and Cheng, Xiuzhen},
  journal={IEEE Transactions on Mobile Computing},
  year={2021},
  publisher={IEEE}
}

@article{wu2020fedhome,
  title={Fedhome: Cloud-edge based personalized federated learning for in-home health monitoring},
  author={Wu, Qiong and Chen, Xu and Zhou, Zhi and Zhang, Junshan},
  journal={IEEE Transactions on Mobile Computing},
  volume={21},
  number={8},
  pages={2818--2832},
  year={2020},
  publisher={IEEE}
}

@article{baghersalimi2023decentralized,
  title={Decentralized Federated Learning for Epileptic Seizures Detection in Low-Power Wearable Systems},
  author={Baghersalimi, Saleh and Teijeiro, Tomas and Aminifar, Amir and Atienza, David},
  journal={IEEE Transactions on Mobile Computing},
  year={2023},
  publisher={IEEE}
}

@article{lim2020federated,
  title={Federated learning in mobile edge networks: A comprehensive survey},
  author={Lim, Wei Yang Bryan and Luong, Nguyen Cong and Hoang, Dinh Thai and Jiao, Yutao and Liang, Ying-Chang and Yang, Qiang and Niyato, Dusit and Miao, Chunyan},
  journal={IEEE Communications Surveys \& Tutorials},
  volume={22},
  number={3},
  pages={2031--2063},
  year={2020},
  publisher={IEEE}
}

@article{sattler2019robust,
  title={Robust and communication-efficient federated learning from non-iid data},
  author={Sattler, Felix and Wiedemann, Simon and M{\"u}ller, Klaus-Robert and Samek, Wojciech},
  journal={IEEE transactions on neural networks and learning systems},
  volume={31},
  number={9},
  pages={3400--3413},
  year={2019},
  publisher={IEEE}
}

@article{kairouz2021advances,
  title={Advances and open problems in federated learning},
  author={Kairouz, Peter and McMahan, H Brendan and Avent, Brendan and Bellet, Aur{\'e}lien and Bennis, Mehdi and Bhagoji, Arjun Nitin and Bonawitz, Kallista and Charles, Zachary and Cormode, Graham and Cummings, Rachel and others},
  journal={Foundations and Trends{\textregistered} in Machine Learning},
  volume={14},
  number={1--2},
  pages={1--210},
  year={2021},
  publisher={Now Publishers, Inc.}
}

@article{lee2019adaptive,
  title={An adaptive sensor fusion framework for pedestrian indoor navigation in dynamic environments},
  author={Lee, Gunwoo and Jung, Suk-Hoon and Han, Dongsoo},
  journal={IEEE Transactions on Mobile Computing},
  volume={20},
  number={2},
  pages={320--336},
  year={2019},
  publisher={IEEE}
}

@article{li2022adaptive,
  title={Adaptive deep feature fusion for continuous authentication with data augmentation},
  author={Li, Yantao and Liu, Li and Qin, Huafeng and Deng, Shaojiang and El-Yacoubi, Mounim A and Zhou, Gang},
  journal={IEEE Transactions on Mobile Computing},
  year={2022},
  publisher={IEEE}
}

@article{zhang2023multimodal,
  title={Multimodal fusion framework based on statistical attention and contrastive attention for sign language recognition},
  author={Zhang, Jiangtao and Wang, Qingshan and Wang, Qi and Zheng, Zhiwen},
  journal={IEEE Transactions on Mobile Computing},
  year={2023},
  publisher={IEEE}
}

@article{yuan2024digital,
  title={Digital Ethics in Federated Learning},
  author={Yuan, Liangqi and Wang, Ziran and Brinton, Christopher G.},
  journal={IEEE Internet Computing}, 
  volume={28},
  number={5},
  pages={66 - 74},
  month={December},
  year={2024},
  publisher={IEEE}
}

@article{pan2023contextual,
  title={Contextual Client Selection for Efficient Federated Learning over Edge Devices},
  author={Pan, Qiying and Cao, Hangrui and Zhu, Yifei and Liu, Jiangchuan and Li, Bo},
  journal={IEEE Transactions on Mobile Computing},
  year={2023},
  publisher={IEEE}
}

@article{xu2023federated,
  title={Federated Learning with Client Selection and Gradient Compression in Heterogeneous Edge Systems},
  author={Xu, Yang and Jiang, Zhida and Xu, Hongli and Wang, Zhiyuan and Qian, Chen and Qiao, Chunming},
  journal={IEEE Transactions on Mobile Computing},
  year={2023},
  publisher={IEEE}
}

@inproceedings{chu2022mitigating,
  title={Mitigating biases in student performance prediction via attention-based personalized federated learning},
  author={Chu, Yun-Wei and Hosseinalipour, Seyyedali and Tenorio, Elizabeth and Cruz, Laura and Douglas, Kerrie and Lan, Andrew and Brinton, Christopher},
  booktitle={Proceedings of the 31st ACM International Conference on Information \& Knowledge Management},
  pages={3033--3042},
  year={2022}
}

@inproceedings{luo2022tackling,
  title={Tackling system and statistical heterogeneity for federated learning with adaptive client sampling},
  author={Luo, Bing and Xiao, Wenli and Wang, Shiqiang and Huang, Jianwei and Tassiulas, Leandros},
  booktitle={IEEE INFOCOM 2022-IEEE conference on computer communications},
  pages={1739--1748},
  year={2022},
  organization={IEEE}
}

@article{fu2023client,
  title={Client selection in federated learning: Principles, challenges, and opportunities},
  author={Fu, Lei and Zhang, Huanle and Gao, Ge and Zhang, Mi and Liu, Xin},
  journal={IEEE Internet of Things Journal},
  year={2023},
  publisher={IEEE}
}

@inproceedings{abdelmoniem2022empirical,
  title={Empirical analysis of federated learning in heterogeneous environments},
  author={Abdelmoniem, Ahmed M and Ho, Chen-Yu and Papageorgiou, Pantelis and Canini, Marco},
  booktitle={Proceedings of the 2nd European Workshop on Machine Learning and Systems},
  pages={1--9},
  year={2022}
}

@article{zhang2022fedada,
  title={FedAda: Fast-convergent adaptive federated learning in heterogeneous mobile edge computing environment},
  author={Zhang, Jinghui and Cheng, Xinyu and Wang, Cheng and Wang, Yuchen and Shi, Zhan and Jin, Jiahui and Song, Aibo and Zhao, Wei and Wen, Liangsheng and Zhang, Tingting},
  journal={World Wide Web},
  volume={25},
  number={5},
  pages={1971--1998},
  year={2022},
  publisher={Springer}
}

\begin{IEEEbiography}
[{\includegraphics[width=1in,height=1.25in,clip,keepaspectratio]{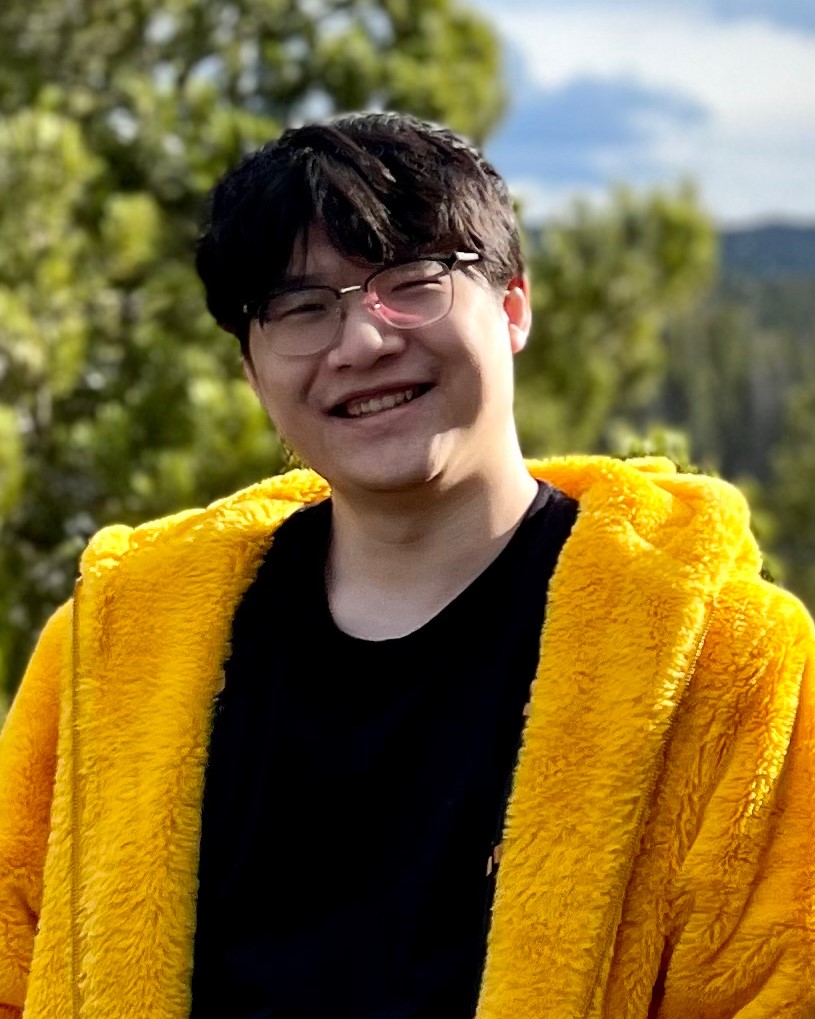}}]
{Liangqi Yuan}
(Student Member, IEEE) received the B.E. degree in Photo-electronic Information Science and Engineering from Beijing Information Science and Technology University, Beijing, China, in 2020, and the M.S. degree in Electrical and Computer Engineering from Oakland University, Rochester, MI, USA, in 2022. He is currently pursuing the Ph.D. degree in Electrical and Computer Engineering in the School of Electrical and Computer Engineering at Purdue University, West Lafayette, IN, USA. His research interests include multimodal learning, mobile computing, and machine learning.
\end{IEEEbiography}

\begin{IEEEbiography}
[{\includegraphics[width=1in,height=1.25in,clip,keepaspectratio]{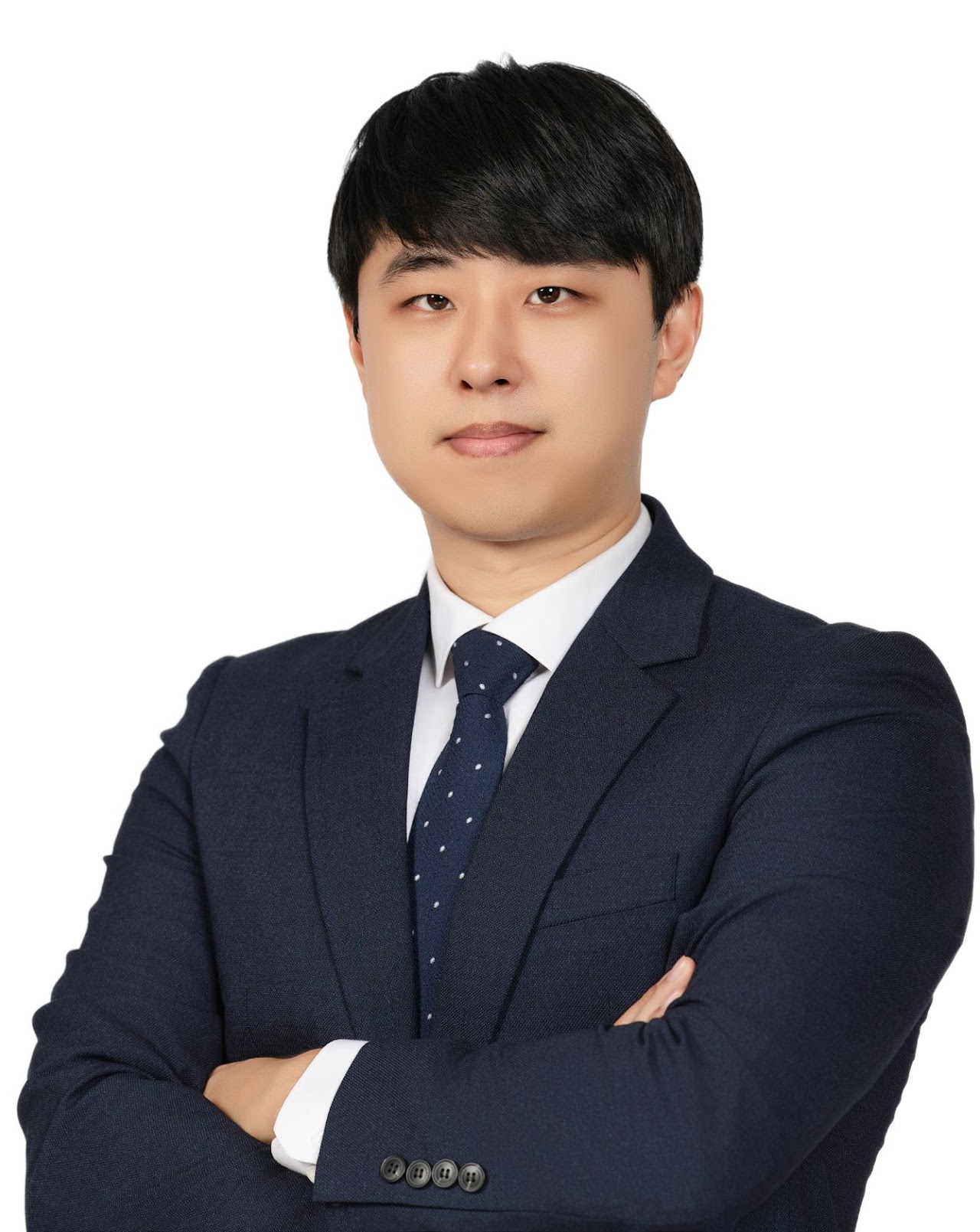}}]
{Dong-Jun Han}
(Member, IEEE) received the B.S. degree in mathematics and electrical engineering and the M.S. and Ph.D. degrees in electrical engineering from Korea Advanced Institute of Science and Technology (KAIST), South Korea, in 2016, 2018, and 2022, respectively. He is currently an Assistant Professor with the Department of Computer Science and Engineering, Yonsei University, South Korea. His research interests include the intersection of communications, networking, and machine learning, specifically in distributed/federated machine learning and network optimization.
\end{IEEEbiography}

\begin{IEEEbiography}
[{\includegraphics[width=1in,height=1.25in,clip,keepaspectratio]{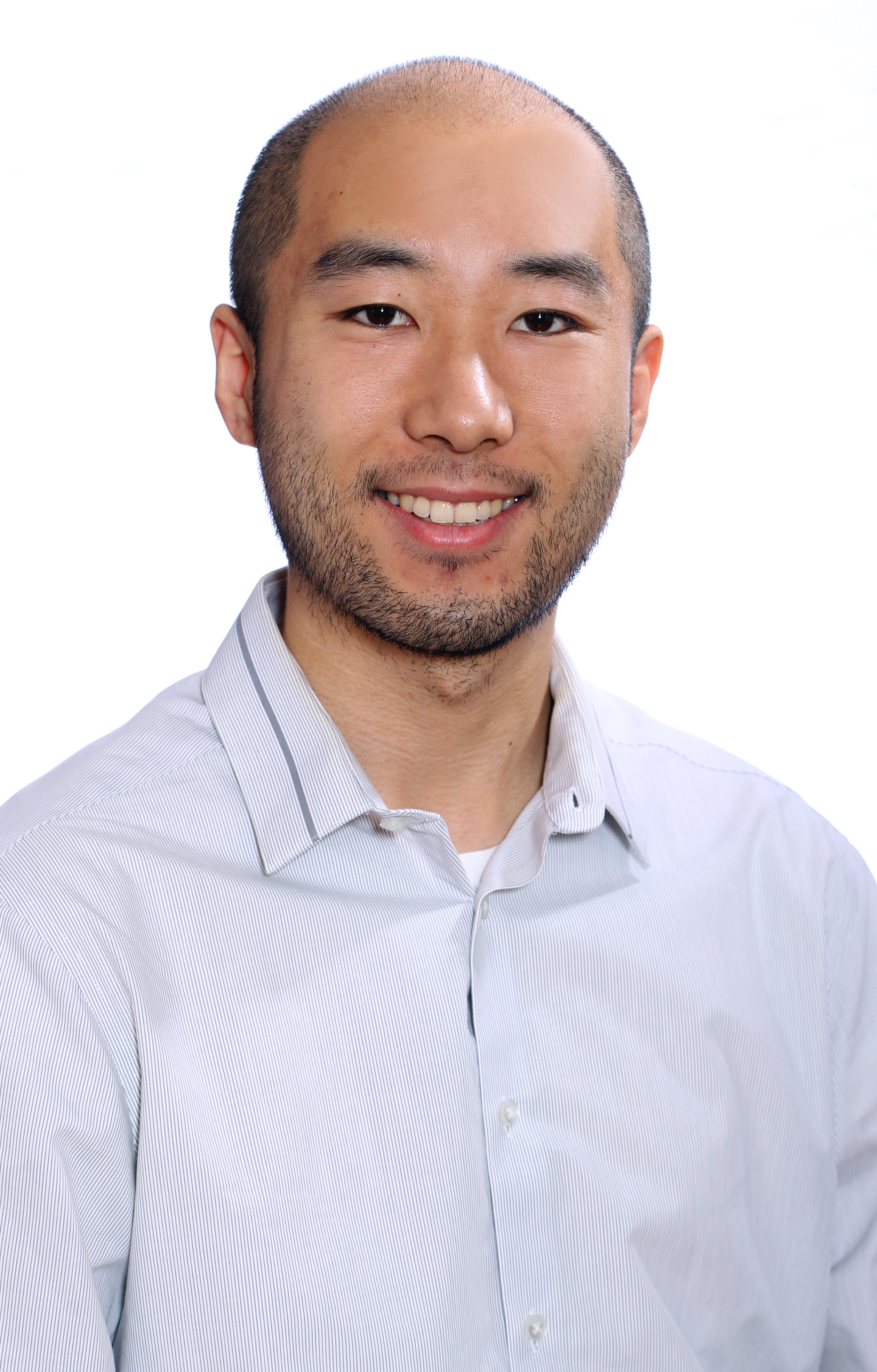}}]
{Su Wang} received his B.S. (with distinction) and Ph.D. in Electrical Engineering from Purdue University, West Lafayette, IN, USA, in 2018 and 2023, respectively. He is currently a joint lecturer and postdoctoral research associate in the School of Electrical and Computer Engineering at Princeton University.
\end{IEEEbiography}

\begin{IEEEbiography}
[{\includegraphics[width=1in,height=1.25in,clip,keepaspectratio]{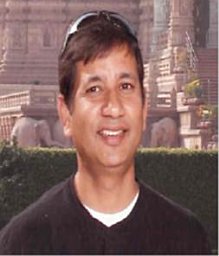}}]
{Devesh Upadhyay}
(Senior Member, IEEE) received the M.S. and Ph.D. degree in mechanical engineering from The Ohio State University, Columbus. He is currently the Technical Director for AI/ML and Autonomy at Saab Inc. Before joining Saab Devesh was a Senior Technical Leader at Ford Research where he led the Core AI/ML Quantum Computing team.
\end{IEEEbiography}

\begin{IEEEbiography}
[{\includegraphics[width=1in,height=1.25in,clip,keepaspectratio]{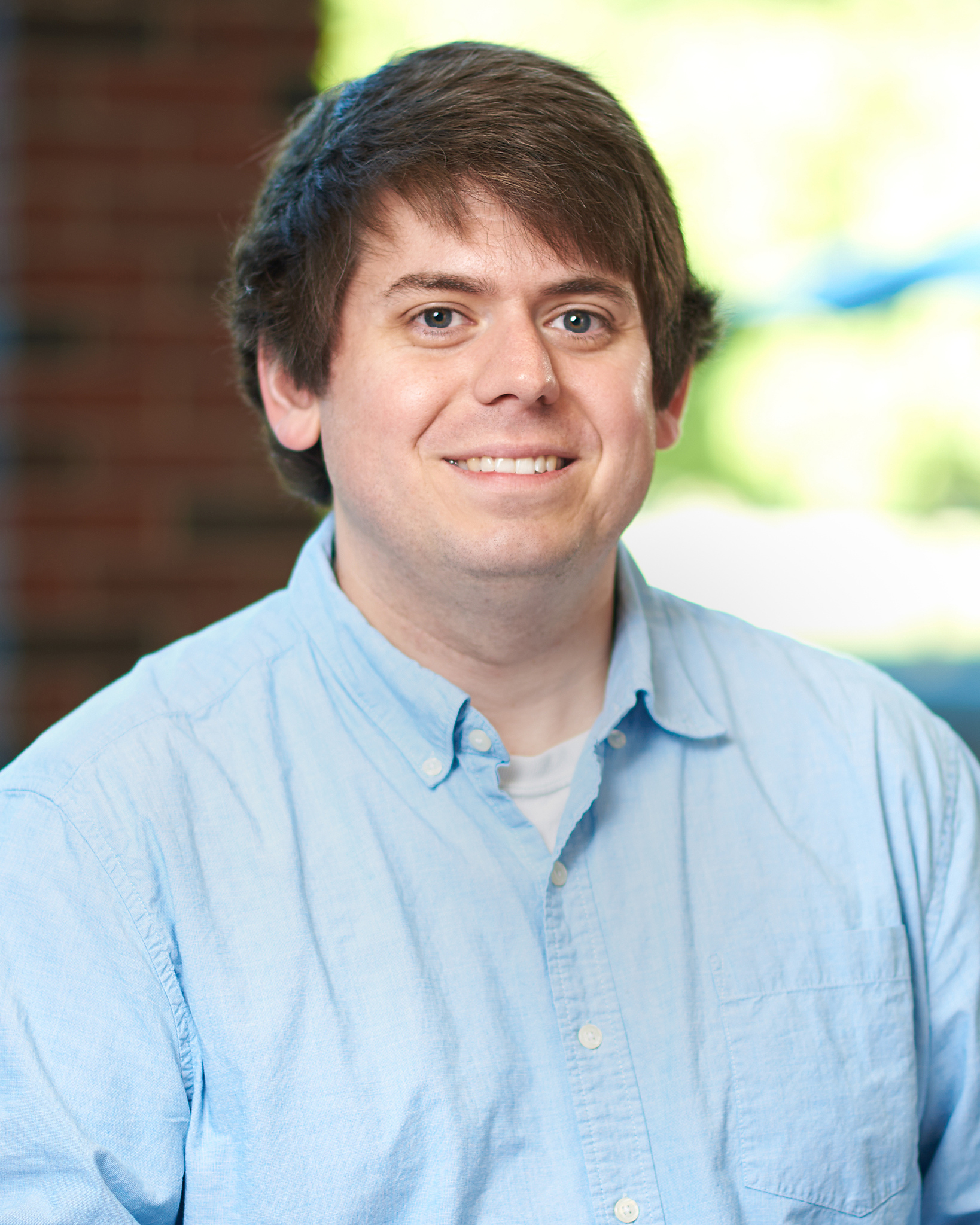}}]
{Christopher G. Brinton}
(Senior Member, IEEE) is the Elmore Associate Professor of ECE at Purdue University. He received his Ph.D. and M.Sc. in EE from Princeton University in 2016 and 2013, respectively. He is a recipient of four of the US top early career awards, from the National Science Foundation (CAREER), Office of Naval Research (YIP), Defense Advanced Research Projects Agency (YFA), and Air Force Office of Scientific Research (YIP). He is also a recipient of the Intel Rising Star Faculty Award and the Qualcomm Faculty Award.
\end{IEEEbiography}

\end{document}